\pdfoutput=1

\documentclass[10pt,twocolumn,letterpaper]{article}

\usepackage{cvpr}      
\usepackage{graphicx}
\usepackage{amsmath}
\usepackage{amssymb}
\usepackage{booktabs}
\usepackage{animate}
\usepackage{enumitem}
\usepackage{multirow}
\usepackage{diagbox}
\usepackage{booktabs}
\usepackage{graphicx}
\usepackage{amsmath}
\usepackage{amssymb}
\usepackage{booktabs}
\usepackage{float}
\usepackage{diagbox}
\usepackage{multirow}
\usepackage{enumitem}
\usepackage{tabularx}
\usepackage{microtype}
\usepackage{soul}

\makeatletter
\@namedef{ver@everyshi.sty}{}
\makeatother
\usepackage{tikz}
\usetikzlibrary{decorations.fractals,spy}
\usepackage{makecell}
\usepackage[accsupp]{axessibility}

\usepackage[pagebackref,breaklinks,colorlinks]{hyperref}

\usepackage[capitalize]{cleveref}
\crefname{section}{Sec.}{Secs.}
\Crefname{section}{Section}{Sections}
\Crefname{table}{Table}{Tables}
\crefname{table}{Tab.}{Tabs.}

\def\cvprPaperID{3275} 
\def\confName{CVPR}
\def\confYear{2024}

\definecolor{Gray}{gray}{0.5}
\definecolor{LightCyan}{rgb}{0.88,1,1}

\newcolumntype{a}{>{\columncolor{Gray}}c}
\newcolumntype{b}{>{\columncolor{white}}c}

\def\eg{\textit{e.g.}}

\renewcommand{\thefootnote}{\fnsymbol{footnote}}

\newcommand{\midparaheading}[1]{\vspace*{-0.8em}\paragraph{#1}}

\newcommand{\unrfunc}{h}

\newcommand{\scinum}[2]{$#1$$\cdot$$10^{#2}$}

\begin{document}


\title{
    Learning with Unreliability: Fast Few-shot Voxel Radiance Fields \\with Relative Geometric Consistency
    \vspace{-4mm}}

\author{%
Yingjie Xu$^{1,2}$\footnotemark[1]\quad Bangzhen Liu$^{2}$\footnotemark[1]\quad Hao Tang$^{1,3}$ \quad Bailin Deng$^{4}$\quad Shengfeng He$^{1}$\footnotemark[2] \\
$^{1}$Singapore Management University \quad $^{2}$South China University of Technology \\
$^{3}$Nanjing University of Science and Technology $^{4}$Cardiff University\\
}


\teaser{
    \centering
        \begin{subfigure}{.145\linewidth}
        \centering
        \includegraphics[width=\linewidth]{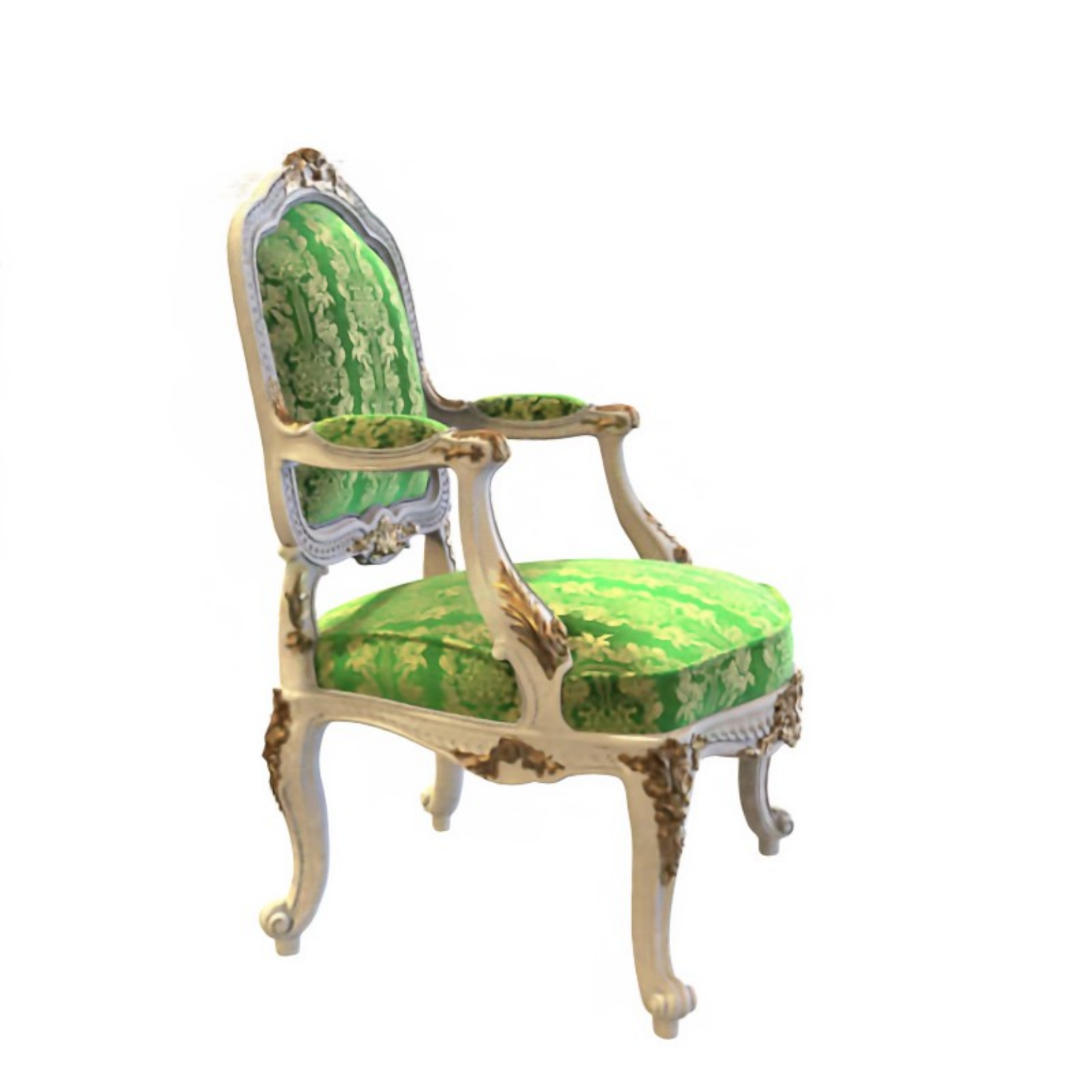}
        \includegraphics[width=\linewidth]{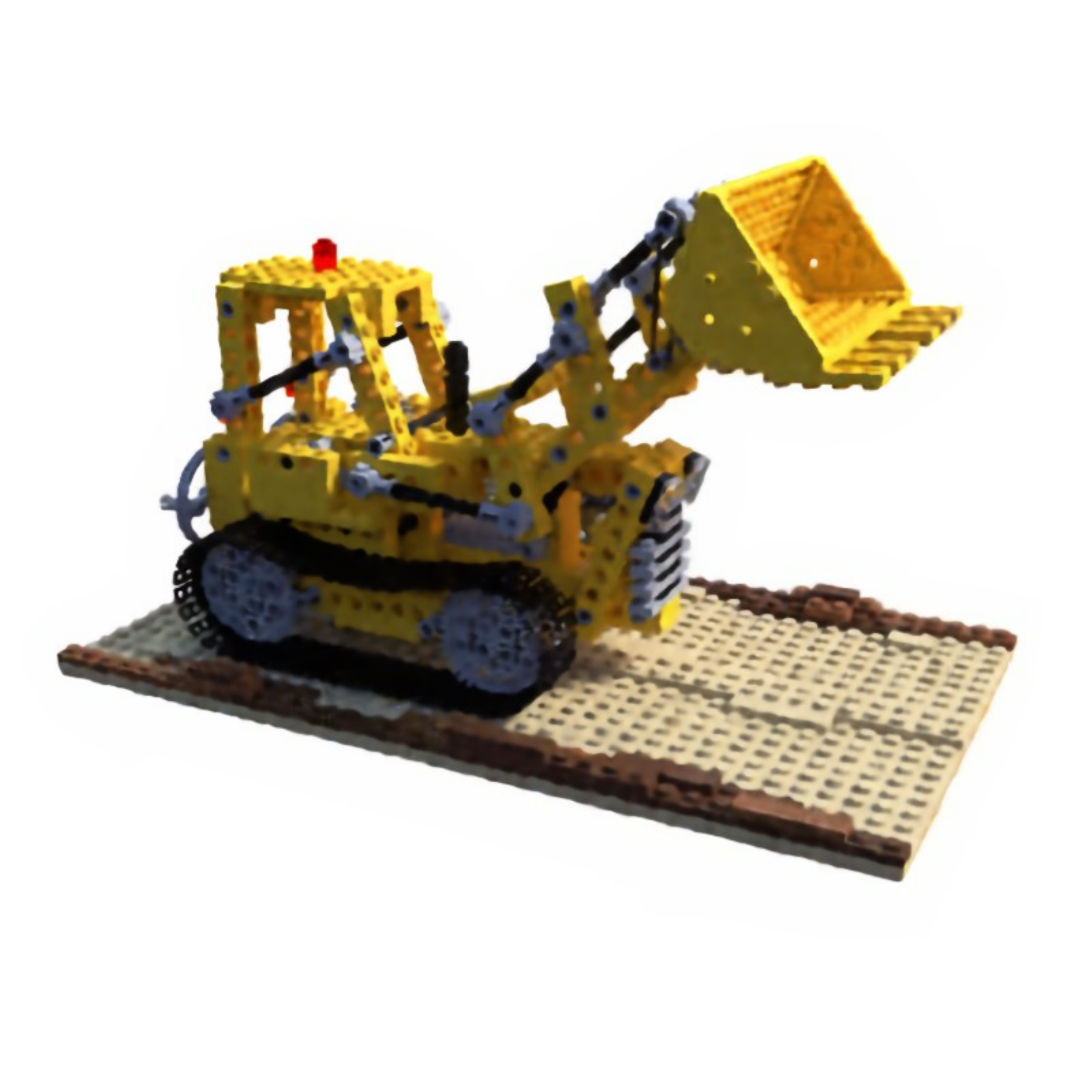}
        \caption{\footnotesize{Input Views}\label{fig_teaser_render-img}}
        \end{subfigure}
        \begin{subfigure}{.145\linewidth}
            \centering
            \includegraphics[width=\linewidth]{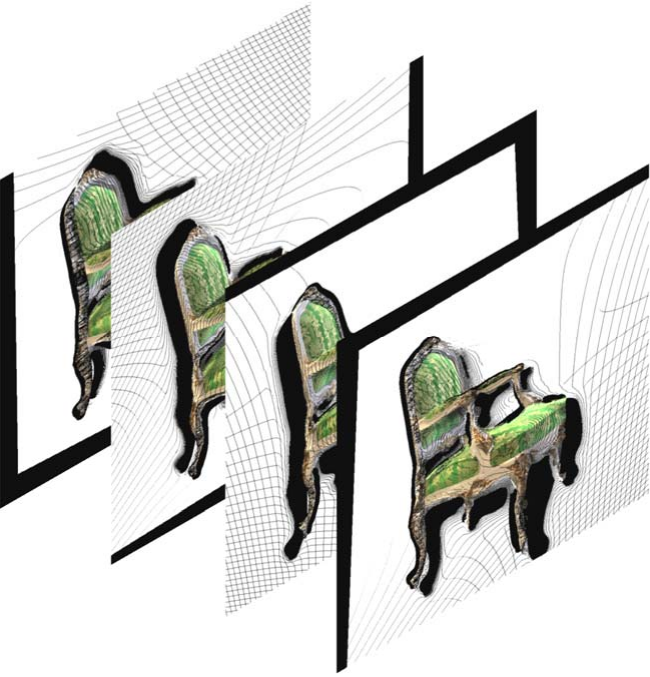}
            \includegraphics[width=\linewidth]{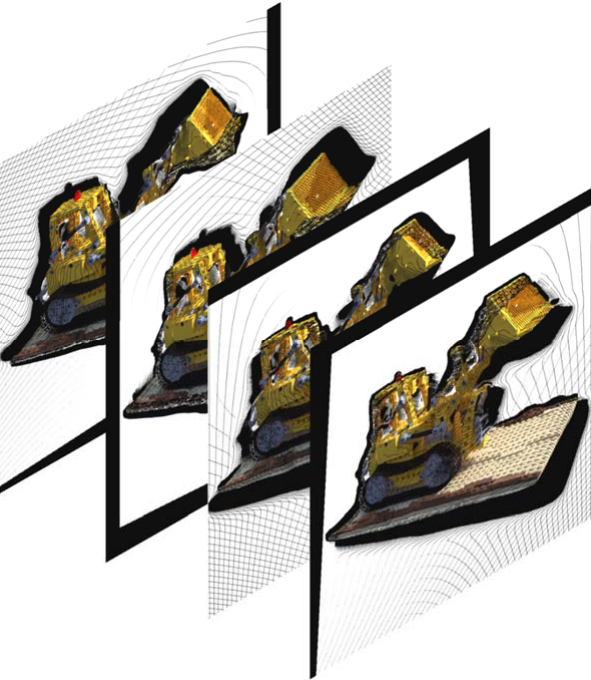}
        \caption{\footnotesize{Warped Images}\label{fig_teaser_warped-img}}
        \end{subfigure}
        \begin{subfigure}{.145\linewidth}
            \centering
            \includegraphics[width=\linewidth]{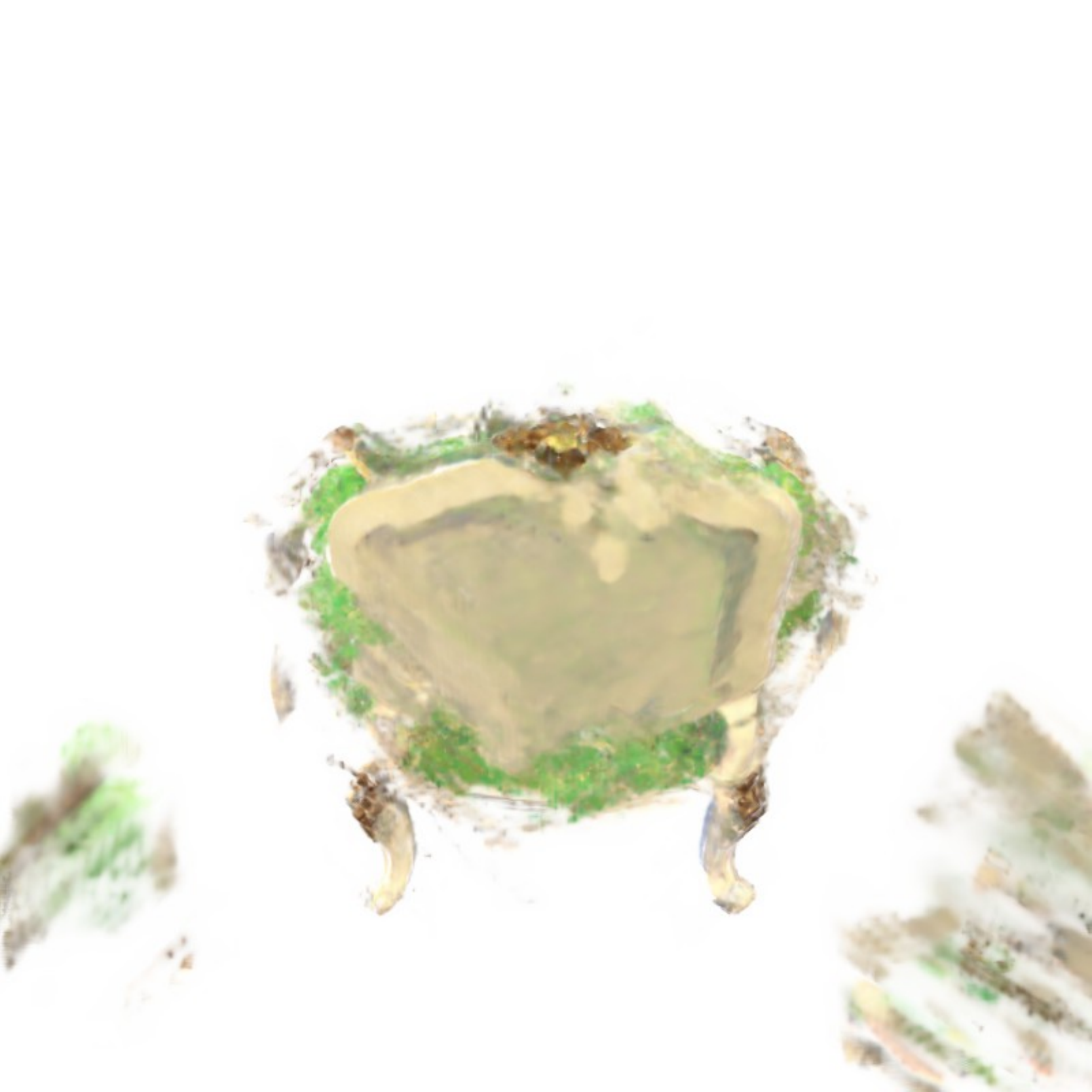}
            \includegraphics[width=\linewidth]{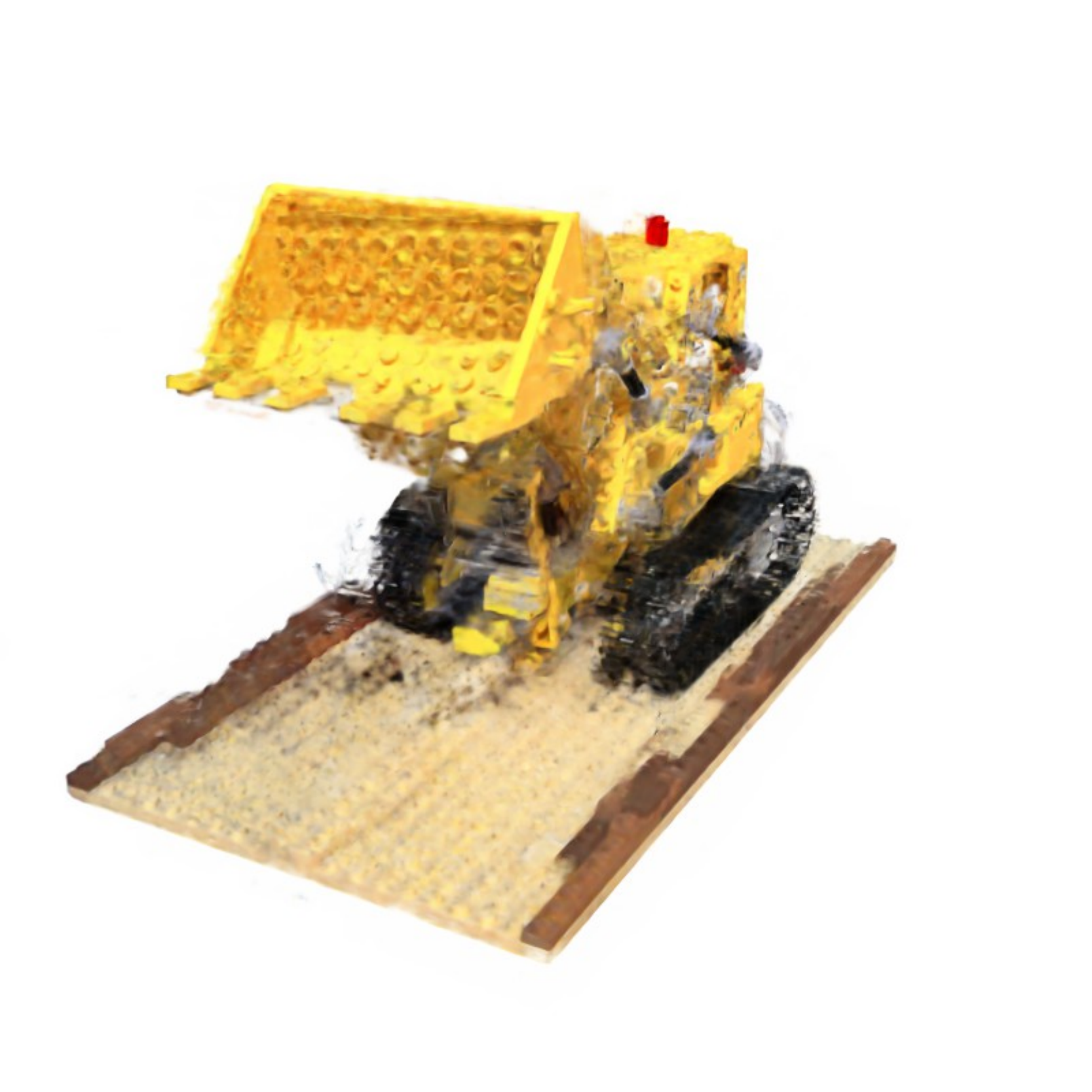}
        \caption{\footnotesize{w/o Unreliability}\label{fig_teaser_depth-img}}
        \end{subfigure}
        \begin{subfigure}{.145\linewidth}
            \centering
            \includegraphics[width=\linewidth]{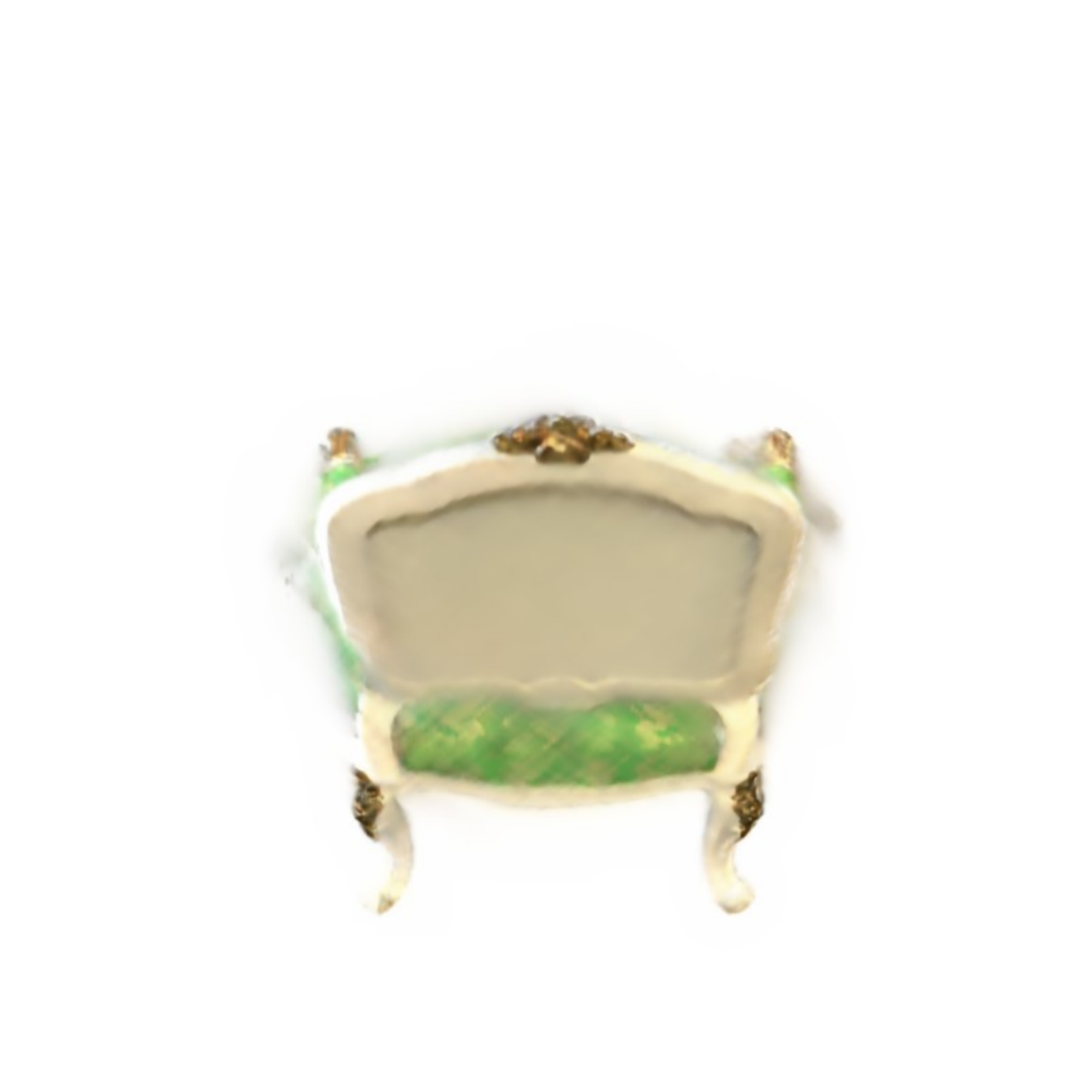}
            \includegraphics[width=\linewidth]{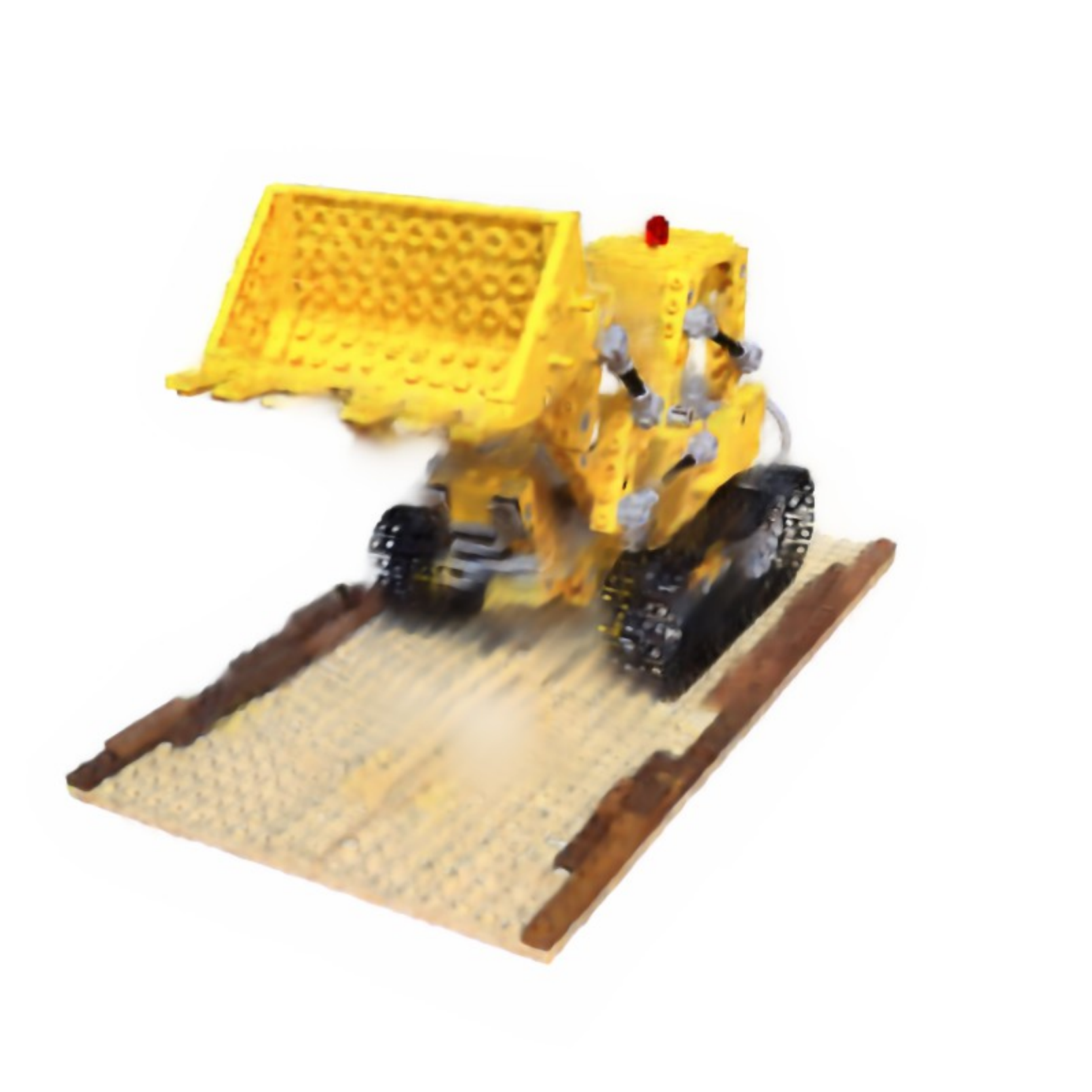}
            \caption{\footnotesize{Ours}\label{good-img}}
            \end{subfigure}
    \hspace{-3mm}
    \begin{subfigure}{.4\linewidth}
        \centering
        \includegraphics[width=\linewidth]{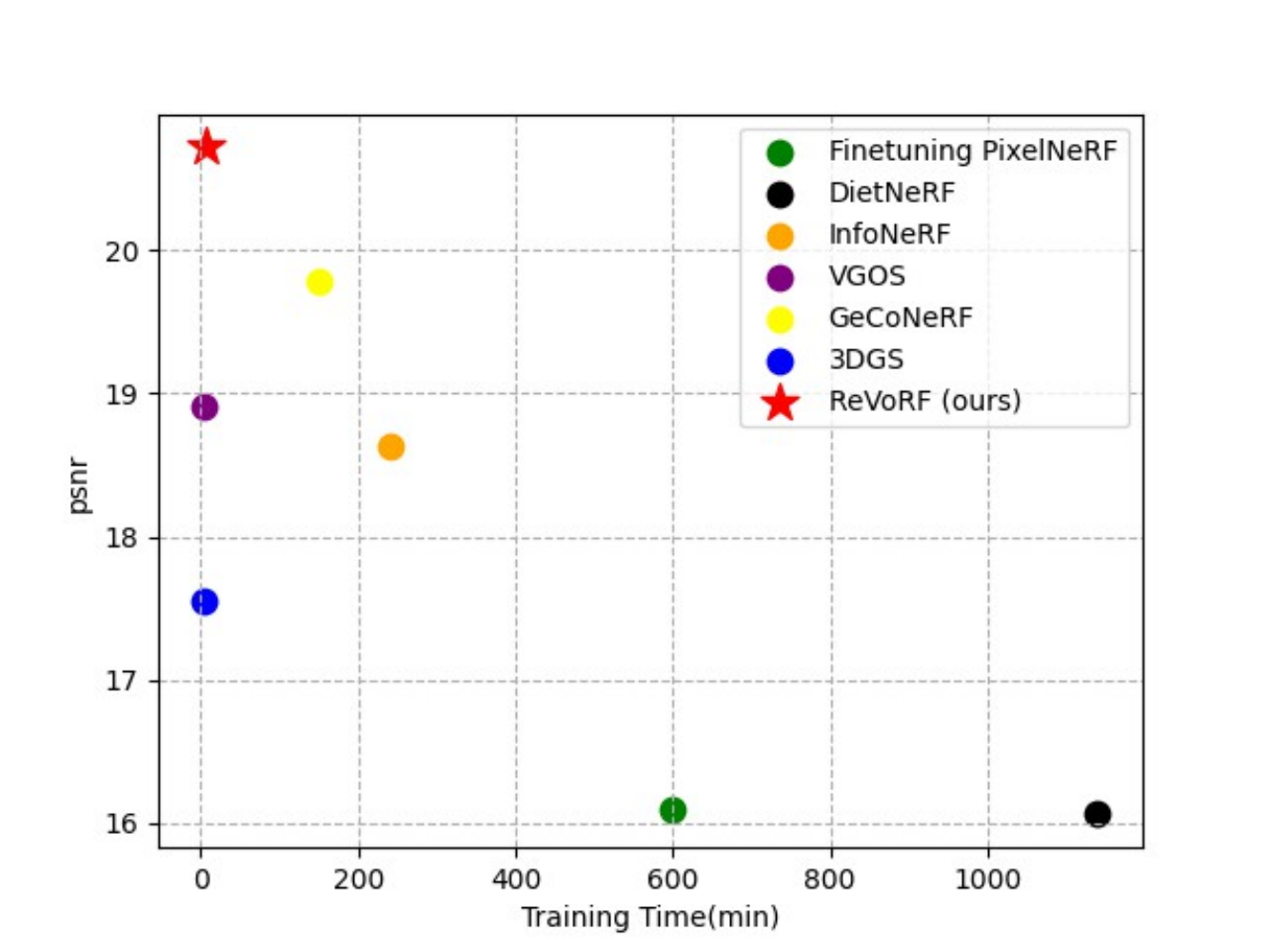}
        \caption{Comparisons on the Realistic Synthetic $360^\circ$ dataset~\cite{mildenhall2020nerf}\label{fig_teaser_d}}
    \end{subfigure}
\vspace{-3mm}
\caption{We present ReVoRF, a voxel-based framework designed to capitalize on the unreliability inherent in warped novel views. (b) demonstrates the warping outcomes, where black holes signify unmatched pixels from the original view. (c) illustrates the results of training when these holes are masked out, which unfortunately results in ambiguous geometric structures. In contrast, (d) showcases our approach's ability to maintain correct geometric consistency. ReVoRF achieves this by leveraging relational depth prior knowledge within these unreliable hole regions. Our approach demonstrates the best reconstruction quality while being one of the fastest few-shot approaches in (e).}
\label{fig:teaser}
\vspace{-5mm}
}
\maketitle

\renewcommand{\thefootnote}{\fnsymbol{footnote}}
\footnotetext[1]{The first two authors contributed equally.}
\footnotetext[2]{Corresponding author (\emph{shengfenghe@smu.edu.sg}).}

\begin{abstract}\vspace{-4mm}
    We propose a voxel-based optimization framework, ReVoRF, for few-shot radiance fields that strategically address the unreliability in pseudo novel view synthesis. Our method pivots on the insight that relative depth relationships within neighboring regions are more reliable than the absolute color values in disoccluded areas. Consequently, we devise a bilateral geometric consistency loss that carefully navigates the trade-off between color fidelity and geometric accuracy in the context of depth consistency for uncertain regions. Moreover, we present a reliability-guided learning strategy to discern and utilize the variable quality across synthesized views, complemented by a reliability-aware voxel smoothing algorithm that smoothens the transition between reliable and unreliable data patches. Our approach allows for a more nuanced use of all available data, promoting enhanced learning from regions previously considered unsuitable for high-quality reconstruction. Extensive experiments across diverse datasets reveal that our approach attains significant gains in efficiency and accuracy, delivering rendering speeds of 3 FPS, 7 mins to train a $360^\circ$ scene, and a 5\% improvement in PSNR over existing few-shot methods. Code is available at \href{https://github.com/HKCLynn/ReVoRF}{https://github.com/HKCLynn/ReVoRF}.\vspace{-5mm}
\end{abstract}

\section{Introduction}

Neural Radiance Fields (NeRF) have revolutionized the fields of novel view synthesis and 3D reconstruction by leveraging an implicit function optimized from a collection of 2D images~\cite{mildenhall2020nerf, barron2021mip, muller2022instant, hu2023tri, jiang2023diffuse3d}. Despite their remarkable rendering capabilities, NeRFs are hampered by the substantial cost and time required to gather dense image sets for a given scene~\cite{yu2021pixelnerf, tancik2022block, chen2022geoaug}. This challenge has spurred the development of Few-shot NeRF, an emerging domain focused on reconstructing 3D scenes with minimal image data~\cite{deng2022depth,chen2022geoaug,guangcong2023sparsenerf,kwak2023geconerf}.

The performance of NeRF in accurately reconstructing geometry and texture diminishes when faced with sparse observations, as it tends to overfit the limited views available~\cite{zhang2022ray,deng2022depth}. To address this issue, there has been a push in recent studies to fortify NeRF with additional priors~\cite{yu2021pixelnerf,yang2023freenerf}, including semantic relations~\cite{jain2021putting}, depth cues~\cite{zhang2022ray}, and entropy constraints~\cite{kim2022infonerf}. These enhancements strive to extract maximum information from limited data. However, the reconstructions are inherently limited by the insufficient coherence of the sparse views provided.

Recent research has explored overcoming the challenges posed by very limited observations through pseudo-view synthesis~\cite{bortolon2023vm, kwak2023geconerf, zhou2023single}. By using known camera poses and coarse depth estimates, these methods generate warped images from sparse viewpoints to enhance cross-view consistency. However, as shown in Fig.~\ref{fig_teaser_warped-img}, these generated images often include noisy areas with artifacts, which, if used for learning, can lead to inconsistent training signals and compromise scene integrity. To address this, Kwak et al.~\cite{kwak2023geconerf} implement self-occlusion aware masking to exclude unreliable regions. While this selective masking successfully filters out areas of uncertainty, it also introduces voids, presenting a conundrum: refining these images can bring in inconsistent noise and floaters, as illustrated in Fig.~\ref{fig_teaser_depth-img}, yet the limited number of usable samples necessitates using all available pseudo supervision for quality reconstruction.

In light of the issues identified with unreliable warped areas, our paper proposes a novel method for fully exploiting these uncertain regions to achieve multi-view consistency learning. The rationale behind our method is that, although absolute supervision is not reliable in those disoccluded regions, we observe that they maintain consistent relative depth relationships. The unreliability of certain regions in warped images can still bear geometric resemblances to their original view counterparts. We find that local depth information within these images can indicate geometric disparities, offering a self-supervised signal that aids in discerning the geometry of regions lacking precise textural information. While reliable regions offer more accurate supervision, our approach seeks to fully utilize all the information present, exploiting depth cues in coarsely warped images to inform the learning process across both reliable and traditionally discarded unreliable areas.

Drawing from the insights above, we propose a novel voxel-based optimization framework, ReVoRF, tailored for fast and multi-view consistent reconstruction of few-shot radiance fields, which incorporates the relative depth priors from several aspects. The objective of this work is to concurrently explore information from both dependable and less reliable regions within the warped novel view images. In the first step, we randomly warp the sparse images onto a series of novel views, subsequently delineating reliable and unreliable regions based on the pixel-wise correlation between the input and novel view. We then introduce a bilateral geometric consistency loss to enable self-training on novel synthesized images. This loss encompasses a reconstruction term in a bilateral manner, including a color and density regularization term for reliable regions and a relative depth consistency term for unreliable regions, respectively. While the former term aims at explicitly learning the geometric context of the reliable regions, the relative depth regularization is applied for implicitly exploring the geometric consistency guided by relative depth. Moreover, we integrate unreliability into our voxelization of scene features: 1) a reliability-guided learning strategy that dynamically adjusts learning priorities towards more reliable regions; 2) a reliability-aware voxel smoothing procedure that preserves structural integrity in reliable zones and mitigates inconsistencies in less reliable ones, ensuring a balanced and coherent scene reconstruction. As illustrated in Fig.~\ref{fig_teaser_d}, assisted by both bilateral geometric consistency loss and reliability-aware regularization, our method is the second fastest while achieving the best reconstruction fidelity, with a large margin over the others.

Our contributions can be summarized as follows:
\begin{itemize}
  \item We present the first attempt to explore pseudo-views unreliability within few-shot radiance fields, presenting the first framework to incorporate these areas for enhanced multi-view consistency learning with a bilateral geometric consistency loss.
  \item We introduce a reliability-guided learning strategy and voxelization smoothing procedure that tailors the learning process to the reliability of data, thus optimizing the training emphasis for improved reconstruction quality in few-shot radiance fields.
  \item We demonstrate superior performance of our approach against existing state-of-the-art few-shot methods in efficiency and accuracy, through extensive experiments on both synthetic and real-world datasets.
\end{itemize}

\section{Related Work}

\paragraph{Neural Radiance Fields (NeRF).} NeRF~\cite{mildenhall2020nerf, MartinBruallaR21, barron2021mip, chen2021mvsnerf, hu2023tri} have emerged as a significant advancement in 3D reconstruction and novel view synthesis. These methods employ an implicit function to represent a 3D scene, enabling the extraction of detailed geometric and textural information from a set of multi-view images.
Subsequent researchers have broadened the scope of NeRF applications, including generative modeling~\cite{gao2022get3d, tan2022volux, zheng2023my}, video synthesis~\cite{li2021neural,shao2023tensor4d, du2021nerflow}, and scene editing~\cite{liu2021editing, yuan2022nerf}.
Despite the impressive rendering quality, the training of vanilla NeRF often spans several days for a single scene reconstruction.
Recent advancements~\cite{sun2022direct,sun2022improved,fridovich2022plenoxels,muller2022instant} have endeavored to mitigate this computational burden. Approaches such as DVGO~\cite{sun2022direct,sun2022improved,cao2023hexplane} employ dense voxel grids in conjunction with shallow multilayer perceptrons to expedite the reconstruction process. Similarly, Plenoxels~\cite{fridovich2022plenoxels} utilizes sparse voxel grids, and Instant-NGP~\cite{muller2022instant} employs a multi-resolution hash table to delineate the radiance field more efficiently.
Diverse from these methods, which typically require dense inputs, we aim to address the challenge of achieving fast and high-fidelity scene reconstruction in the case where only a few observed views are available. 

\midparaheading{Few-Shot NeRFs.}
Recent advances~\cite{kim2022infonerf, jain2021putting, yu2021pixelnerf, chen2021mvsnerf, zhang2022ray, zheng2024learning} have sought to reduce the dependency on densely collected data for scene reconstruction, leveraging sparse inputs and scene priors.
Notably, PixelNeRF~\cite{yu2021pixelnerf} and StereoRF~\cite{chibane2021stereo} utilize local semantic relationships across multiple scenes, while MVSNeRF~\cite{chen2021mvsnerf} incorporates cost volume to enhance performance. These methodologies, however, require pretraining on numerous scenes to acquire necessary scene priors.
Further developments~\cite{jain2021putting, kim2022infonerf, radford2021learning, ijcai2023p157, yang2023freenerf} have introduced various regularization techniques to maximize the utility of sparse input views.
InfoNeRF~\cite{kim2022infonerf} enhances ray adjacency consistency through entropy regularization. DietNeRF~\cite{jain2021putting} facilitates cross-view semantic consistency by harnessing the semantic space of the pretrained CLIP~\cite{radford2021learning}, while DiffusioNeRF~\cite{wynn-2023-diffusionerf} explores the diffusion prior of pretrained diffusion models. Additionally, FreeNeRF~\cite{yang2023freenerf} applies frequency regularization, and VGOS~\cite{ijcai2023p157} introduces voxel regularization to optimize both feature representation and density.
Another group of research focuses on augmenting sparse inputs with synthetically generated views.
RapNeRF~\cite{zhang2022ray} utilizes geometric re-projection for novel view extrapolation, while VmNeRF~\cite{bortolon2023vm} employs depth maps for view-morphing. GeCoNeRF~\cite{kwak2023geconerf} aims to refine geometric consistency by separating reliable regions from warped images and discarding unreliable areas prone to self-occlusion.
Our work diverges from these approaches by considering the inherent information of both reliable and unreliable regions of the novel view images, facilitating cross-view geometric consistency. 

\midparaheading{Unreliability/Uncertainty Modeling.} In the rapid development of NeRF, the incorporation of uncertainty modeling has become crucial for achieving robustness in 3D reconstruction from sparse views. Previous efforts have employed diverse strategies, including Bayesian approaches~\cite{Neal95,KendallG17} and evidential neural networks~\cite{SensoyKK18,AminiSSR20}, to quantify uncertainty in neural networks. In the context of NeRF, uncertainty has been harnessed to enhance rendering and guide input capture. Some methods assume Gaussian noise in RGB space for pixel-wise uncertainty~\cite{MartinBruallaR21,PanLSH22}, employ volumetric entropy for scene geometry~\cite{LeeCWLKY22,YanLQCF23}, or adopt variational inference or Latent Variable Modeling for radiance field uncertainty as seen in S-NeRF~\cite{ShenRAM21} and CF-NeRF~\cite{ShenAMR22}. However, these approaches have not comprehensively addressed uncertainty quantification in unseen regions. Our approach differs from these methods by not only capturing uncertainty in the geometry and appearance of visible areas but also explicitly accounting for unseen spaces, including occluded points, which previous methods have not considered. This distinction allows for a more nuanced and accurate reconstruction, promising to elevate the fidelity of few-shot NeRF models.

\section{Methodology}

\begin{figure*}[t]
    \centering
    \includegraphics[width=\linewidth]{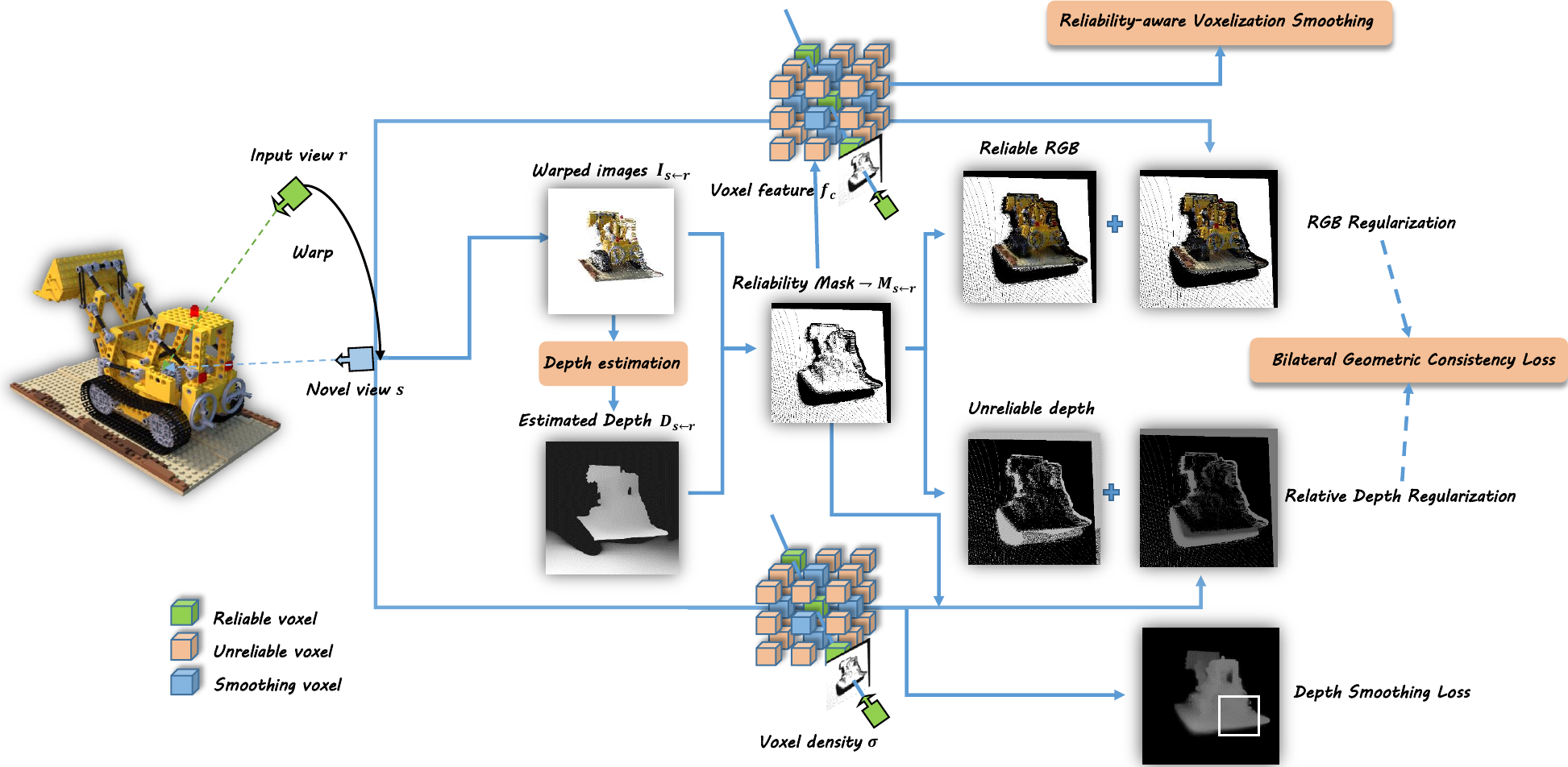}
    \vspace{-5mm}
    \caption{Overview of our proposed ReVoRF. Specifically, we first warp the sparse images onto several novel views and determine both the dependable and unreliable regions. Based on the dependability of each image region, we introduce a bilateral geometric consistency loss for multi-view consistent learning, which is composed of a color and density regularization term for reliable regions and a relative depth consistency term for unreliable regions. These two terms are responsible for explicitly learning the reliable geometric contents and implicitly exploring the geometric consistency via the guidance of relative depth, respectively. For voxel feature regularization, we integrate the unreliability through a reliability-guided learning strategy and a reliability-aware voxel smoothing procedure. By prioritizing the learning of more reliable regions and mitigating the inconsistencies in less reliable ones, ReVoRF ensures a more balanced and coherent reconstruction. \label{fig_pipeline}}
    \vspace{-3mm}
\end{figure*}  

\subsection{Preliminaries}
NeRF~\cite{mildenhall2020nerf} represents a scene as a continuously differentiable function $\textit{f}$ via a Multi-Layer Perceptron (MLP). Given a 3D position $ \mathbf{x} \in \mathbb{R}^3 $ and the associated 2D viewing directions $ \mathbf{d} \in \mathbb{R}^2 $, NeRF maps them into a volume density $\sigma \in \mathbb{R}$ and an RGB value $ \mathbf{c} \in \mathbb{R}^3 $, such that: $(\mathbf{c}, \sigma) = f (\gamma(\mathbf{x}), \gamma(\mathbf{d})),$
where the $\gamma$ is a positional encoding that projects $\mathbf{x}$ and $ \mathbf{d}$ into a higher dimensional feature space~\cite{tancik2020fourier}. With a ray parameterized as $\mathbf{r}_p(\mathit{t}) = \mathbf{o} + \mathit{t}\mathbf{d}_p$ cast from the camera's optical center $\mathbf{o}$ along direction $\mathbf{d}_p$, the expected color $\hat{C}(\mathbf{r}_p)$ of pixel $p$ is rendered as follows: 
\begin{equation}
    \hat{C}(\mathbf{r}_p) = \int_{t_n}^{t_f} T(t) \sigma(\mathbf{r}_p(t)) \mathbf{c}(\mathbf{r}_p(t), \mathbf{d}_p) \, dt,
\end{equation}
where $t_n$ and $t_f$ are the near and far bounds of the ray for sampling, and $T(t) = \exp\left(-\int_{t_n}^{t} \sigma(\mathbf{r}(s)) \, ds\right)$ denotes the cumulative transparency along the ray from $t_n$ to $t$.
Therefore, the NeRF can be optimized by a reconstruction loss between the rendered color $\hat{C}(\mathbf{r})$ and the real color $C(\mathbf{r})$:
\begin{equation}
    L_{rgb} = \sum_{\mathbf{r} \in \mathcal{R}} \| \hat{C}(\mathbf{r}) - C(\mathbf{r}) \|^2, 
\end{equation}
where $\mathcal{R}$ denotes the set of training rays.

\subsection{Overview}
We propose ReVoRF, a novel voxel-based optimization framework tailored for fast and multi-view consistent scene reconstruction from sparse input views. Our key idea is to incorporate the unreliability information for fully exploring the warped novel view images. 
By treating depth priors as the unreliability metrics, ReVoRF facilitates the reconstruction of the few-shot radiance field from several aspects, including the multi-view consistency learning~(Sec.~\ref{sec:multi-view-consistency}) and the regularization of voxel features~(Sec.~\ref{sec:voxel-regularization}). 
The overall pipeline of ReVoRF is displayed in Fig.~\ref{fig_pipeline}.

\subsection{Unreliability for Multi-view Consistency} \label{sec:multi-view-consistency}

In this section, we explore the potential of unreliability in facilitating multi-view geometric consistency via the proposed bilateral geometric consistency loss.

\midparaheading{Novel View Warping.}
Starting from a set of sparse input images $\mathit{I}^i_r$ for $i \in {1,...,N_r}$, where $i$ represents the view number and $N_r$ is a small number, \eg, $N_r=3$ or $N_r=4$, we propose to synthesize novel view images $\mathit{I}^{i,j}_{s \leftarrow r}$ through a fast and flexible warping process on several novel views $j \in {1,...,N_s}$. To preserve the cross-view consistency, the warping is guided by a coarse depth map $\mathit{D}^i_{r}$, where the depth value on each pixel $p$ is accumulated by the density along each ray $r_p$ omitted from the camera: $\mathit{D}^i_{r}(p) = \int_{t_f}^{t_n} T(t) \sigma(r_p(t)) t \, dt$. Subsequently, we obtain the warped image $\mathit{I}^{i,j}_{s \leftarrow r}$ through a cross-view transformation $H_{s \leftarrow r}$, which deforms each pixel $p_r$ of $\mathit{I}^i_r$ to its corresponding position $p_s$ on the target view:
\begin{equation}
    \begin{split}
    p_{s \leftarrow r} & = H_{s \leftarrow r}(p_r) \\
                    & = f_{s\leftarrow w}(f_{w\leftarrow r}(p_r)), 
    \end{split}
\end{equation}
$f_{w\leftarrow r}$ is a mapping matrix from the pixel coordinate of $\mathit{I}^i_{r}$ to the world coordinate and $f_{s\leftarrow w}$ is the inverse operation to the coordinate of $\mathit{I}^{i,j}_{s \leftarrow r}$, such that:
\begin{align}
f_{w\leftarrow r}(p_r) &= D^i_r(p_r) T^{-1}_rK^{-1}_r p_r,\label{eq:forward} \\
f_{s\leftarrow w}(p_w) &= K_sT_s(p_w),  
\end{align}
where $K$ and $T$ represent the camera's intrinsic and extrinsic parameter matrices in their corresponding views, respectively. Since the warping function is not surjective, voids may occur in $\mathit{I}^{i,j}_{s \leftarrow r}$. We empirically obtain a binary mask $M_{warp}$, where the pixels of void areas are set as 1 to identify the initial unreliable regions. To further refine the mask, we employ the cross-view pixel correspondences within the world coordinate of $\mathit{I}^i_{r}$ and $\mathit{I}^{i,j}_{s\leftarrow r}$. To achieve this, we obtain the pseudo depth $\mathit{D}^j_{s}$ of view $j$ by rendering from the radiance fields. Following Eq.~\ref{eq:forward}, we map the pixel of $\mathit{I}^i_{r}$ and $\mathit{I}^{i,j}_{s\leftarrow r}$ into the same coordinate and obtain the correlation map $M_{cor}$ by comparing the distance between each pixel pair $(p_r, p_{s\leftarrow r})$: 
\begin{equation}
    M_{cor}(p_{s\leftarrow r}) = \begin{cases}
        1, &\|f_{w\leftarrow r}(p_r)-f_{w\leftarrow s}(p_{s \leftarrow r})\|_2>\mathbf{\epsilon}. \\
        0, &\text{otherwise}.
    \end{cases}
\end{equation}
In this way, the final unreliability mask can be calculated as follows:
\begin{equation}
    M_{s\leftarrow r} = M_{cor} \cup M_{warp}.
\end{equation}

\midparaheading{Bilateral Geometric Consistency Loss.} According to the obtained unreliability mask $M_{s\leftarrow r}$, we categorize the warped novel view image into reliable regions $R_{rel}$ and unreliable regions $R_{unr}$. Subsequently, we propose a bilateral geometric consistency loss to facilitate the self-training. Since the contents within $R_{rel}$ are considered reliable, we explicitly constrain the appearance of rendered image $\mathit{I}^{j}_{s}$ on view $j$ via a reconstruction loss defined as:
\begin{equation}
    L_{rel} = \sum_{p \in \mathcal{R}_{rel}} \| \mathit{I}^{i,j}_{s\leftarrow r}(p) - \mathit{I}^{j}_{s}(p) \|^2. \label{eq:rgb}
\end{equation}

For $R_{unr}$, we propose to leverage the relative depth prior of the warped $\mathit{I}^{i,j}_{s\leftarrow r}$ to improve the geometric consistency. Specifically, we extract the depth map $\mathit{D}_{s\leftarrow r}$ with a powerful pretrained depth estimation model DPT~\cite{Ranftl2020}. By analyzing the semantics of the surrounding context, we could inpaint the voids occurring in $\mathit{D}_{s\leftarrow r}$. Subsequently, a relative depth regularization loss~\cite{mertan2020siralama} is introduced to constrain the geometric consistency on the rendered depth $\mathit{D}_{s}$, which is defined as:
\begin{equation}
    \begin{split}
        L_{unr} =\sum_{p \in R_{unr}}\sum_{\hat{p} \in {N}(D^{p}_{s\leftarrow r})} \unrfunc({p}, \hat{p}),
    \end{split}\label{eq:lossUnr}
\end{equation}
where 
\begin{equation}
    \begin{aligned}
    &~\unrfunc({p}, \hat{p}) =\\
    &
    \begin{cases}
    \max(|D^{\hat{p}}_s - D^{p}_s | - m, 0) & \text{if}~(D^{\hat{p}}_{s\leftarrow r} - D^{p}_{s \leftarrow r})  \\
    & ~~~\times (D^{\hat{p}}_s - D^{p}_s) < 0,\\
    0 & \text{otherwise}.
    \end{cases}
    \end{aligned}
    \label{eq:UnrErrTerm}
\end{equation}
Here $p$ denotes any pixel within the unreliable region $R_{unr}$; $\hat{p}$ represents each pixel within a neighborhood ${N}(D^{p}_{s\leftarrow r})$ of $p$, where ${N}(D^{p}_{s\leftarrow r})$ is obtained by calculating pixels that have close depth values with $\mathit{D}^p_{s\leftarrow r}$ in the warped depth map; $D^{p}_s, D^{\hat{p}}_s$ and $D^{p}_{s \leftarrow r}, D^{\hat{p}}_{s \leftarrow r}$ denote the depth values of $p$ and $\hat{p}$ within the rendered depth map $D_s$ and estimated depth map $D_{s \leftarrow r}$, respectively. The function $\unrfunc({p}, \hat{p})$ penalizes inconsistent relative ordering between the depth values of  $p$ and $\hat{p}$ in the two depth maps. Specifically, if $D^{p}_s < D^{\hat{p}}_s$ but $D^{p}_{s \leftarrow r} > D^{\hat{p}}_{s \leftarrow r}$, or $D^{p}_s > D^{\hat{p}}_s$ but $D^{p}_{s \leftarrow r} < D^{\hat{p}}_{s \leftarrow r}$, then $\unrfunc({p}, \hat{p})$ penalizes the depth difference $|D^{\hat{p}}_s - D^{p}_s|$ beyond a threshold $m$, to prevent the depth values from shifting dramatically.

Our bilateral geometric consistency loss is then defined as a weighted sum of $L_{rel}$ and $L_{unr}$:
\begin{equation}
    L_{bgc} = \lambda_{rel} L_{rel} + \lambda_{unr} L_{unr}. \label{eq:bgc}
\end{equation}
This loss enables us to thoroughly explore the information from both reliable and unreliable regions, facilitating the learning of cross-view consistency.
Note that we also apply Eq.~\ref{eq:lossUnr} on input views $\mathit{I}_r$ as a depth regularization between the rendered depth and $\mathit{D}_{r}$ for fully exploring the depth prior. 

\begin{figure*}
    \centering
    \begin{subfigure}{.19\linewidth}
    \centering
            \includegraphics[width=\linewidth]{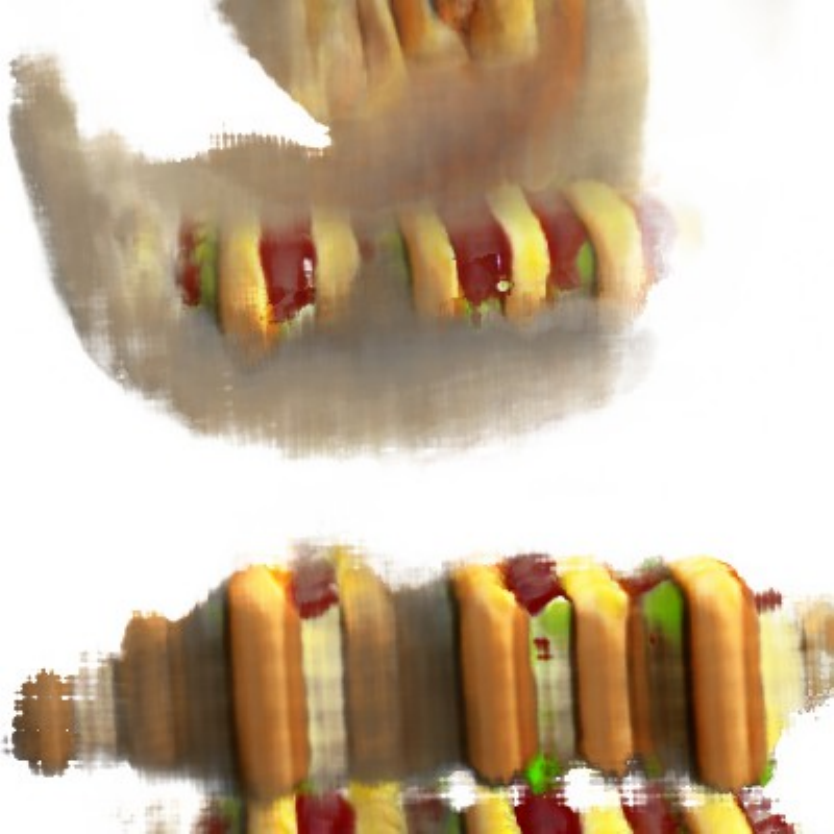} \\
            \includegraphics[width=\linewidth]{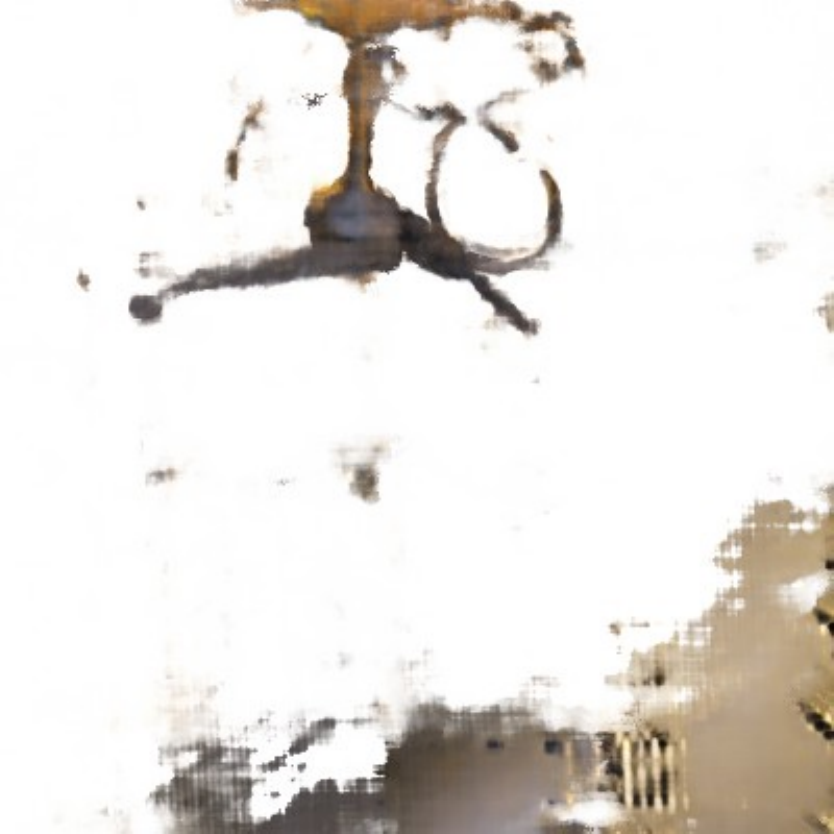} \\
        \caption*{\footnotesize{Diet-NeRF~\cite{jain2021putting}}}
    \end{subfigure}
    \hspace{-1.7mm}
    \begin{subfigure}{.19\linewidth}
    \centering
            \includegraphics[width=\linewidth]{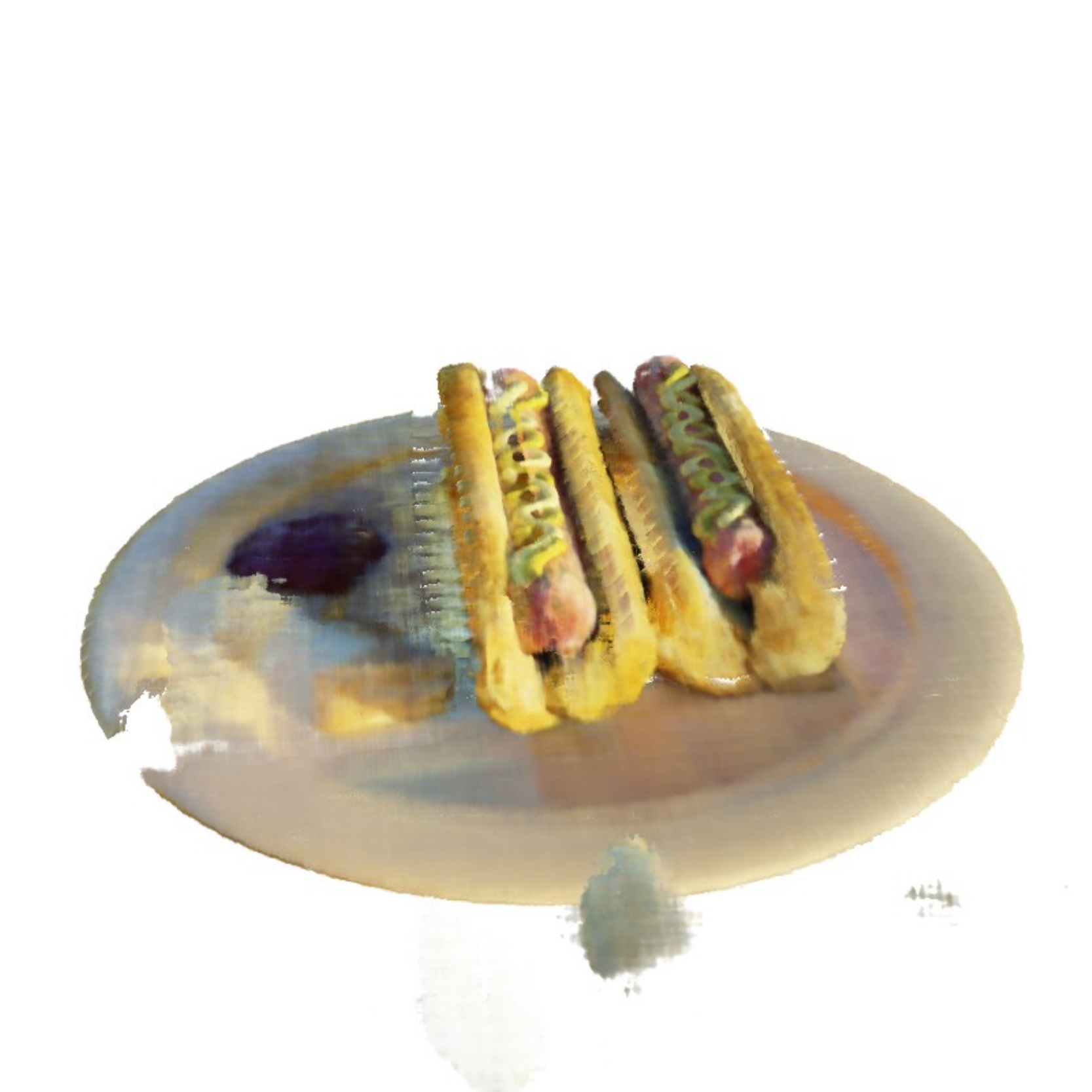}\\
            \includegraphics[width=\linewidth]{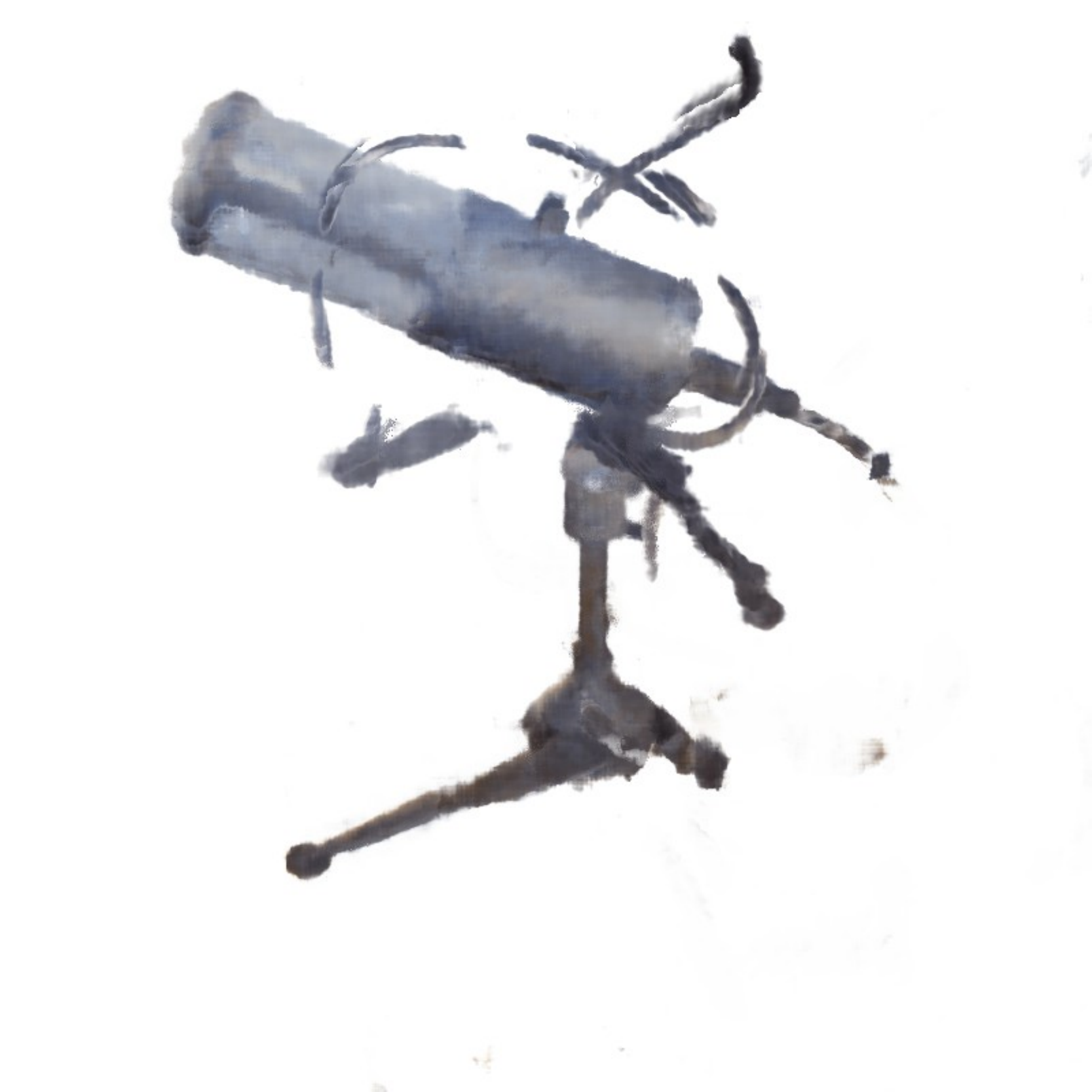}\\
        \caption*{\footnotesize{InfoNeRF~\cite{kim2022infonerf}}\label{edge_a}}
    \end{subfigure}
    \hspace{-1.7mm}
    \begin{subfigure}{.19\linewidth}
    \centering
            \includegraphics[width=\linewidth]{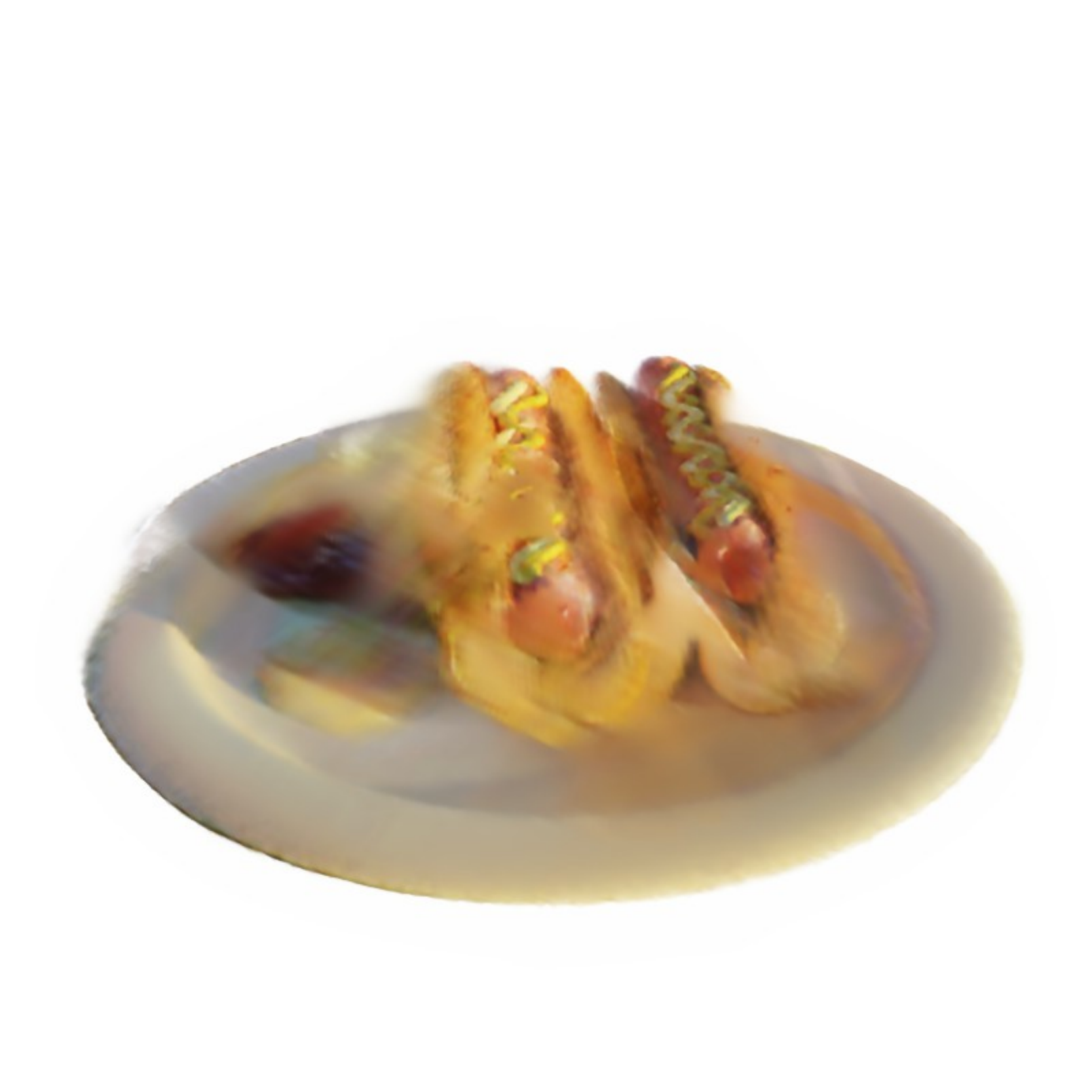}\\
            \includegraphics[width=\linewidth]{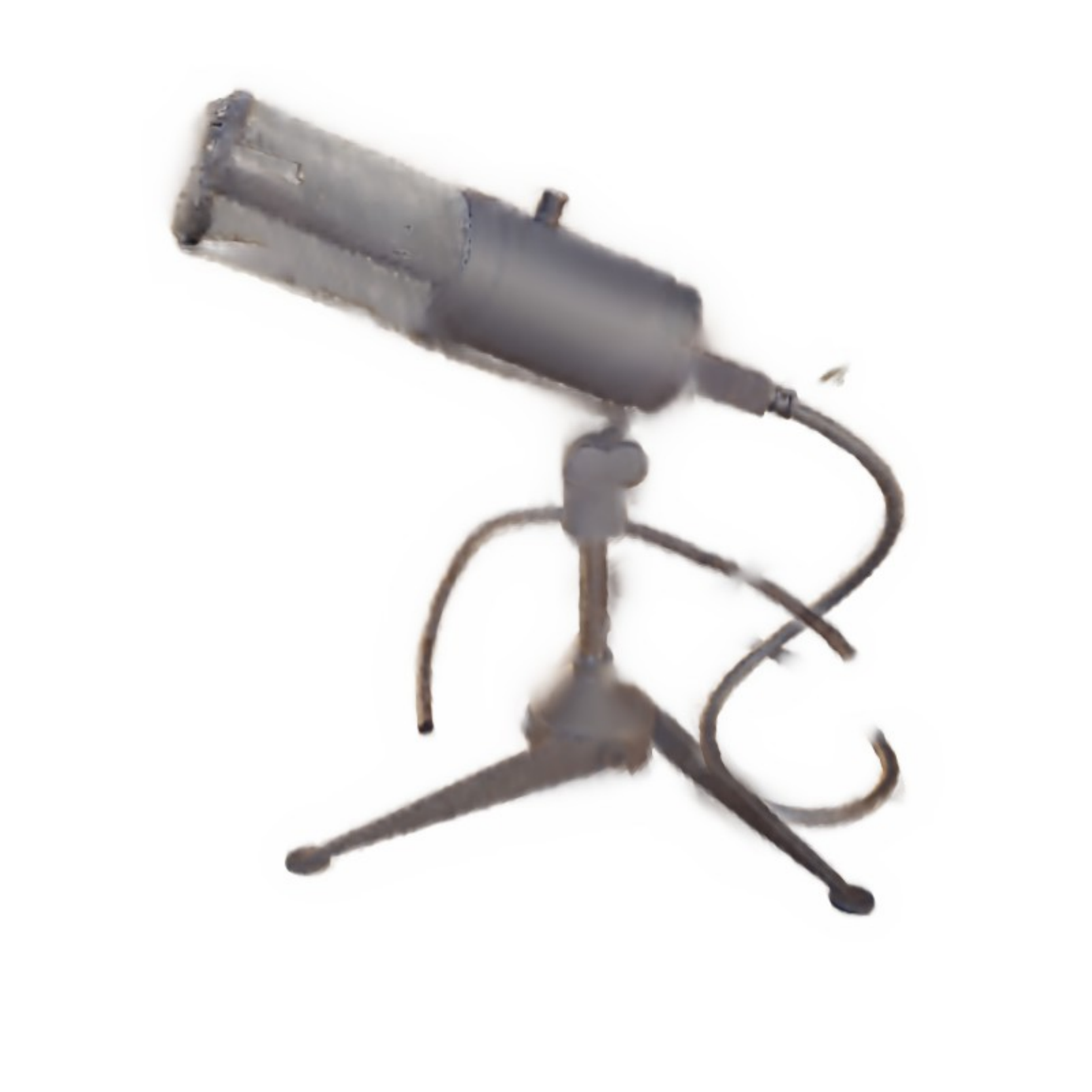}\\
        \caption*{\footnotesize{VGOS~\cite{ijcai2023p157}}\label{edge_a}}
    \end{subfigure}
    \hspace{-1.7mm}
    \begin{subfigure}{.19\linewidth}
        \centering
        \includegraphics[width=\linewidth]{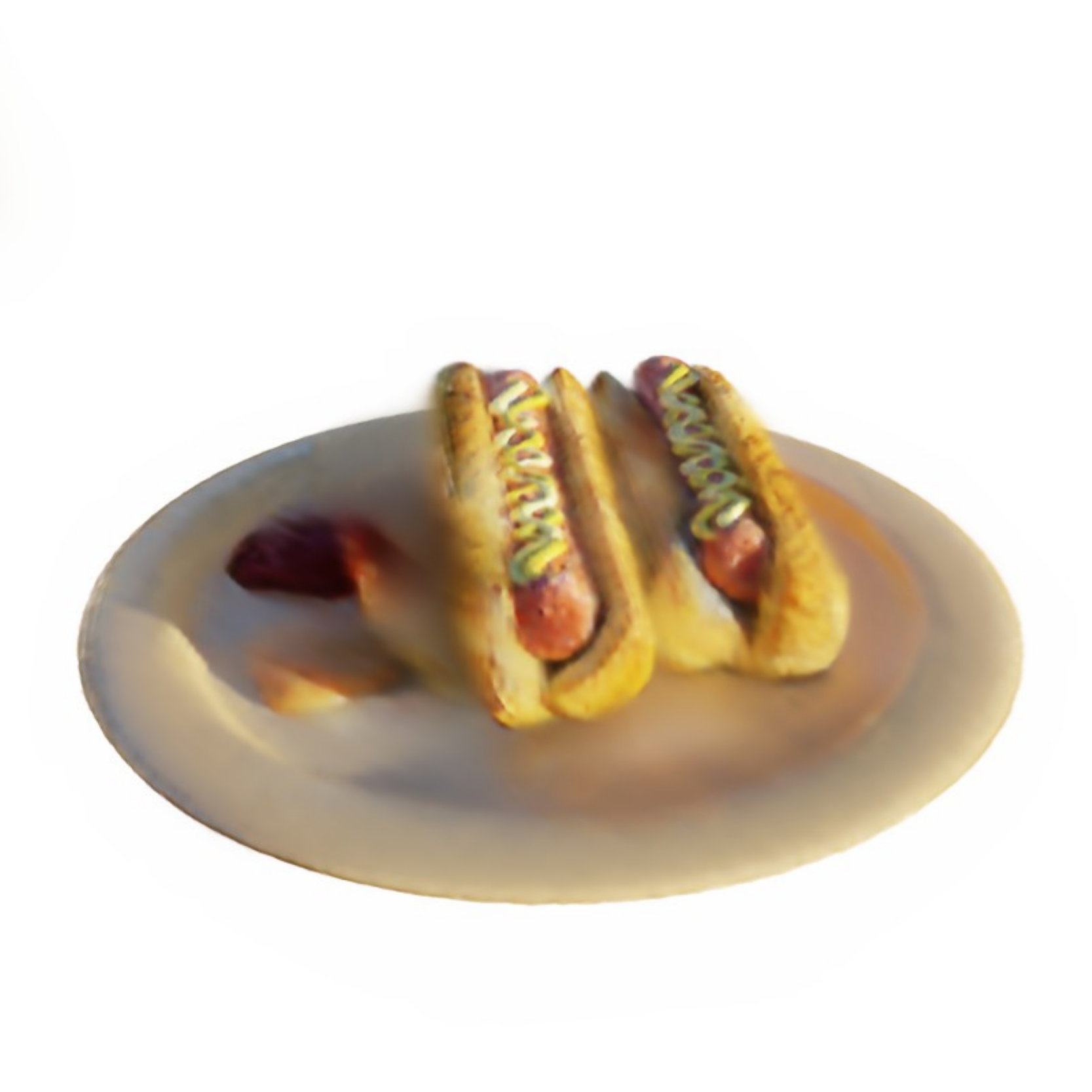}\\
        \includegraphics[width=\linewidth]{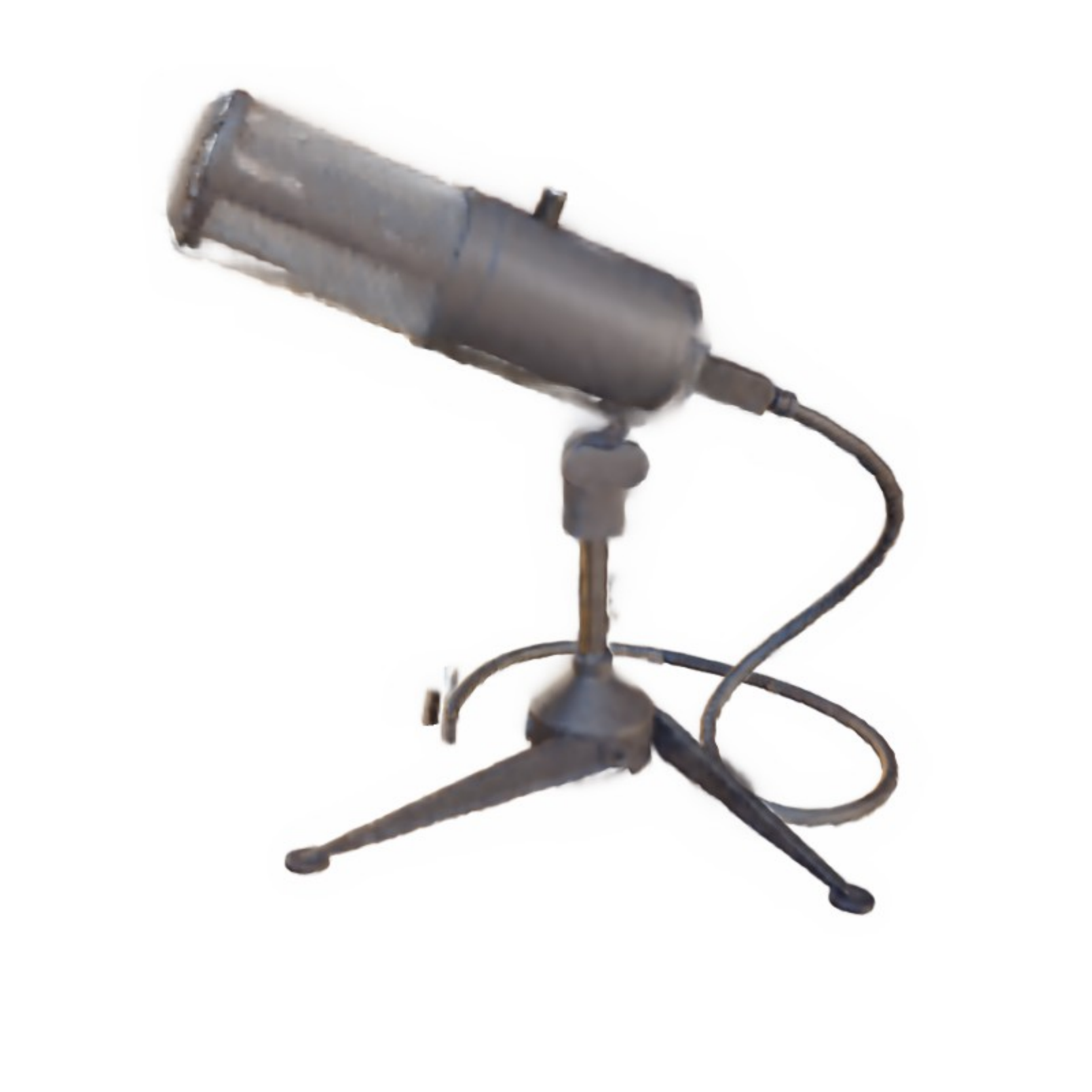}\\
        \caption*{\footnotesize{ReVoRF (Ours)}\label{edge_b}}
    \end{subfigure}
    \hspace{-1.7mm}
    \begin{subfigure}{.19\linewidth}
    \centering
    \includegraphics[width=\linewidth]{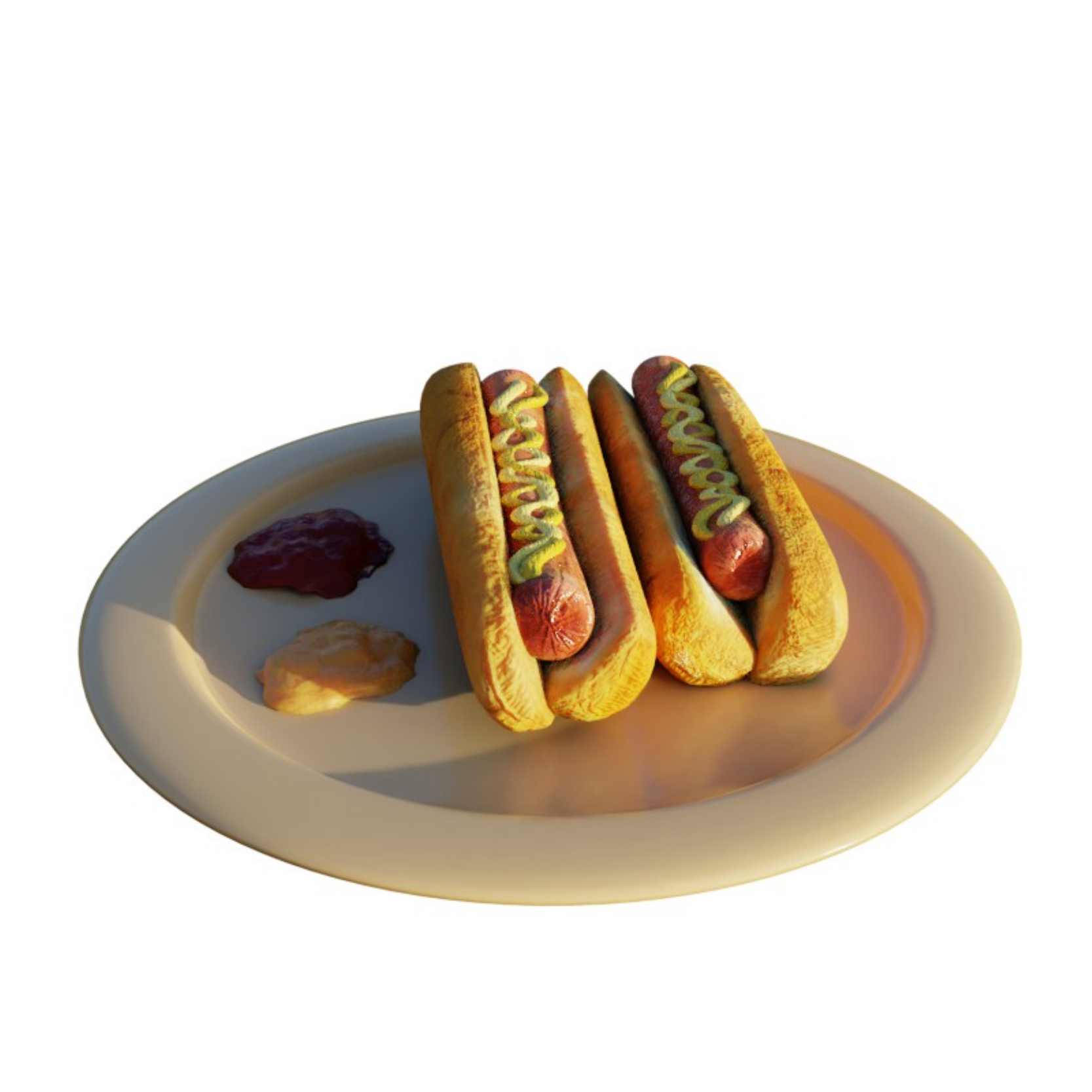}\\
    \includegraphics[width=\linewidth]{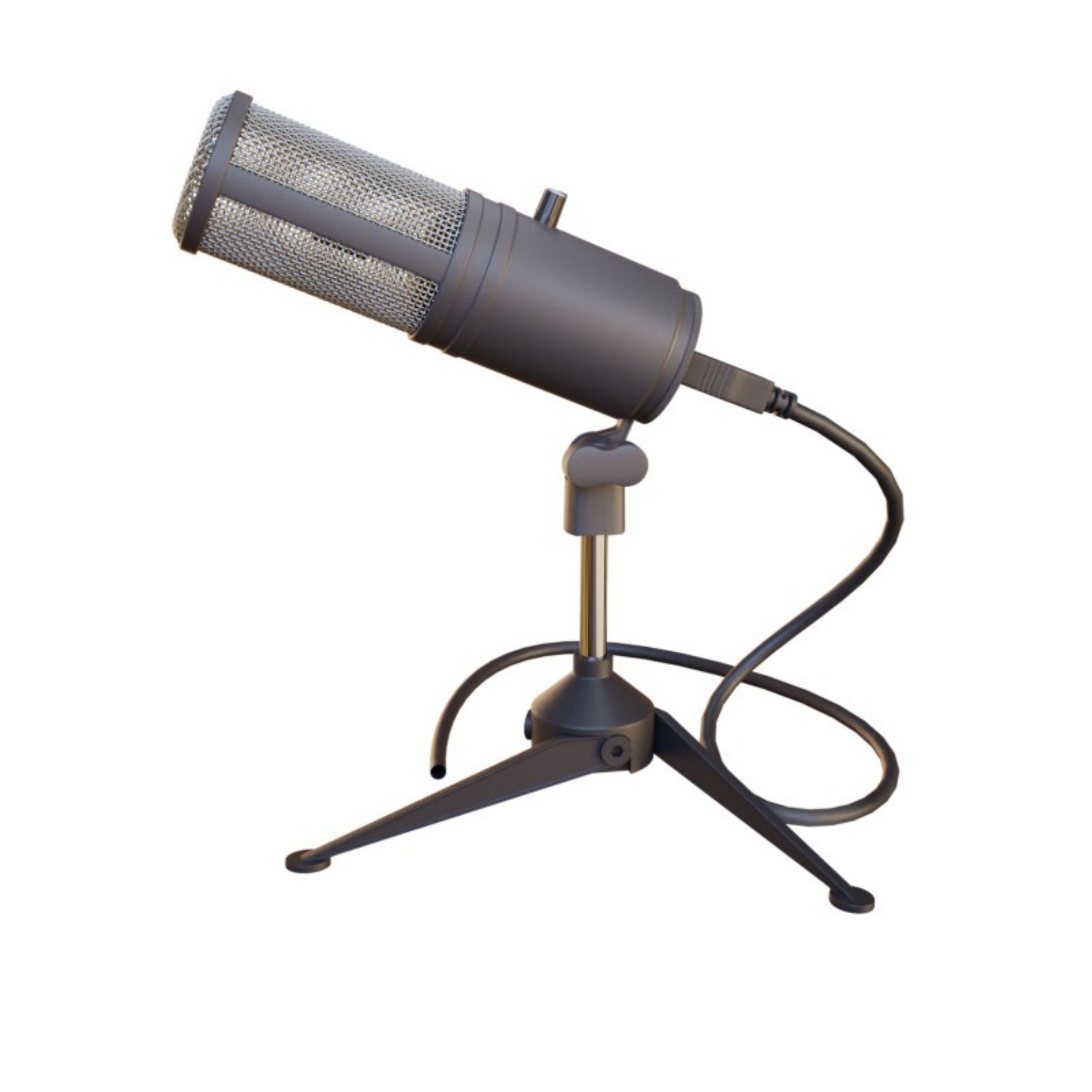}\\
        \caption*{\footnotesize{GT}\label{edge_ours}}
    \end{subfigure}
    \vspace{-3mm}
    \caption{4-views reconstructions on Realistic Synthetic 360°~\cite{MildenhallSTBRN20}. ReVoRF enables more consistent reconstruction with detailed appearance.}
    \vspace{-4mm}
    \label{fig:comparisons}

\end{figure*}

\subsection{Unreliability for Voxel Feature Regularization}\label{sec:voxel-regularization}

In this section, we incorporate the unreliability on regularizing the feature of each position of the voxel grid. With a proper design of reliability-aware voxel smoothing and reliability-aware learning adjustment, we further improve the quality of the rendered image, avoiding suboptimal scene reconstruction.

\midparaheading{Reliability-aware Voxel Smoothing.}
Employing voxelized feature representations~\cite{sun2022direct,sun2022improved} can significantly improve the training and rendering speed of NeRF, by storing the RGB features $f_c$ and density $\sigma$ in a voxel grids. 
To facilitate the learning of voxel representation, DVGO~\cite{sun2022direct,sun2022improved} propose a differentiable voxel smoothing loss, which regularizes the difference of $f_c$ and $\sigma$ between a given voxel $\mathbf{v}$ with its six adjacent points $V$ as follows:
\begin{equation}
    L(\mathbf{v}) = \sum_{\hat{\mathbf{v}} \in V} \Delta_{f_c}(\mathbf{v}, \hat{\mathbf{v}}) + \Delta_{\sigma}(\mathbf{v}, \hat{\mathbf{v}}),
\end{equation}
where $\Delta(\cdot,\cdot)$ denotes an error metric for the difference  (\eg, $L_1$, $L_2$, or Huber loss). 

However, under the few-shot scenario, learning on the sparse input views can easily overfit the view-specific image, which could lead to a degenerated voxel grid that may contain fluctuant features. To address this issue, we propose to regularize the voxel features for balanced and smooth learning. By taking the unreliability of each synthesized novel view image into consideration, we mitigate the influence of unreliable regions while promoting learning in more reliable areas.  
We design a reliability-aware smooth factor $\rho(\mathbf{v})$ for each voxel in the grid. Specifically, given a warped image, we cast ${O}$ rays $\mathbf{r}$ through each pixel of the reliable regions $R_{rel}$. For each voxel $\mathbf{v}$, we obtain a reliability score by accumulating the number of rays that pass this voxel, denoted by $S(\mathbf{v})$. The maximum number of times being passed by rays is denoted as $S(\mathbf{v})_{max}$. Then the reliability-aware smooth factor is  defined as $\rho(v) = \frac{S(\mathbf{v})}{S(\mathbf{v})_{max}}$. Finally, we formulate the reliability-aware voxel smoothing losses on $f_c$ and $\sigma$ as:
\begin{equation}
    \begin{split}
    L_{f_c} &= \sum_{\mathbf{v}}\sum_{\hat{\mathbf{v}} \in V} (1+e^{-\rho(\mathbf{v})})\Delta_{f_c}(\mathbf{v}, \hat{\mathbf{v}}), \\
    L_{\sigma} &= \sum_{\mathbf{v}}\sum_{\hat{\mathbf{v}} \in V} (1+e^{-\rho(\mathbf{v})})\Delta_{\sigma}(\mathbf{v}, \hat{\mathbf{v}}).
    \end{split} \label{eq:rho}
\end{equation}
The final regularization loss is defined as follows:
\begin{equation}
    L_{rs} = \lambda_{f} L_{f_c} + \lambda_{d} L_{\sigma}. \label{eq:rs}
\end{equation}
In this case, the unreliable regions will have smoother supervision during training, mitigating the inconsistency caused by potential overfitting. 

\midparaheading{Reliability-guided Learning Adjustment.} 
To further facilitate the learning of geometric and appearance information, we apply a reliability-guided learning strategy to dynamically prioritize the learning towards the reliable zones, while eliminating the false supervision of unreliable regions at the beginning of training. Concretely, we adjust the importance of each voxel $\mathbf{v}$ with a reliability weight $\mathbf{w_{v}} = 1+\rho(\mathbf{v})$, to control the gradients of each voxel during the back-propagation.

\subsection{Optimization of ReVoRF}

To avoid dramatic variation of the unreliable depth, we further adopt a depth smoothness loss function~\cite{Niemeyer2021Regnerf} as an extra regularization for better relative depth supervision:
\begin{equation}
    \begin{split}
        L_{ds} = \frac{1}{|R|} \sum_ {r \in R} \sum_{(x,y) \in D} &\|d(x,y)-d(x,y+1)\|^2_2 \\ 
        &+\|d(x,y)-d(x+1,y)\|^2_2,  \label{eq:ds}
    \end{split}
\end{equation}
where $\mathit{R}$ represents the set of rays emanating from the sampled views, $\mathit{D}$ refers to the depth patch that is centered around $\mathit{r}$, and $d(x,y)$ is of the depth value in position $(x,y)$. 

The final objective of ReVoRF is formulated as:
\begin{equation}
    L_{total} = L_{rgb}+L_{bgc}+L_{rs}+\lambda_{ds}L_{ds}. 
\end{equation}
\begin{table}[t]
    \resizebox{0.5\textwidth}{!}{
        \begin{tabular}{c|cccc}
        \hline
        \toprule[1.1pt]
        & \multicolumn{4}{c}{Realistic Synthetic 360° dataset}  \\
        \hline
        Methods          & PSNR↑          & SSIM↑          & LPIPS↓         & Training Time↓     \\
        \hline
        NeRF~\cite{mildenhall2020nerf}             & 15.93          & 0.780          & 0.320          & 2 hrs             \\
    
        PixelNeRF~\cite{yu2021pixelnerf}        & 16.09          & 0.738          & 0.390          & 3-4 days* + 10 hrs \\
        DietNeRF~\cite{jain2021putting}         & 16.06          & 0.793          & 0.306          & 19 hrs             \\
        3DGS~\cite{kerbl3Dgaussians}            & 17.55             & 0.701          & 0.250          & \textbf{3 mins}          \\
        InfoNeRF~\cite{kim2022infonerf}         & 18.62          & 0.811          & 0.230          & 4 hrs              \\
        VGOS~\cite{ijcai2023p157}             & 18.91          & 0.825          & 0.205          & \textbf{3 mins}    \\
        GeCoNeRF~\cite{kwak2023geconerf}         & \underline{19.78} & \textbf{0.880} & {0.185} & $>$ 2 hrs            \\
        \hline
        Ours             & \textbf{20.72} & \underline{0.848} & \textbf{0.179} & \underline{7 mins}\\
    
        \bottomrule[1.1pt]
        \end{tabular}
        }
        \vspace{-3.5mm}
        \caption{Quantitative comparison for 4-views setting in the Realistic Synthetic $360^\circ$ dataset~\cite{mildenhall2020nerf}. The best and the second-best results are highlighted in \textbf{bold} and \underline{underlined}, respectively. (*) denotes the time cost of pre-training.}
        \label{tab:blender}
        \vspace{-4mm}
        \end{table}
    
\begin{figure*}
    \centering
    \resizebox{0.205\linewidth}{!}{
    \begin{subfigure}{.24\linewidth}
        \centering
        \begin{tikzpicture}[spy using outlines={rectangle, magnification=3, size=33pt}]
            \node[anchor=south west, inner sep=0] at (0,0){\includegraphics[width=\linewidth]{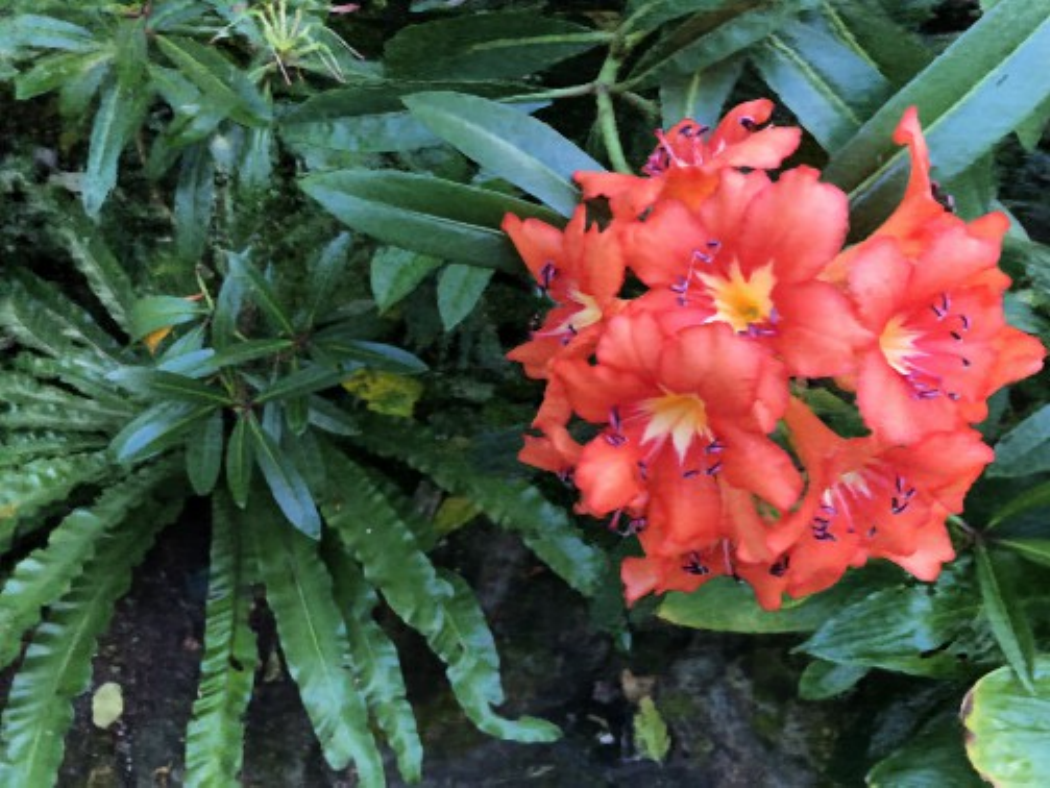}};
            \spy [blue] on (1.3,2.13) in node [right] at (0.05,17pt);
            \spy [red] on (1.7,2.9) in node [right] at (3.0,72pt);
            \end{tikzpicture}
                \\
            \begin{tikzpicture}[spy using outlines={rectangle, magnification=3, size=33pt}]
            \node[anchor=south west, inner sep=0] at (0,0){\includegraphics[width=\linewidth]{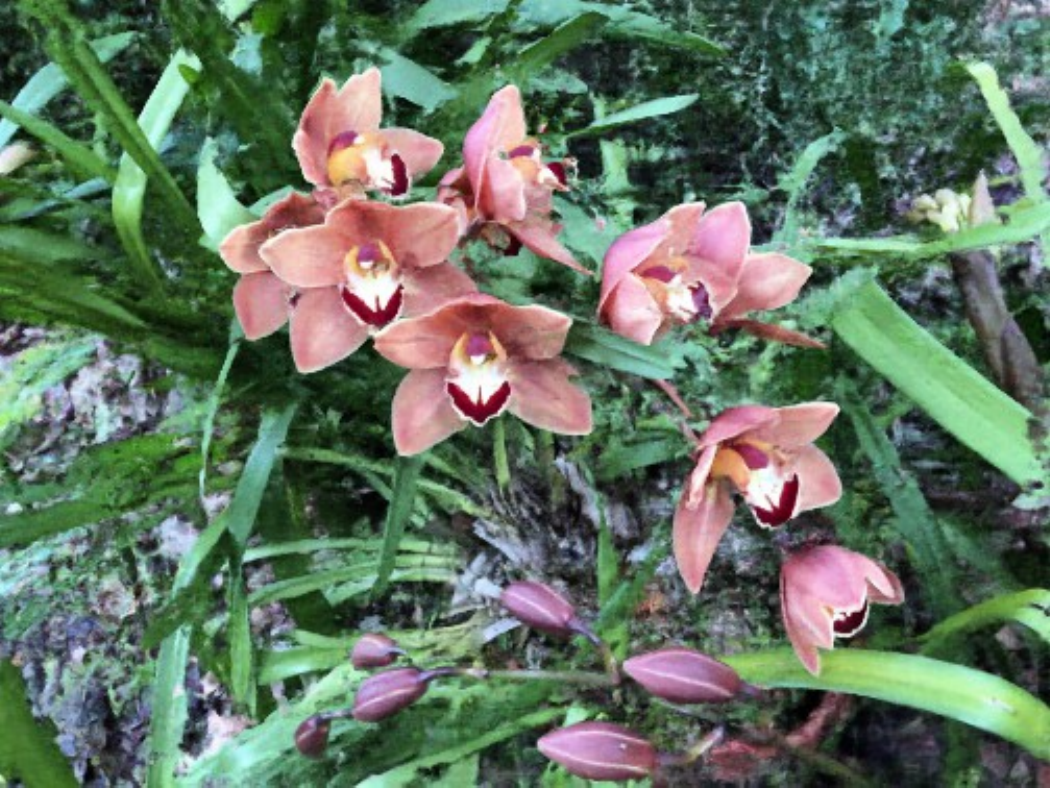}};
            \spy [blue] on (2.7,1.6) in node [right] at (0.05,17pt);
            \spy [red] on (2.3,2.5) in node [right] at (3.0,72pt);
            \end{tikzpicture}
                \\
            \caption{\footnotesize{RegNeRF~\cite{Niemeyer2021Regnerf}}} \label{fig:llff_reg}
    \end{subfigure}
    }
    \hspace{-3mm}
    \resizebox{0.205\linewidth}{!}{
        \begin{subfigure}{.24\linewidth}
        \centering
        \begin{tikzpicture}[spy using outlines={rectangle, magnification=3, size=33pt}]
            \node[anchor=south west, inner sep=0] at (0,0){\includegraphics[width=\linewidth]{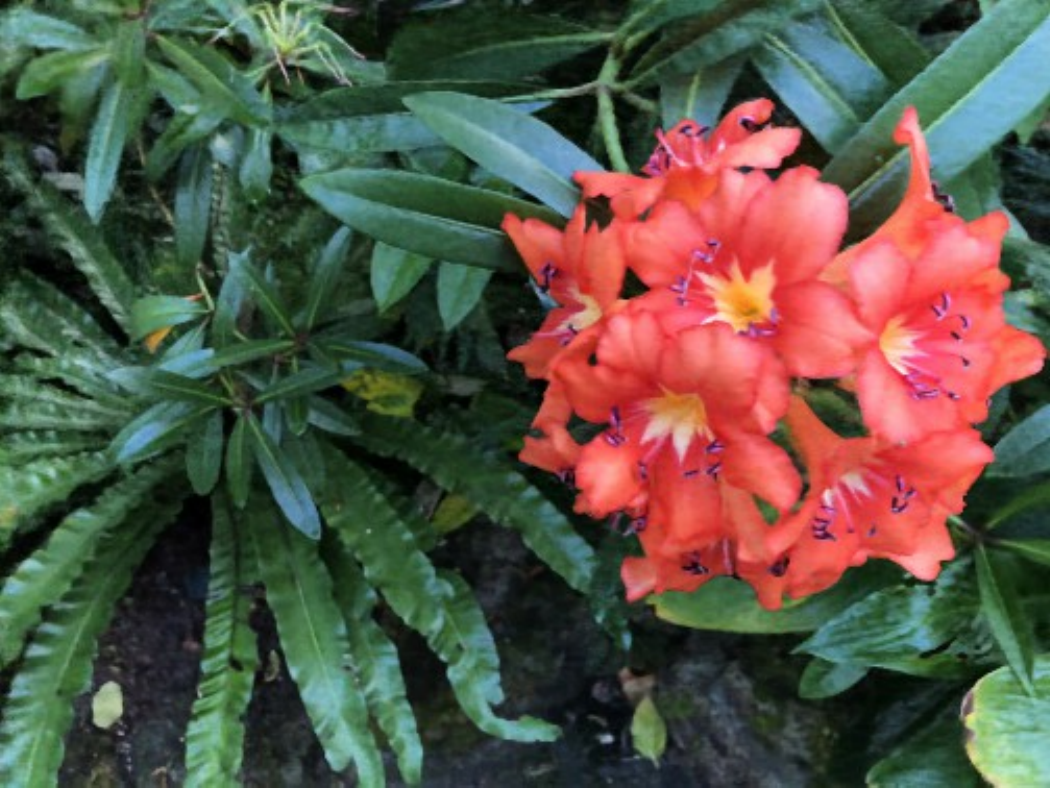}};
            \spy [blue] on (1.3,2.13) in node [right] at (0.05,17pt);
            \spy [red] on (1.7,2.9) in node [right] at (3.0,72pt);
            \end{tikzpicture}
                    \\
            \begin{tikzpicture}[spy using outlines={rectangle, magnification=3, size=33pt}]
            \node[anchor=south west, inner sep=0] at (0,0){\includegraphics[width=\linewidth]{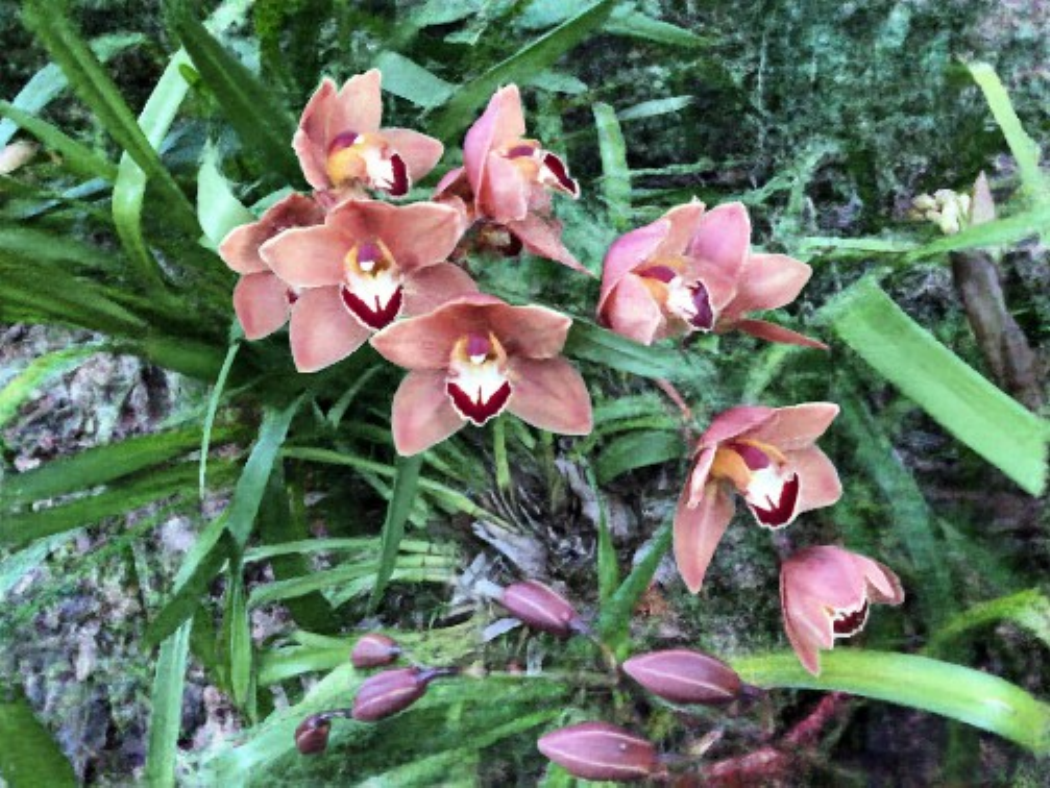}};
            \spy [blue] on (2.7,1.6) in node [right] at (0.05,17pt);
            \spy [red] on (2.3,2.5) in node [right] at (3.0,72pt);
            \end{tikzpicture}
                    \\
            \caption{\footnotesize{SparseNeRF~\cite{guangcong2023sparsenerf}}} \label{fig:llff_sparse}
        \end{subfigure}
    }
    \hspace{-3mm}
    \resizebox{0.205\linewidth}{!}{
        \begin{subfigure}{.24\linewidth}
        \centering
        \begin{tikzpicture}[spy using outlines={rectangle, magnification=3, size=33pt}]
            \node[anchor=south west, inner sep=0] at (0,0){\includegraphics[width=\linewidth]{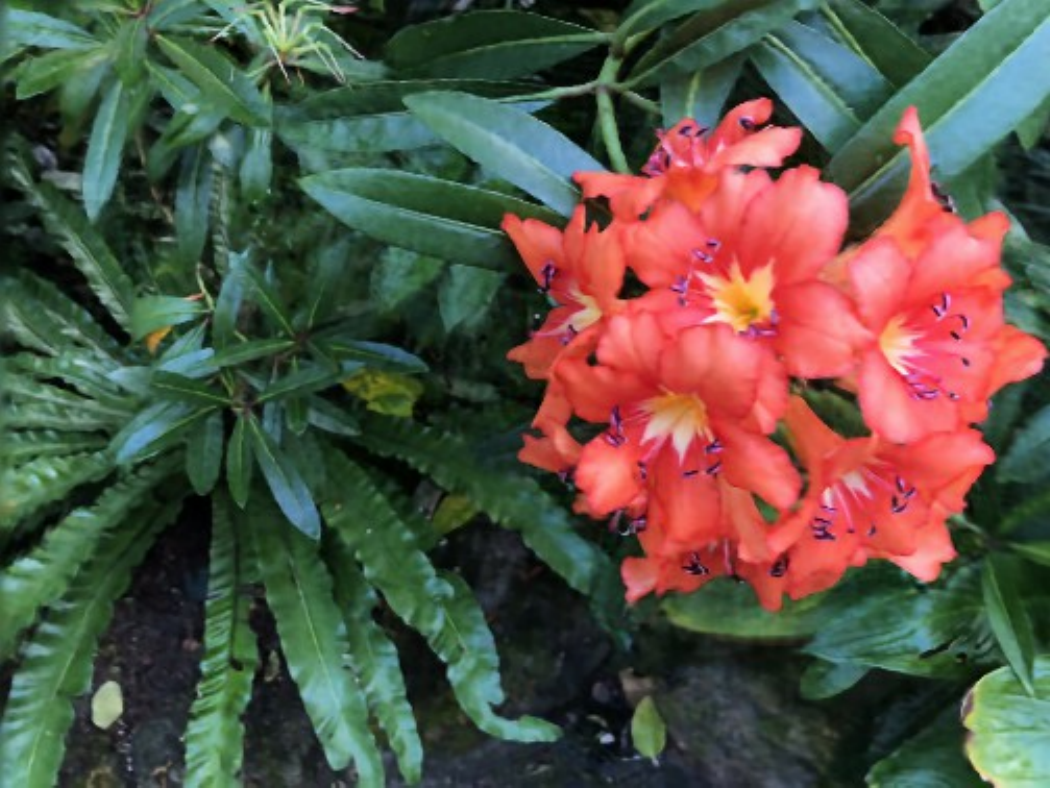}};
            \spy [blue] on (1.3,2.13) in node [right] at (0.05,17pt);
            \spy [red] on (1.7,2.9) in node [right] at (3.0,72pt);
            \end{tikzpicture}
                    \\
            \begin{tikzpicture}[spy using outlines={rectangle, magnification=3, size=33pt}]
            \node[anchor=south west, inner sep=0] at (0,0){\includegraphics[width=\linewidth]{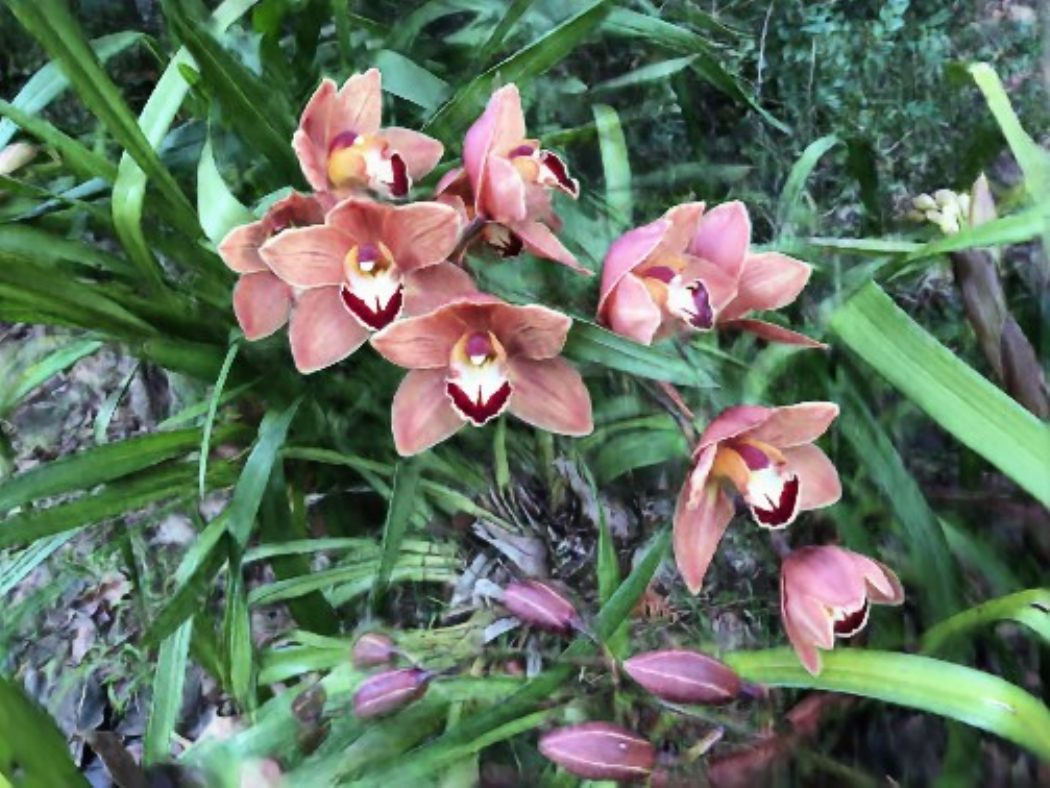}};
            \spy [blue] on (2.7,1.6) in node [right] at (0.05,17pt);
            \spy [red] on (2.3,2.5) in node [right] at (3.0,72pt);
            \end{tikzpicture}
                \\
            \caption{\footnotesize{VGOS~\cite{ijcai2023p157}}} \label{fig:llff_vgos}
        \end{subfigure}
    }
    \hspace{-3mm}
    \resizebox{0.205\linewidth}{!}{
        \begin{subfigure}{.24\linewidth}
        \centering
        \begin{tikzpicture}[spy using outlines={rectangle, magnification=3, size=33pt}]
            \node[anchor=south west, inner sep=0] at (0,0){\includegraphics[width=\linewidth]{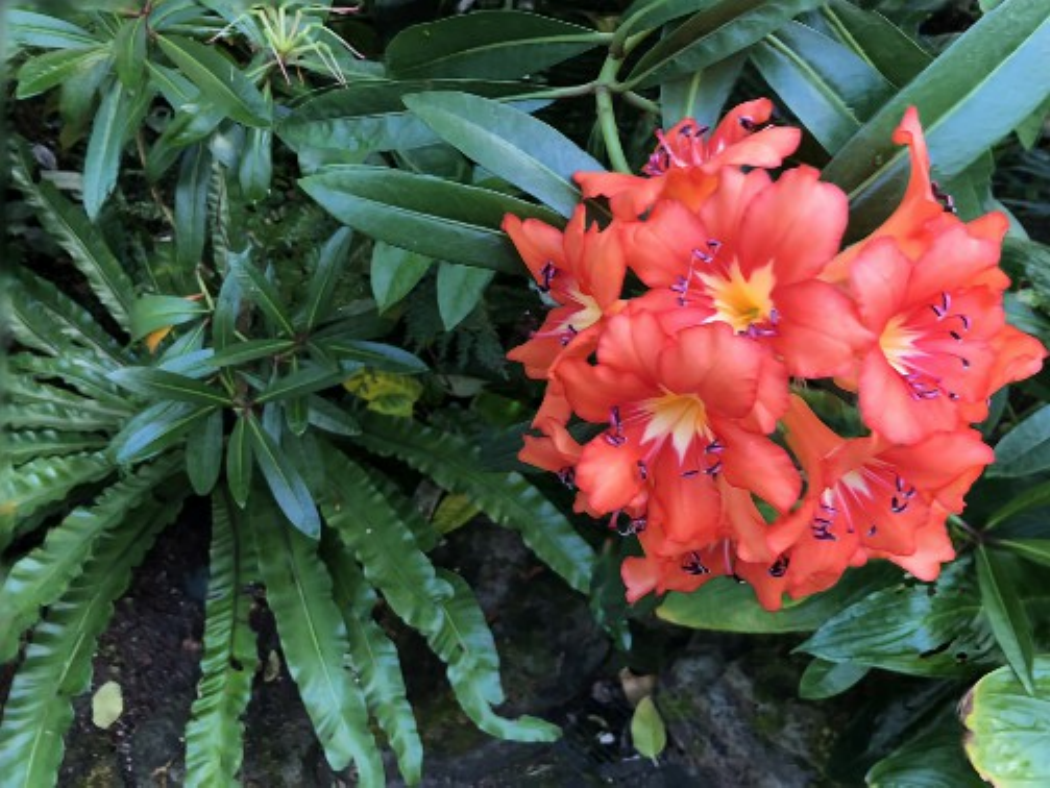}};
            \spy [blue] on (1.3,2.13) in node [right] at (0.05,17pt);
            \spy [red] on (1.7,2.9) in node [right] at (3.0,72pt);
            \end{tikzpicture}
                 \\
            \begin{tikzpicture}[spy using outlines={rectangle, magnification=3, size=33pt}]
            \node[anchor=south west, inner sep=0] at (0,0){\includegraphics[width=\linewidth]{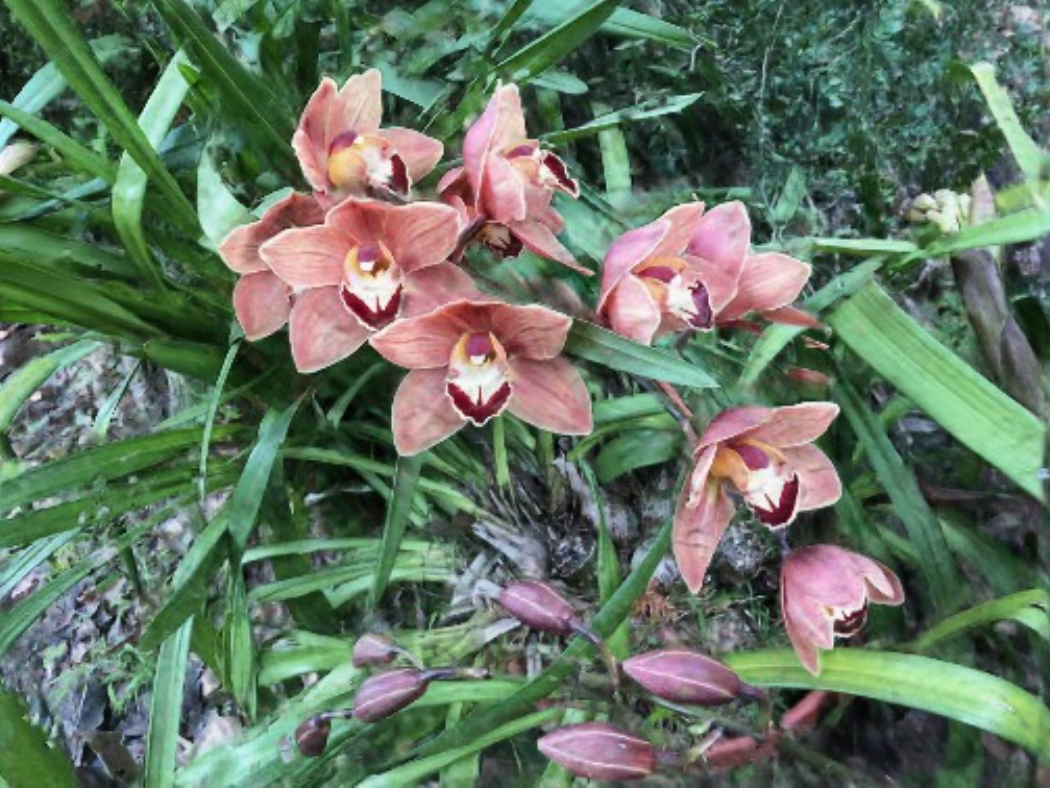}};
            \spy [blue] on (2.7,1.6) in node [right] at (0.05,17pt);
            \spy [red] on (2.3,2.5) in node [right] at (3.0,72pt);
            \end{tikzpicture}
                 \\
            \caption{\footnotesize{ReVoRF (Ours)}} \label{fig:llff_ours}
        \end{subfigure}
    }
    \hspace{-3mm}
    \resizebox{0.205\linewidth}{!}{
        \begin{subfigure}{.24\linewidth}
        \centering
        \begin{tikzpicture}[spy using outlines={rectangle, magnification=3, size=33pt}]
            \node[anchor=south west, inner sep=0] at (0,0){\includegraphics[width=\linewidth]{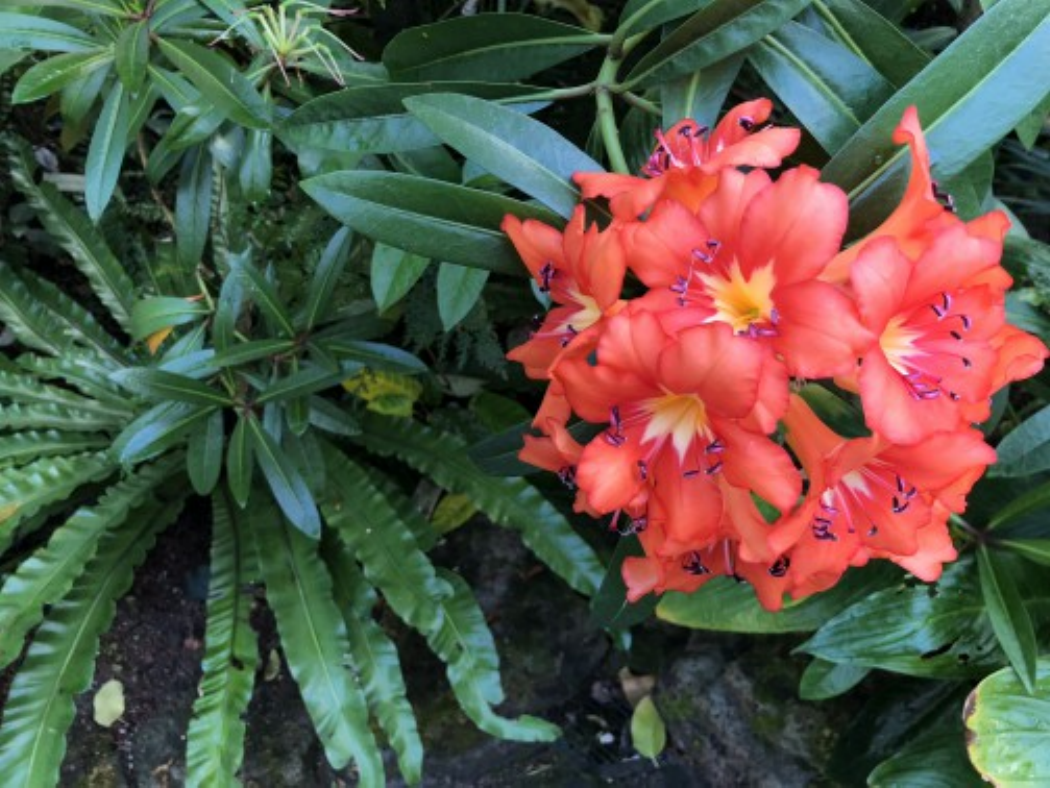}};
            \spy [blue] on (1.3,2.13) in node [right] at (0.05,17pt);
            \spy [red] on (1.7,2.9) in node [right] at (3.0,72pt);
            \end{tikzpicture}
                    \\
            \begin{tikzpicture}[spy using outlines={rectangle, magnification=3, size=33pt}]
            \node[anchor=south west, inner sep=0] at (0,0){\includegraphics[width=\linewidth]{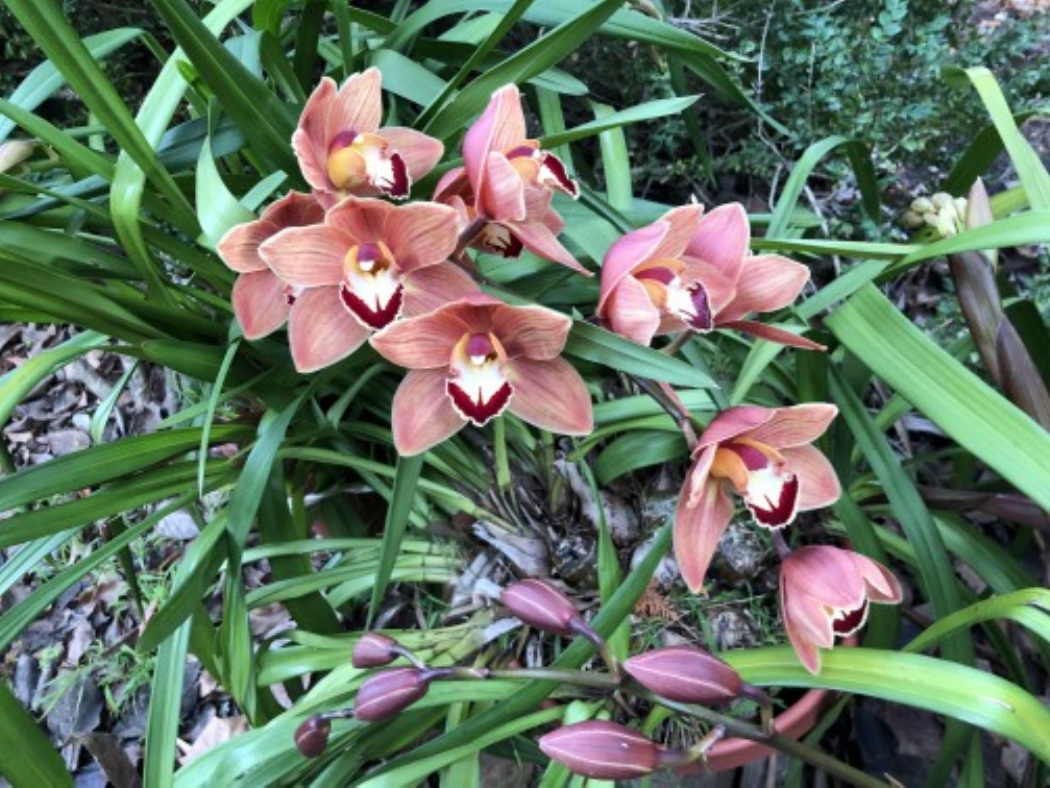}};
            \spy [blue] on (2.7,1.6) in node [right] at (0.05,17pt);
            \spy [red] on (2.3,2.5) in node [right] at (3.0,72pt);
            \end{tikzpicture}
                    \\
            \caption{\footnotesize{GT}} \label{fig:llff_gt}
        \end{subfigure}
    }
    \vspace{-2mm}
    \caption{Comparisons on the LLFF dataset~\cite{MildenhallSCKRN19} in 3-view setting. The red and blue boxes denote compared regions. Our approach achieves better results in reconstructing fine details with enhanced clarity. Please zoom in for details.}
    \label{fig:comparisonsllff}
    \vspace{-4mm}

\end{figure*} 
\section{Experiments}

In this section, we demonstrate the superiority of the proposed ReVoRF through extensive experiments. The details of experiment settings are discussed in Sec.~\ref{sec:settings}. Analysis on comparison experiments and ablation study are performed in Sec.~\ref{sec:compare} and Sec.~\ref{sec:ablation}, respectively.

\subsection{Experiment Settings}\label{sec:settings}

\paragraph{Datasets.}
The experiments are conducted on inward-facing scenes from the Realistic Synthetic 360° dataset~\cite{MildenhallSTBRN20} and forward-facing scenes from the LLFF dataset~\cite{MildenhallSCKRN19}. 

\textbf{Realistic Synthetic 360°} comprises path-traced images from 8 synthetic scenes, which are characterized by their complex geometry and realistic rendering of non-Lambertian materials. Each scene is represented by 400 images, rendered by inward-facing virtual cameras positioned at varying viewpoints. We adhere to the protocol established by InfoNeRF~\cite{kim2022infonerf} and randomly select 4 views from 100 training images as sparse inputs. The model's performance is then evaluated on a set of 200 testing images.

\textbf{LLFF} consists of 8 real-world scenes captured with a handheld cellphone, featuring 20 to 62 forward-facing images per scene. These scenes encompass a range of complex environments. In line with the standard protocol~\cite{MildenhallSTBRN20}, we reserve 1/8 of these images for testing purposes. The remaining images are used for training, from which we randomly sample three views for input into our model.

\midparaheading{Implementation Details.}
Following DVGO~\cite{sun2022direct,sun2022improved}, we adopt a coarse-to-fine optimization scheme to stabilize the training of ReVoRF and gradually improve the geometric details. During the whole training period, we set the values of $\lambda_{rel}$ and $\lambda_{unr}$ in Eq.~\ref{eq:bgc} as $10^{-1}$ and $10^{-2}$, respectively. The values of $\lambda_{d}, \lambda_{f}$ in Eq.~\ref{eq:rs}, and $\lambda_{ds}$ in Eq.~\ref{eq:ds} are set as \scinum{5}{-4}, \scinum{5}{-5}, and \scinum{5}{-5} in the coarse stage, and decreased to \scinum{5}{-5}, $10^{-5}$, and \scinum{5}{-5} in the fine stage. 
The warping poses collected for the Realistic Synthetic 360° dataset~\cite{mildenhall2020nerf} and the LLFF dataset~\cite{li2021neural} are different. For Realistic Synthetic 360°, we randomly vary the polar angle $\mathbf{\theta}$ and azimuthal angle $\mathbf{\phi}$ in the range of $\left[5^\circ, 10^\circ\right]$ based on the input view, and subsequently warp each input sparse view to its four neighboring views defined by $\left\{(\mathbf{\theta}, \mathbf{\phi}), (-\mathbf{\theta}, \mathbf{\phi}), (\mathbf{\theta}, -\mathbf{\phi}), (-\mathbf{\theta}, -\mathbf{\phi})\right\}$.   
For the LLFF dataset, the warped views are obtained by randomly interpolating between every adjacent input view. To speed up the training stage and improve the quality of depth supervision, the warping is performed periodically, which updates the warped depth maps and RGB images every 1000 training steps.

\midparaheading{Evaluation Metrics.}
To assess the effectiveness of our method, we employ several established metrics,
including PSNR (Peak Signal-to-Noise Ratio) for assessing image reconstruction accuracy, SSIM (Structural Similarity Index Measure)~\cite{WangBSS04} for evaluating changes in luminance and contrast that affect structural integrity, and LPIPS (Learned Perceptual Image Patch Similarity)~\cite{ZhangIESW18}, which uses deep learning to approximate human visual perception. These metrics provide a comprehensive analysis of our model's performance, covering aspects of accuracy, perceptual quality, and structural fidelity in the reconstructed images.

\begin{table}
    \resizebox{0.5\textwidth}{!}{
    \begin{tabular}{c|cccc}
    \hline
    \toprule[1.1pt]
    & \multicolumn{4}{c}{NeRF LLFF dataset}  \\
    \hline
    Methods                         & PSNR↑ & SSIM↑ & LPIPS↓ & Training Time↓     \\
    \hline
    PixelNeRF~\cite{yu2021pixelnerf}                       & 16.17 & 0.438 & 0.512  & 3-4 days* + 10 hrs \\
    SRF~\cite{chibane2021stereo}                             & 17.07 & 0.436 & 0.529  & 2-3 days* + 43mins \\
    MVSNeRF~\cite{chen2021mvsnerf}                         & 17.88 & 0.584 & \underline{0.327}  & 1-2 days* + 10mins \\
    Mip-NeRF~\cite{barron2021mip}                        & 14.62 & 0.351 & 0.495  & 14 hrs             \\
    DietNeRF~\cite{jain2021putting}                        & 14.94 & 0.370 & 0.496  & 18 hrs             \\
    3DGS~\cite{kerbl3Dgaussians}                           & 13.05 & 0.407 & 0.388  & 13 mins    \\
    RegNeRF~\cite{Niemeyer2021Regnerf}                         & 19.08 & 0.587 & 0.336  & 4 hrs              \\
    VGOS~\cite{ijcai2023p157}                            & \underline{19.35} & 0.620 & 0.432  & \textbf{5 mins}    \\
    SparseNeRF~\cite{guangcong2023sparsenerf}                      & \textbf{19.86} & \underline{0.624} & 0.328  & $>$ 2 hrs          \\
    \hline
    Ours                            & 19.26 & \textbf{0.644} & \textbf{0.316}  & \underline{11 mins}      \\
    \bottomrule[1.1pt]
    \end{tabular}
    }
    \vspace{-3.5mm}
    \caption{Quantitative comparison for 3-views setting on LLFF~\cite{li2021neural}. The best and the second-best results are highlighted in \textbf{bold} and \underline{underlined}, respectively. (*) signifies the pre-training time.}
    \vspace{-4mm}
    \label{tab:llff}
    \end{table}

\begin{figure*}
    \centering
    \begin{subfigure}{.19\linewidth}
    \centering
            \includegraphics[width=\linewidth]{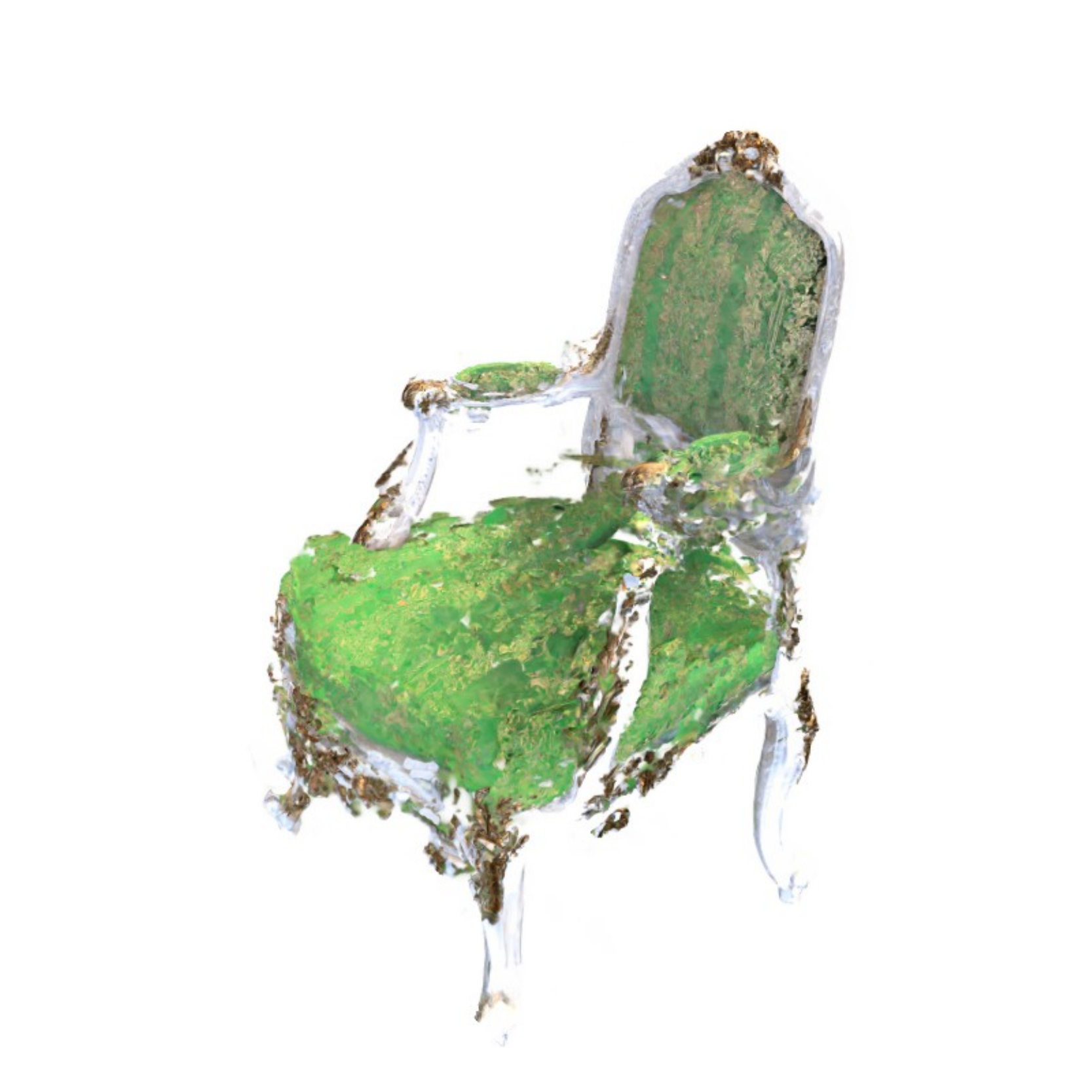} \\
        \caption*{\footnotesize{Baseline}}
    \end{subfigure}
    \hspace{-1.7mm}
    \begin{subfigure}{.19\linewidth}
    \centering
            \includegraphics[width=\linewidth]{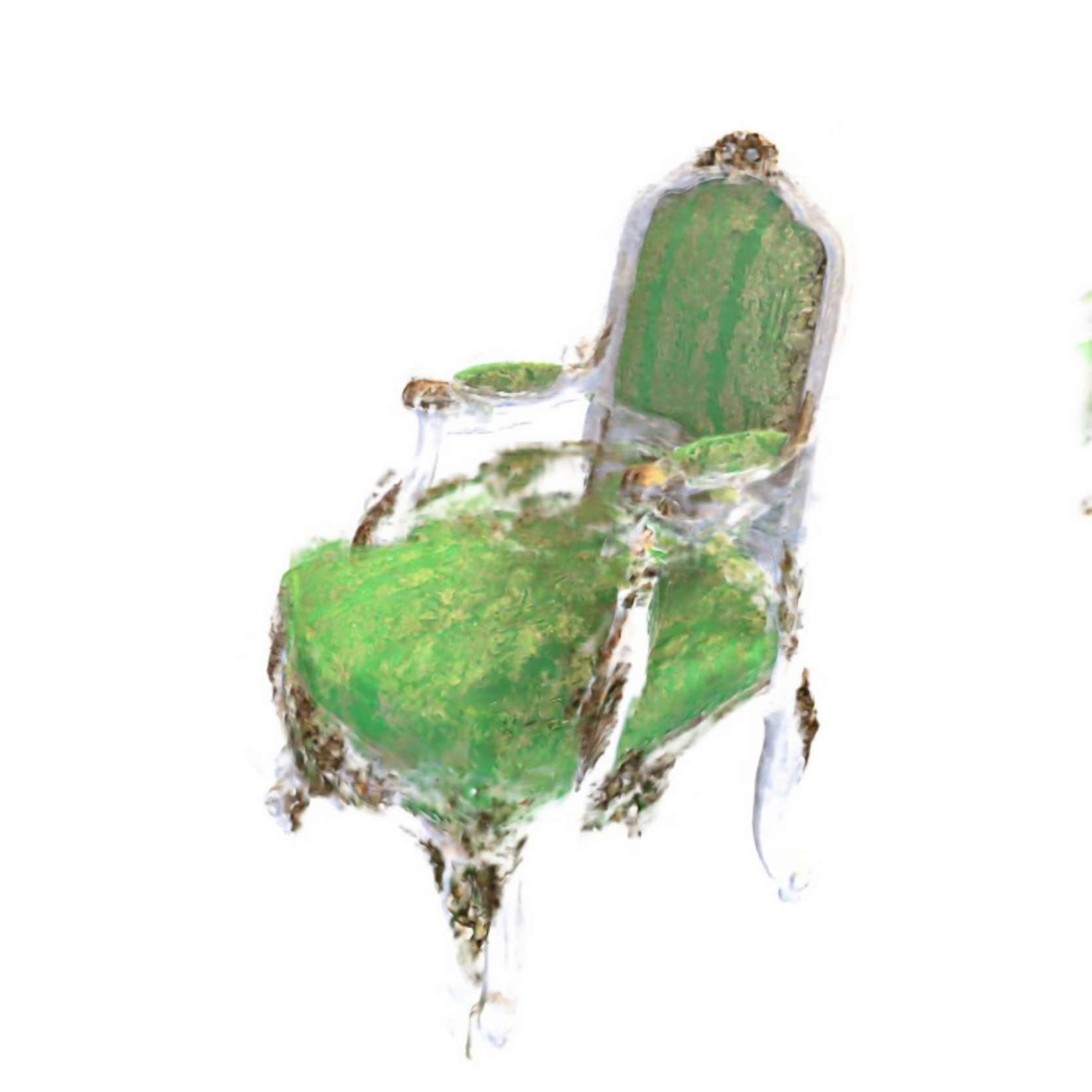}\\
        \caption*{\footnotesize{Baseline+$\textit{L}_{rel}$}\label{edge_a}}
    \end{subfigure}
    \hspace{-1.7mm}
    \begin{subfigure}{.19\linewidth}
    \centering
            \includegraphics[width=\linewidth]{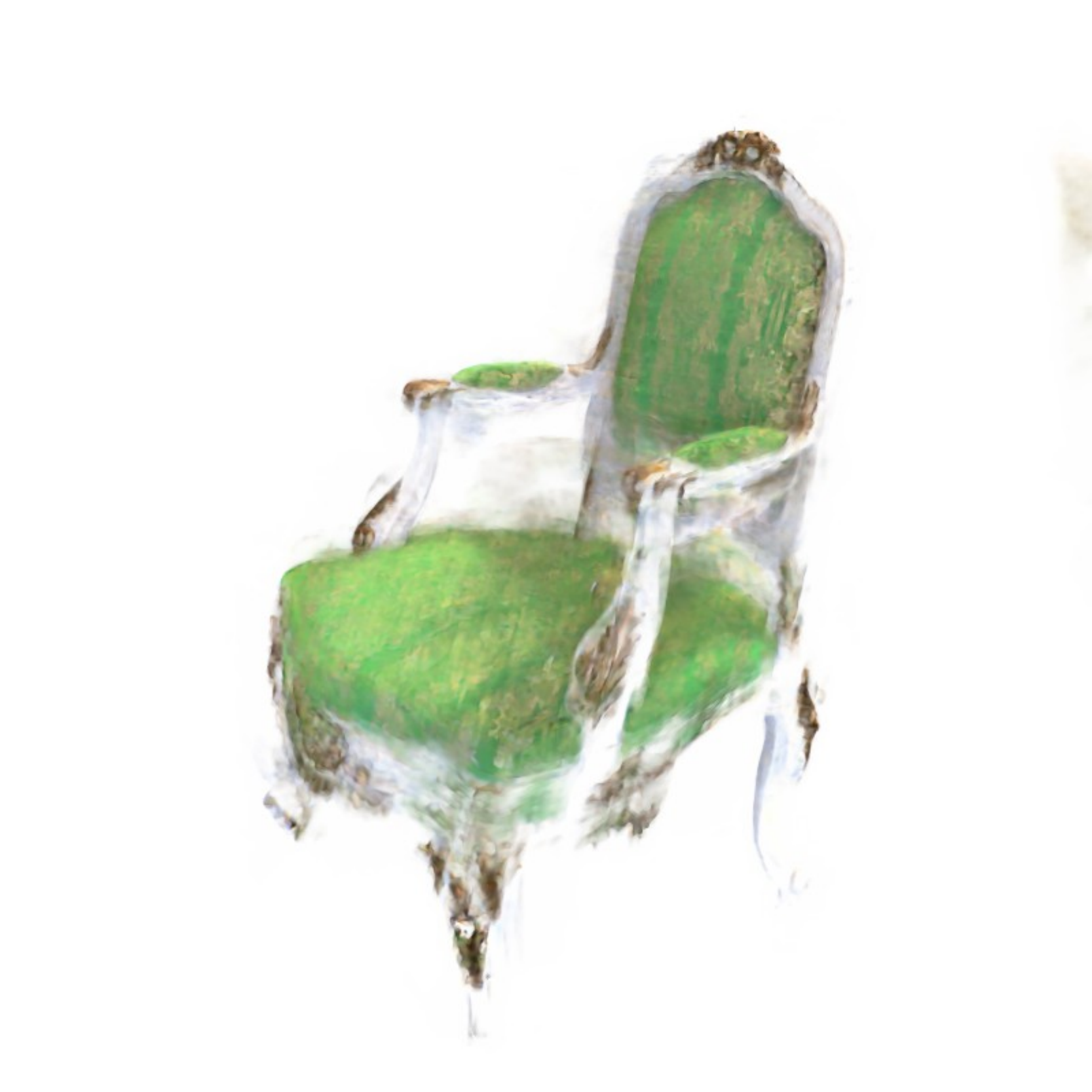}\\
        \caption*{\footnotesize{Baseline+$\textit{L}_{rel}$+$\textit{L}_{unr}$}\label{edge_a}}
    \end{subfigure}
    \hspace{-1.7mm}
    \begin{subfigure}{.19\linewidth}
        \centering
        \includegraphics[width=\linewidth]{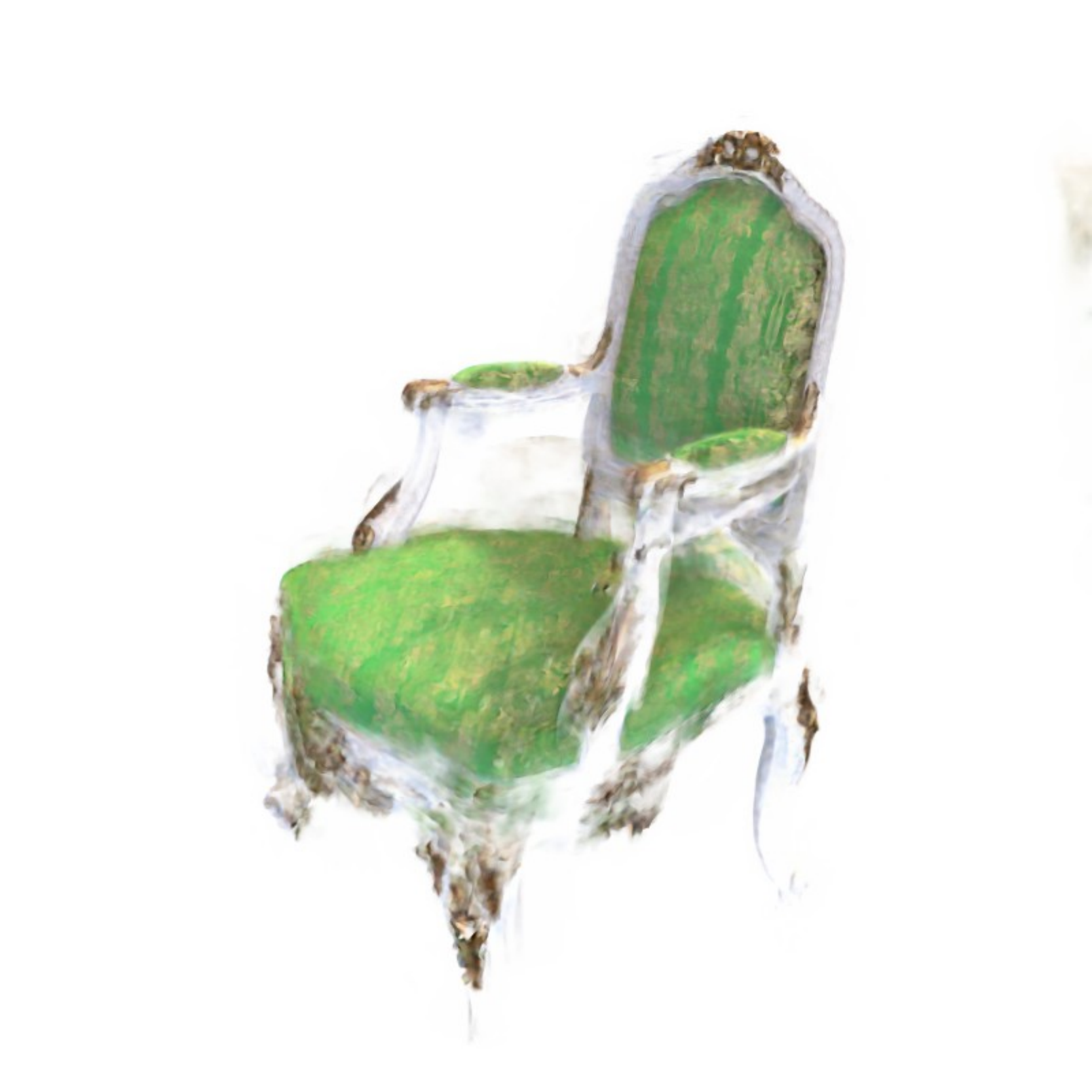}\\
        \caption*{\footnotesize{Baseline+$\textit{L}_{rel}$+$\textit{L}_{unr}$+$\textit{L}_{ds}$}\label{edge_b}}
    \end{subfigure}
    \hspace{-1.7mm}
    \begin{subfigure}{.19\linewidth}
    \centering
    \includegraphics[width=\linewidth]{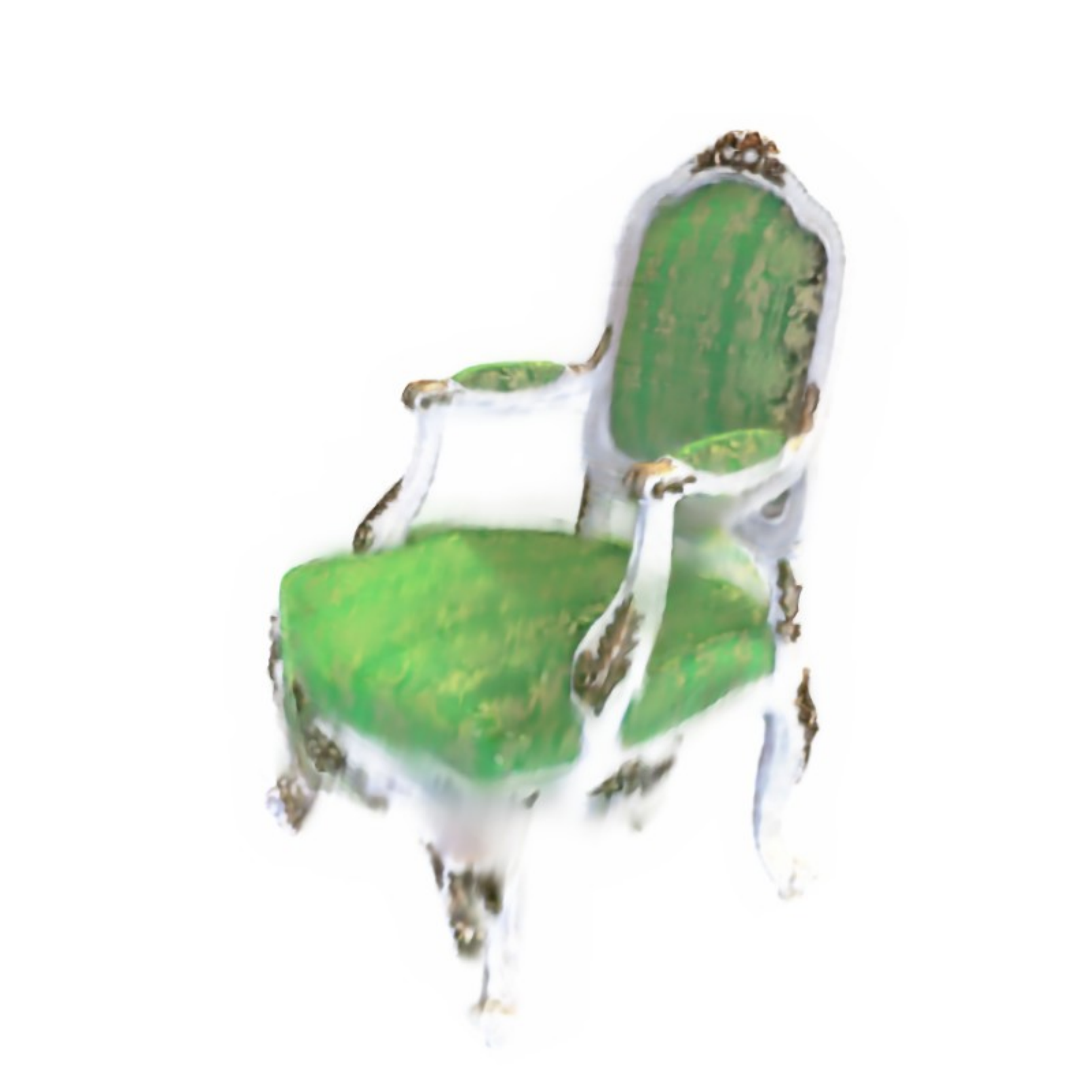}\\
        \caption*{\footnotesize{Full Model}\label{edge_ours}}
    \end{subfigure}
    \vspace{-2mm}
    \caption{Visualizations of the ablation on Chair scene from the Realistic Synthetic 360°~\cite{MildenhallSTBRN20} dataset in 4 views setting. With the proposed losses, our methods could gradually improve the cross-view consistency and reduce the noise compared with the baseline.}
    \vspace{-4mm}
    \label{fig:Ablation}

\end{figure*} 

\subsection{Comparisons}\label{sec:compare}
On the Realistic Synthetic 360° dataset~\cite{mildenhall2020nerf}, we compare our method with state-of-the-art approaches, including RegNeRF~\cite{Niemeyer2021Regnerf}, DietNeRF~\cite{jain2021putting}, infoNeRF~\cite{kim2022infonerf}, VGOS~\cite{ijcai2023p157},PixelNeRF~\cite{yu2021pixelnerf}, and GeCoNeRF~\cite{kwak2023geconerf}, in a 4-view setting.
On the LLFF dataset~\cite{li2021neural}, we implement our method in a 3-view setting and compare with SRF~\cite{chibane2021stereo}, MVSNeRF~\cite{chen2021mvsnerf}, mip-NeRF~\cite{barron2021mip}, DietNeRF~\cite{jain2021putting}, RegNeRF~\cite{Niemeyer2021Regnerf},VGOS~\cite{ijcai2023p157}, SparseNeRF~\cite{guangcong2023sparsenerf} and GeCoNeRF~\cite{kwak2023geconerf}. We adopt the reported results from VGOS~\cite{ijcai2023p157}, sparseNeRF~\cite{guangcong2023sparsenerf}, and GeCoNeRF~\cite{kwak2023geconerf}. Besides, we also compare with the advanced reconstruction method 3DGS~\cite{kerbl3Dgaussians}.

\midparaheading{Qualitative Experiments.}\label{sec:quality_compare}

Fig.~\ref{fig:comparisons} compares our approach with some recent methods on the Realistic Synthetic 360° dataset~\cite{mildenhall2020nerf}. Diet-NeRF~\cite{jain2021putting} performs poorly in the setting of 4 views, while infoNeRF~\cite{kim2022infonerf} and VGOS~\cite{ijcai2023p157} are inferior to our method in terms of both geometric shapes and detail resolution. Our approach demonstrates superior performance in both geometry and details.

Fig.~\ref{fig:comparisonsllff} shows a qualitative comparison on a scene from the LLFF dataset~\cite{li2021neural}. While all methods can recover the overall structure of the scene, our approach excels at the quality of details as shown in the magnified regions. Our method incorporates smoothness while retaining fine details, achieving the most natural results.

\midparaheading{Quantitative Experiments.}\label{sec:quantity_compare}

Table~\ref{tab:blender} shows quantitative results from different methods on the Realistic Synthetic 360° dataset~\cite{mildenhall2020nerf}. In terms of training time, our method is at least an order of magnitude faster than all other methods except for VGOS~\cite{ijcai2023p157}. Although our method is slightly lower than VGOS~\cite{ijcai2023p157}, it significantly enhances the PSNR, LPIPS~\cite{ZhangIESW18}, and SSIM~\cite{WangBSS04} of the rendered images. Our method achieves state-of-the-art accuracy in PSNR and LPIPS~\cite{ZhangIESW18}. Additionally, despite not utilizing a pre-trained model or perceptual loss for high-level semantic information extraction, our method still achieves the second-best performance in perceived SSIM~\cite{WangBSS04}.

Table~\ref{tab:llff} shows a quantitative comparison on the LLFF dataset~\cite{li2021neural}. Our method achieves the highest scores in both SSIM~\cite{WangBSS04} and LPIPS~\cite{ZhangIESW18}, indicating that our images exhibit the best performance in terms of human perceptual reconstruction. We also achieve the third-highest PSNR, which, together with our state-of-the-art performance in SSIM~\cite{WangBSS04} and LPIPS~\cite{ZhangIESW18}, demonstrates that our approach has made improvements in certain aspects of image rendering.

\subsection{Ablation Study}\label{sec:ablation}

\begin{table}
    \centering
\resizebox{0.4\textwidth}{!}{
    \begin{tabular}{ccccccc}
    \toprule[1.1pt]
  $\textit{L}_{rs}$  &$\textit{L}_{ds}$& $\textit{L}_{unr}$ & $\textit{L}_{rel}$ & PSNR↑ & SSIM↑ & LPIPS↓ \\
  \hline
                  &            &               &               & 17.19 & 0.767 & 0.223 \\
                  &            &               & $\checkmark$               & 17.79 & 0.780 & 0.243 \\
                  &            & $\checkmark$               & $\checkmark$               & 19.01 & 0.805 & 0.228 \\
                  & $\checkmark$            & $\checkmark$               & $\checkmark$        & \underline{19.23} & \underline{0.811} & \underline{0.220} \\
\hline
$\checkmark$                & $\checkmark$            & $\checkmark$               & $\checkmark$        & \textbf{20.72} & \textbf{0.848} & \textbf{0.179} \\
    \bottomrule[1.1pt]
    \end{tabular}
    }
    \vspace{-3mm}
    \caption{Ablation study on the Realistic Synthetic $360^\circ$ dataset~\cite{mildenhall2020nerf} in the 4-view setting. The best and the second-best results are highlighted in \textbf{bold} and \underline{underlined}, respectively.}
    \vspace{-2mm}
    \label{tab:ablation}
\end{table}

Our ablation study is segmented into five distinct groups with Table~\ref{tab:ablation}, with DVGO~\cite{sun2022direct,sun2022improved} serving as the baseline. We incrementally introduce our proposed contributions along with various regularization methods to enhance the model's rendering quality. The groups are delineated as: baseline, baseline+$\textit{L}_{rel}$, baseline+$\textit{L}_{rel}$+$\textit{L}_{unr}$, baseline+$\textit{L}_{rel}$+$\textit{L}_{unr}$+$\textit{L}_{ds}$, and baseline+$\textit{L}_{rel}$+$\textit{L}_{unr}$+$\textit{L}_{ds}$+$\textit{L}_{rs}$. The ablation results reveal that each incremental contribution positively impacts the rendering quality in various aspects. After the addition of $\textit{L}_{rel}$, our PSNR increased by 0.6. Subsequent inclusion of $\textit{L}_{unr}$ led to a further rise in PSNR by $\textbf{1.22}$. With the incorporation of $\textit{L}_{ds}$, the PSNR went from 19.01 to 19.23. Finally, after adding $\textit{L}_{rs}$, our PSNR peaked at $\textbf{20.72}$, SSIM~\cite{WangBSS04} reached $\textbf{0.848}$, and LPIPS~\cite{ZhangIESW18} arrived at $\textbf{0.179}$, marking a significant enhancement.

Besides, we explore the potential of ReVoRF under the settings where more input views are available. We report extra results for 6-view and 9-view settings in Table~\ref{tab:ablation-views}, demonstrating that increased input views generally enhance performance. ReVoRF maintains superiority across these settings.

Fig.~\ref{fig:Ablation} presents the visualization of our ablation study on the Chair scene. We incorporated $\textit{L}_{rel}$ to enhance the texture quality of the model, which, however, introduced some noise artifacts. To mitigate these artifacts, $\textit{L}_{unr}$ and $\textit{L}_{ds}$ were subsequently integrated. These adjustments successfully reduced noise but at the cost of blurring the geometric structures in the process. The issue of maintaining geometric consistency while eliminating noise was addressed through the implementation of $\textit{L}_{rs}$. We observe that our method effectively prevents the collapse of new views caused by overfitting due to a limited number of viewpoints.

\begin{table}[t]
    \setlength{\tabcolsep}{0.01cm}{
    \resizebox{0.48\textwidth}{!}{
        \begin{tabular}{c|ccc|ccc|ccc}
        \hline
        \toprule[1.1pt]
        \multirow{2}{*}{Methods}                 & \multicolumn{3}{c|}{PSNR↑}   & \multicolumn{3}{c|}{SSIM↑}   & \multicolumn{3}{c}{LPIPS↓}        \\
        \cline{2-10}
            & 3-view & 6-view & 9-view & 3-view & 6-view & 9-view & 3-view & 6-view & 9-view\\
        \hline
        SRF~\cite{chibane2021stereo}          & 17.07  & 16.75  & 17.39    & 0.436  & 0.438  & 0.465    & 0.529  & 0.521  & 0.503 \\
        PixelNeRF~\cite{yu2021pixelnerf}          & 16.17  & 17.03  & 18.92    & 0.438  & 0.473  & 0.535    & 0.512  & 0.477  & 0.430 \\
        MVSNeRF~\cite{chen2021mvsnerf}          & 17.88  & 19.99  & 20.47    & 0.584  & 0.660  & 0.695    & 0.327  & 0.264  & 0.244 \\
        \hline
        DVGO~\cite{sun2022direct}             & 16.60  & 21.25  & \underline{22.89}    & 0.560  & \underline{0.704}  & \underline{0.746}    & \underline{0.422}  & \textbf{0.246}  & \underline{0.228} \\
        VGOS~\cite{ijcai2023p157}             & \textbf{19.35}  & \underline{21.55}  & 22.39    & \underline{0.620}  & 0.671  & 0.692    & 0.432  & 0.328  & 0.325 \\
        Ours             & \underline{19.26}  & \textbf{22.21}  & \textbf{23.04}    & \textbf{0.644}  & \textbf{0.720}  & \textbf{0.753}   & \textbf{0.316}  & \underline{0.269}  & \textbf{0.225} \\
        \bottomrule[1.1pt]
        \end{tabular}
        }
        \vspace{-3.5mm}
        \caption{Comparison of 3, 6, and 9 input views on LLFF~\cite{li2021neural}. The best and the second-best results are highlighted in \textbf{bold} and \underline{underlined}, respectively.}
        \vspace{-4mm}
        \label{tab:ablation-views}
    }
\end{table} 

\section{Conclusion, Limitation, and Future Work}
In this paper, we address the challenge of view deformation by discerning reliable and unreliable areas, subsequently introducing a bilateral geometric consistency regularization. This approach maximizes the use of reliable regions while delicately exploring the depth in unreliable areas, applying a more lenient constraint to these zones. Further extending our method into voxel space, we transform 2D reliable areas into 3D space through a reliability-aware voxelization smoothing process. Our method, when applied to various datasets, has proven to be highly precise, significantly bolstering geometric consistency and demonstrating its efficacy in intricate 3D reconstruction tasks.

Our method shares a common limitation of the voxel-based method: the tendency to produce smoothed results, leading to a loss in fine details. Besides, the exceptionally challenging context for NeRF with sparse input also limits its application in more complex scenes, such as large-scale scene reconstruction.
For future work, we aim to refine the voxelization technique to better preserve details, potentially exploring hybrid models that combine voxel-based methods with alternative geometric representations for a more detailed reconstruction.

\midparaheading{Acknowledgement.}
The work is supported by the Guangdong Natural Science Funds for Distinguished Young Scholar (No. 2023B1515020097), Singapore MOE Tier 1 Funds (MSS23C002), and the NRF Singapore under the AI Singapore Programme (No. AISG3-GV-2023-011).


\begin{thebibliography}{10}\itemsep=-1pt

    \bibitem{AminiSSR20}
    Alexander Amini, Wilko Schwarting, Ava Soleimany, and Daniela Rus.
    \newblock Deep evidential regression.
    \newblock In {\em NeurIPS}, volume~33, pages 14927--14937, 2020.
    
    \bibitem{barron2021mip}
    Jonathan~T Barron, Ben Mildenhall, Matthew Tancik, Peter Hedman, Ricardo
      Martin-Brualla, and Pratul~P Srinivasan.
    \newblock Mip-nerf: A multiscale representation for anti-aliasing neural
      radiance fields.
    \newblock In {\em ICCV}, pages 5855--5864, 2021.
    
    \bibitem{bortolon2023vm}
    Matteo Bortolon, Alessio Del~Bue, and Fabio Poiesi.
    \newblock Vm-nerf: Tackling sparsity in nerf with view morphing.
    \newblock In {\em ICIAP}, pages 63--74. Springer, 2023.
    
    \bibitem{cao2023hexplane}
    Ang Cao and Justin Johnson.
    \newblock Hexplane: A fast representation for dynamic scenes.
    \newblock In {\em CVPR}, pages 130--141, 2023.
    
    \bibitem{chen2021mvsnerf}
    Anpei Chen, Zexiang Xu, Fuqiang Zhao, Xiaoshuai Zhang, Fanbo Xiang, Jingyi Yu,
      and Hao Su.
    \newblock Mvsnerf: Fast generalizable radiance field reconstruction from
      multi-view stereo.
    \newblock In {\em ICCV}, pages 14124--14133, 2021.
    
    \bibitem{chen2022geoaug}
    Di Chen, Yu Liu, Lianghua Huang, Bin Wang, and Pan Pan.
    \newblock Geoaug: Data augmentation for few-shot nerf with geometry
      constraints.
    \newblock In {\em ECCV}, pages 322--337. Springer, 2022.
    
    \bibitem{chibane2021stereo}
    Julian Chibane, Aayush Bansal, Verica Lazova, and Gerard Pons-Moll.
    \newblock Stereo radiance fields (srf): Learning view synthesis for sparse
      views of novel scenes.
    \newblock In {\em CVPR}, pages 7911--7920, 2021.
    
    \bibitem{deng2022depth}
    Kangle Deng, Andrew Liu, Jun-Yan Zhu, and Deva Ramanan.
    \newblock Depth-supervised nerf: Fewer views and faster training for free.
    \newblock In {\em CVPR}, pages 12882--12891, 2022.
    
    \bibitem{du2021nerflow}
    Yilun Du, Yinan Zhang, Hong-Xing Yu, Joshua~B. Tenenbaum, and Jiajun Wu.
    \newblock Neural radiance flow for 4d view synthesis and video processing.
    \newblock In {\em ICCV}, 2021.
    
    \bibitem{fridovich2022plenoxels}
    Sara Fridovich-Keil, Alex Yu, Matthew Tancik, Qinhong Chen, Benjamin Recht, and
      Angjoo Kanazawa.
    \newblock Plenoxels: Radiance fields without neural networks.
    \newblock In {\em CVPR}, pages 5501--5510, 2022.
    
    \bibitem{gao2022get3d}
    Jun Gao, Tianchang Shen, Zian Wang, Wenzheng Chen, Kangxue Yin, Daiqing Li, Or
      Litany, Zan Gojcic, and Sanja Fidler.
    \newblock Get3d: A generative model of high quality 3d textured shapes learned
      from images.
    \newblock {\em NeurIPS}, 35:31841--31854, 2022.
    
    \bibitem{guangcong2023sparsenerf}
    Zhaoxi~Chen Guangcong, Chen~Change Loy, and Ziwei Liu.
    \newblock Sparsenerf: Distilling depth ranking for few-shot novel view
      synthesis.
    \newblock In {\em ICCV}, 2023.
    
    \bibitem{hu2023tri}
    Wenbo Hu, Yuling Wang, Lin Ma, Bangbang Yang, Lin Gao, Xiao Liu, and Yuewen Ma.
    \newblock Tri-miprf: Tri-mip representation for efficient anti-aliasing neural
      radiance fields.
    \newblock In {\em ICCV}, pages 19774--19783, 2023.
    
    \bibitem{jain2021putting}
    Ajay Jain, Matthew Tancik, and Pieter Abbeel.
    \newblock Putting nerf on a diet: Semantically consistent few-shot view
      synthesis.
    \newblock In {\em ICCV}, pages 5885--5894, 2021.
    
    \bibitem{jiang2023diffuse3d}
    Yutao Jiang, Yang Zhou, Yuan Liang, Wenxi Liu, Jianbo Jiao, Yuhui Quan, and
      Shengfeng He.
    \newblock Diffuse3d: Wide-angle 3d photography via bilateral diffusion.
    \newblock In {\em ICCV}, pages 8998--9008, 2023.
    
    \bibitem{KendallG17}
    Alex Kendall and Yarin Gal.
    \newblock What uncertainties do we need in bayesian deep learning for computer
      vision?
    \newblock In {\em NeurIPS}, pages 5574--5584, 2017.
    
    \bibitem{kerbl3Dgaussians}
    Bernhard Kerbl, Georgios Kopanas, Thomas Leimk{\"u}hler, and George Drettakis.
    \newblock 3d gaussian splatting for real-time radiance field rendering.
    \newblock {\em ACM Transactions on Graphics}, 42(4), July 2023.
    
    \bibitem{kim2022infonerf}
    Mijeong Kim, Seonguk Seo, and Bohyung Han.
    \newblock Infonerf: Ray entropy minimization for few-shot neural volume
      rendering.
    \newblock In {\em CVPR}, pages 12912--12921, 2022.
    
    \bibitem{kwak2023geconerf}
    Min-Seop Kwak, Jiuhn Song, and Seungryong Kim.
    \newblock Geconerf: Few-shot neural radiance fields via geometric consistency.
    \newblock In {\em ICML}. JMLR.org, 2023.
    
    \bibitem{LeeCWLKY22}
    Soomin Lee, Le Chen, Jiahao Wang, Alexander Liniger, Suryansh Kumar, and Fisher
      Yu.
    \newblock Uncertainty guided policy for active robotic 3d reconstruction using
      neural radiance fields.
    \newblock {\em {IEEE} Robotics Autom. Lett.}, 7(4):12070--12077, 2022.
    
    \bibitem{li2021neural}
    Zhengqi Li, Simon Niklaus, Noah Snavely, and Oliver Wang.
    \newblock Neural scene flow fields for space-time view synthesis of dynamic
      scenes.
    \newblock In {\em CVPR}, pages 6498--6508, 2021.
    
    \bibitem{liu2021editing}
    Steven Liu, Xiuming Zhang, Zhoutong Zhang, Richard Zhang, Jun-Yan Zhu, and
      Bryan Russell.
    \newblock Editing conditional radiance fields.
    \newblock In {\em ICCV}, pages 5773--5783, 2021.
    
    \bibitem{MartinBruallaR21}
    Ricardo Martin-Brualla, Noha Radwan, Mehdi~SM Sajjadi, Jonathan~T Barron,
      Alexey Dosovitskiy, and Daniel Duckworth.
    \newblock Nerf in the wild: Neural radiance fields for unconstrained photo
      collections.
    \newblock In {\em CVPR}, pages 7210--7219, 2021.
    
    \bibitem{mertan2020siralama}
    Alican Mertan, Damien~Jade Duff, and G{\"o}zde {\"U}nal.
    \newblock Siralama sorunu olarak nispi derinlik tahmini relative depth
      estimation as a ranking problem.
    \newblock In {\em 2020 28th Signal Processing and Communications Applications
      Conference (SIU)}, pages 1--6. IEEE, 2020.
    
    \bibitem{mildenhall2020nerf}
    B Mildenhall, PP Srinivasan, M Tancik, JT Barron, R Ramamoorthi, and R Ng.
    \newblock Nerf: Representing scenes as neural radiance fields for view
      synthesis.
    \newblock In {\em ECCV}, 2020.
    
    \bibitem{MildenhallSCKRN19}
    Ben Mildenhall, Pratul~P. Srinivasan, Rodrigo~Ortiz Cayon, Nima~Khademi
      Kalantari, Ravi Ramamoorthi, Ren Ng, and Abhishek Kar.
    \newblock Local light field fusion: practical view synthesis with prescriptive
      sampling guidelines.
    \newblock {\em ACM TOG}, 38(4):29:1--29:14, 2019.
    
    \bibitem{MildenhallSTBRN20}
    Ben Mildenhall, Pratul~P. Srinivasan, Matthew Tancik, Jonathan~T. Barron, Ravi
      Ramamoorthi, and Ren Ng.
    \newblock Nerf: Representing scenes as neural radiance fields for view
      synthesis.
    \newblock In {\em ECCV}, volume 12346, pages 405--421, 2020.
    
    \bibitem{muller2022instant}
    Thomas M{\"u}ller, Alex Evans, Christoph Schied, and Alexander Keller.
    \newblock Instant neural graphics primitives with a multiresolution hash
      encoding.
    \newblock {\em ACM TOG}, 41(4):1--15, 2022.
    
    \bibitem{Neal95}
    Radford~M. Neal.
    \newblock {\em Bayesian learning for neural networks}.
    \newblock PhD thesis, University of Toronto, Canada, 1995.
    
    \bibitem{Niemeyer2021Regnerf}
    Michael Niemeyer, Jonathan~T. Barron, Ben Mildenhall, Mehdi S.~M. Sajjadi,
      Andreas Geiger, and Noha Radwan.
    \newblock Regnerf: Regularizing neural radiance fields for view synthesis from
      sparse inputs.
    \newblock In {\em CVPR}, 2022.
    
    \bibitem{PanLSH22}
    Xuran Pan, Zihang Lai, Shiji Song, and Gao Huang.
    \newblock Activenerf: Learning where to see with uncertainty estimation.
    \newblock In {\em ECCV}, volume 13693, pages 230--246, 2022.
    
    \bibitem{radford2021learning}
    Alec Radford, Jong~Wook Kim, Chris Hallacy, Aditya Ramesh, Gabriel Goh,
      Sandhini Agarwal, Girish Sastry, Amanda Askell, Pamela Mishkin, Jack Clark,
      et~al.
    \newblock Learning transferable visual models from natural language
      supervision.
    \newblock In {\em ICML}, pages 8748--8763. PMLR, 2021.
    
    \bibitem{Ranftl2020}
    Ren\'{e} Ranftl, Katrin Lasinger, David Hafner, Konrad Schindler, and Vladlen
      Koltun.
    \newblock Towards robust monocular depth estimation: Mixing datasets for
      zero-shot cross-dataset transfer.
    \newblock {\em IEEE TPAMI}, 2020.
    
    \bibitem{SensoyKK18}
    Murat Sensoy, Lance Kaplan, and Melih Kandemir.
    \newblock Evidential deep learning to quantify classification uncertainty.
    \newblock In {\em NeurIPS}, volume~31, 2018.
    
    \bibitem{shao2023tensor4d}
    Ruizhi Shao, Zerong Zheng, Hanzhang Tu, Boning Liu, Hongwen Zhang, and Yebin
      Liu.
    \newblock Tensor4d: Efficient neural 4d decomposition for high-fidelity dynamic
      reconstruction and rendering.
    \newblock In {\em CVPR}, pages 16632--16642, 2023.
    
    \bibitem{ShenAMR22}
    Jianxiong Shen, Antonio Agudo, Francesc Moreno{-}Noguer, and Adria Ruiz.
    \newblock Conditional-flow nerf: Accurate 3d modelling with reliable
      uncertainty quantification.
    \newblock In {\em ECCV}, volume 13663, pages 540--557, 2022.
    
    \bibitem{ShenRAM21}
    Jianxiong Shen, Adria Ruiz, Antonio Agudo, and Francesc Moreno{-}Noguer.
    \newblock Stochastic neural radiance fields: Quantifying uncertainty in
      implicit 3d representations.
    \newblock In {\em 3DV}, pages 972--981, 2021.
    
    \bibitem{sun2022direct}
    Cheng Sun, Min Sun, and Hwann-Tzong Chen.
    \newblock Direct voxel grid optimization: Super-fast convergence for radiance
      fields reconstruction.
    \newblock In {\em CVPR}, pages 5459--5469, 2022.
    
    \bibitem{sun2022improved}
    Cheng Sun, Min Sun, and Hwann-Tzong Chen.
    \newblock Improved direct voxel grid optimization for radiance fields
      reconstruction.
    \newblock {\em arXiv preprint arXiv:2206.05085}, 2022.
    
    \bibitem{ijcai2023p157}
    Jiakai Sun, Zhanjie Zhang, Jiafu Chen, Guangyuan Li, Boyan Ji, Lei Zhao, and
      Wei Xing.
    \newblock Vgos: Voxel grid optimization for view synthesis from sparse inputs.
    \newblock In {\em IJCAI}, pages 1414--1422, 8 2023.
    
    \bibitem{tan2022volux}
    Feitong Tan, Sean Fanello, Abhimitra Meka, Sergio Orts-Escolano, Danhang Tang,
      Rohit Pandey, Jonathan Taylor, Ping Tan, and Yinda Zhang.
    \newblock Volux-gan: A generative model for 3d face synthesis with hdri
      relighting.
    \newblock In {\em ACM TOG}, pages 1--9, 2022.
    
    \bibitem{tancik2022block}
    Matthew Tancik, Vincent Casser, Xinchen Yan, Sabeek Pradhan, Ben Mildenhall,
      Pratul~P Srinivasan, Jonathan~T Barron, and Henrik Kretzschmar.
    \newblock Block-nerf: Scalable large scene neural view synthesis.
    \newblock In {\em CVPR}, pages 8248--8258, 2022.
    
    \bibitem{tancik2020fourier}
    Matthew Tancik, Pratul Srinivasan, Ben Mildenhall, Sara Fridovich-Keil, Nithin
      Raghavan, Utkarsh Singhal, Ravi Ramamoorthi, Jonathan Barron, and Ren Ng.
    \newblock Fourier features let networks learn high frequency functions in low
      dimensional domains.
    \newblock In {\em NeurIPS}, volume~33, pages 7537--7547, 2020.
    
    \bibitem{WangBSS04}
    Zhou Wang, Alan~C. Bovik, Hamid~R. Sheikh, and Eero~P. Simoncelli.
    \newblock Image quality assessment: from error visibility to structural
      similarity.
    \newblock {\em IEEE TIP}, 13(4):600--612, 2004.
    
    \bibitem{wynn-2023-diffusionerf}
    Jamie Wynn and Daniyar Turmukhambetov.
    \newblock {DiffusioNeRF: Regularizing Neural Radiance Fields with Denoising
      Diffusion Models}.
    \newblock In {\em CVPR}, 2023.
    
    \bibitem{YanLQCF23}
    Dongyu Yan, Jianheng Liu, Fengyu Quan, Haoyao Chen, and Mengmeng Fu.
    \newblock Active implicit object reconstruction using uncertainty-guided
      next-best-view optimization.
    \newblock {\em {IEEE} Robotics Autom. Lett.}, 8(10):6395--6402, 2023.
    
    \bibitem{yang2023freenerf}
    Jiawei Yang, Marco Pavone, and Yue Wang.
    \newblock Freenerf: Improving few-shot neural rendering with free frequency
      regularization.
    \newblock In {\em CVPR}, pages 8254--8263, 2023.
    
    \bibitem{yu2021pixelnerf}
    Alex Yu, Vickie Ye, Matthew Tancik, and Angjoo Kanazawa.
    \newblock pixelnerf: Neural radiance fields from one or few images.
    \newblock In {\em CVPR}, pages 4578--4587, 2021.
    
    \bibitem{yuan2022nerf}
    Yu-Jie Yuan, Yang-Tian Sun, Yu-Kun Lai, Yuewen Ma, Rongfei Jia, and Lin Gao.
    \newblock Nerf-editing: geometry editing of neural radiance fields.
    \newblock In {\em CVPR}, pages 18353--18364, 2022.
    
    \bibitem{zhang2022ray}
    Jian Zhang, Yuanqing Zhang, Huan Fu, Xiaowei Zhou, Bowen Cai, Jinchi Huang,
      Rongfei Jia, Binqiang Zhao, and Xing Tang.
    \newblock Ray priors through reprojection: Improving neural radiance fields for
      novel view extrapolation.
    \newblock In {\em CVPR}, pages 18376--18386, 2022.
    
    \bibitem{ZhangIESW18}
    Richard Zhang, Phillip Isola, Alexei~A. Efros, Eli Shechtman, and Oliver Wang.
    \newblock The unreasonable effectiveness of deep features as a perceptual
      metric.
    \newblock In {\em CVPR}, pages 586--595, 2018.
    
    \bibitem{zheng2024learning}
    Chenxi Zheng, Bangzhen Liu, Xuemiao Xu, Huaidong Zhang, and Shengfeng He.
    \newblock Learning an interpretable stylized subspace for 3d-aware animatable
      artforms.
    \newblock {\em IEEE TVCG}, 2024.
    
    \bibitem{zheng2023my}
    Chenxi Zheng, Bangzhen Liu, Huaidong Zhang, Xuemiao Xu, and Shengfeng He.
    \newblock Where is my spot? few-shot image generation via latent subspace
      optimization.
    \newblock In {\em CVPR}, pages 3272--3281, 2023.
    
    \bibitem{zhou2023single}
    Yang Zhou, Hanjie Wu, Wenxi Liu, Zheng Xiong, Jing Qin, and Shengfeng He.
    \newblock Single-view view synthesis with self-rectified pseudo-stereo.
    \newblock {\em IJCV}, 131(8):2032--2043, 2023.
    
\end{thebibliography}
\end{document}



\title{Learning with Unreliability: \\ Fast Few-shot Voxel Radiance Fields with Relative Geometric Consistency\\\emph{--Supplementary Materials--}}

\author{%
Yingjie Xu$^{1,2}$\footnotemark[1]\quad Bangzhen Liu$^{2}$\footnotemark[1]\quad Hao Tang$^{1,3}$ \quad Bailin Deng$^{4}$\quad Shengfeng He$^{1}$\footnotemark[2] \\
$^{1}$Singapore Management University \quad $^{2}$South China University of Technology \\
$^{3}$Nanjing University of Science and Technology $^{4}$Cardiff University\\
}

\maketitle
\renewcommand{\thefootnote}{\fnsymbol{footnote}}
\footnotetext[1]{The first two authors contributed equally.}
\footnotetext[2]{Corresponding author (\emph{shengfenghe@smu.edu.sg}).}

This supplementary material includes visualizations of more experimental results, further analysis on the selection of different warping angles, and extra implementation details. Besides, we also record a video demo, which gives an explanation of the overall framework and the visualizations.

\section{Comparisons on More Scenes}

\subsection{Realistic Synthetic 360°}

\begin{figure*}
    \centering
    \begin{subfigure}{.17\linewidth}
    \centering
            \includegraphics[width=\linewidth]{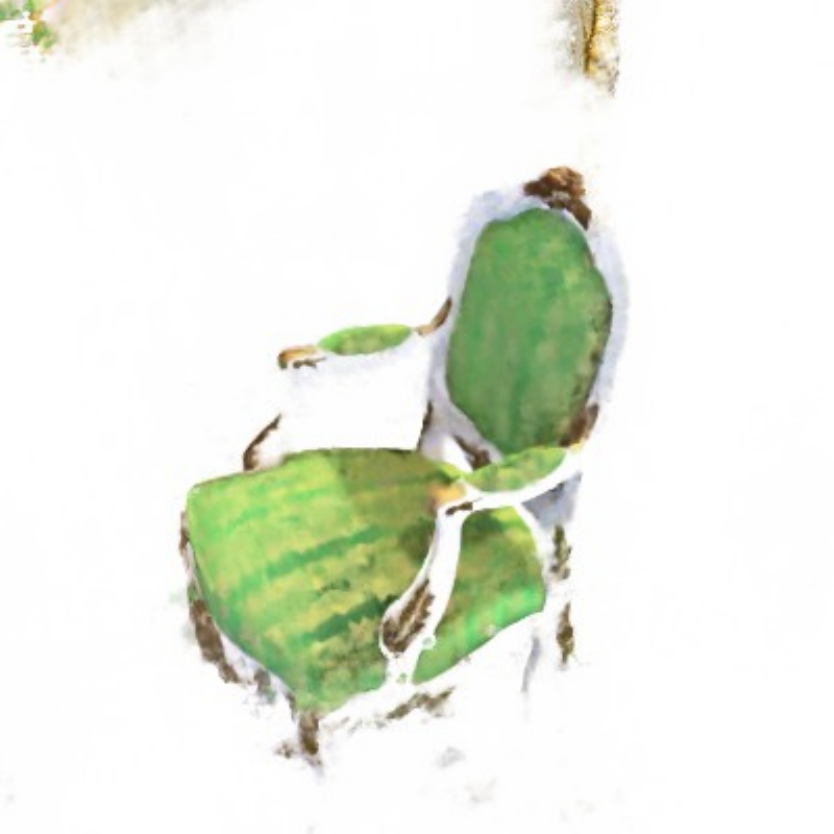} \\
            \includegraphics[width=\linewidth]{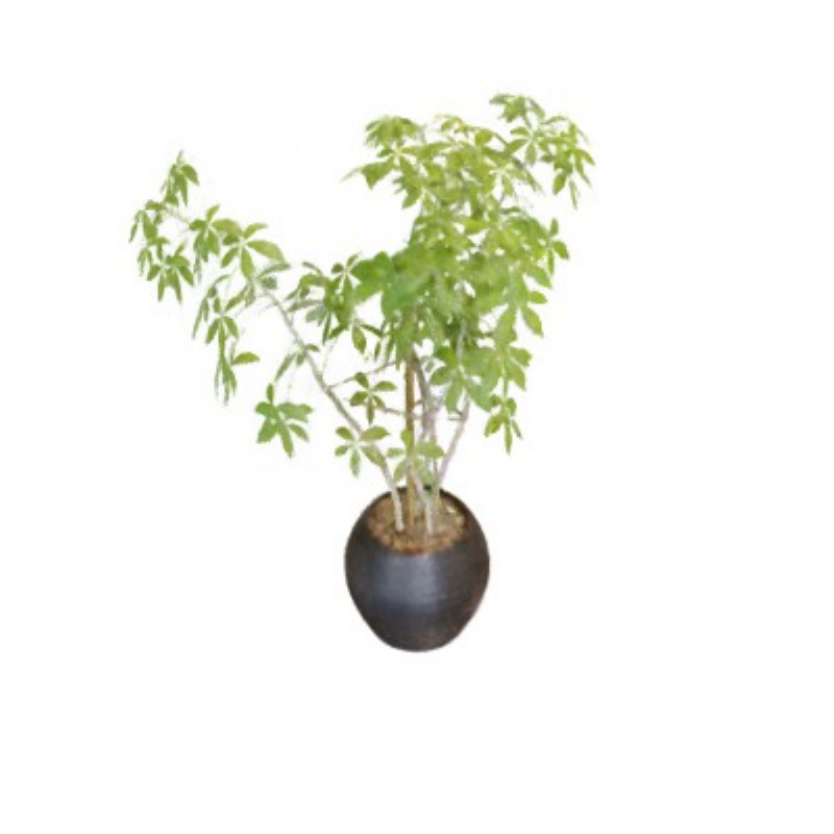} \\
            \includegraphics[width=\linewidth]{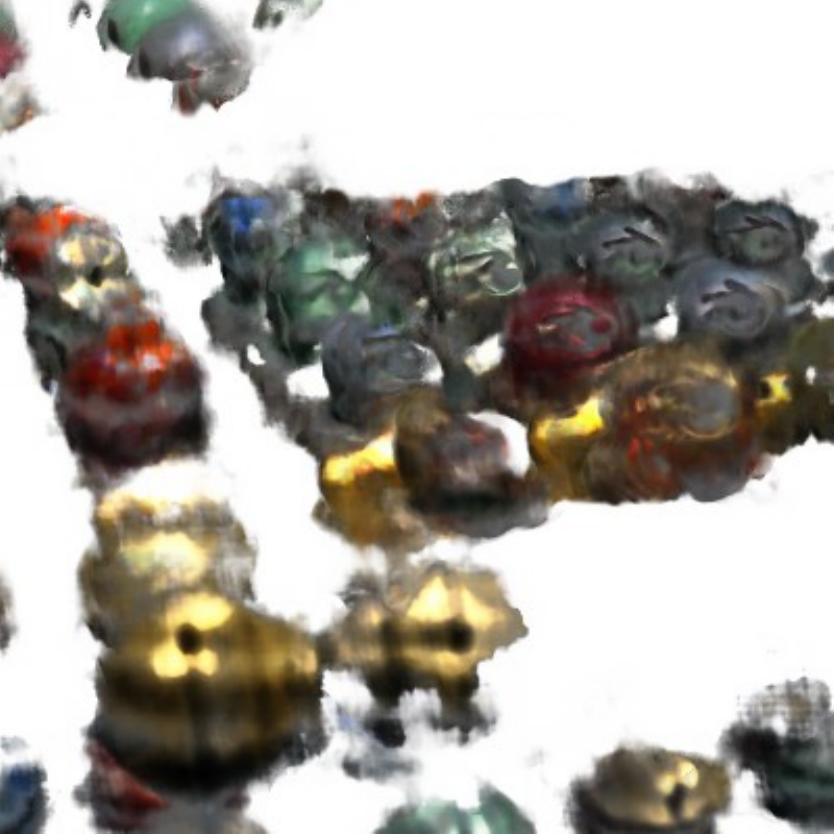}\\
        \caption*{\footnotesize{Diet-NeRF~\cite{jain2021putting}}}
    \end{subfigure}
    \hspace{-1.7mm}
    \begin{subfigure}{.17\linewidth}
    \centering
            \includegraphics[width=\linewidth]{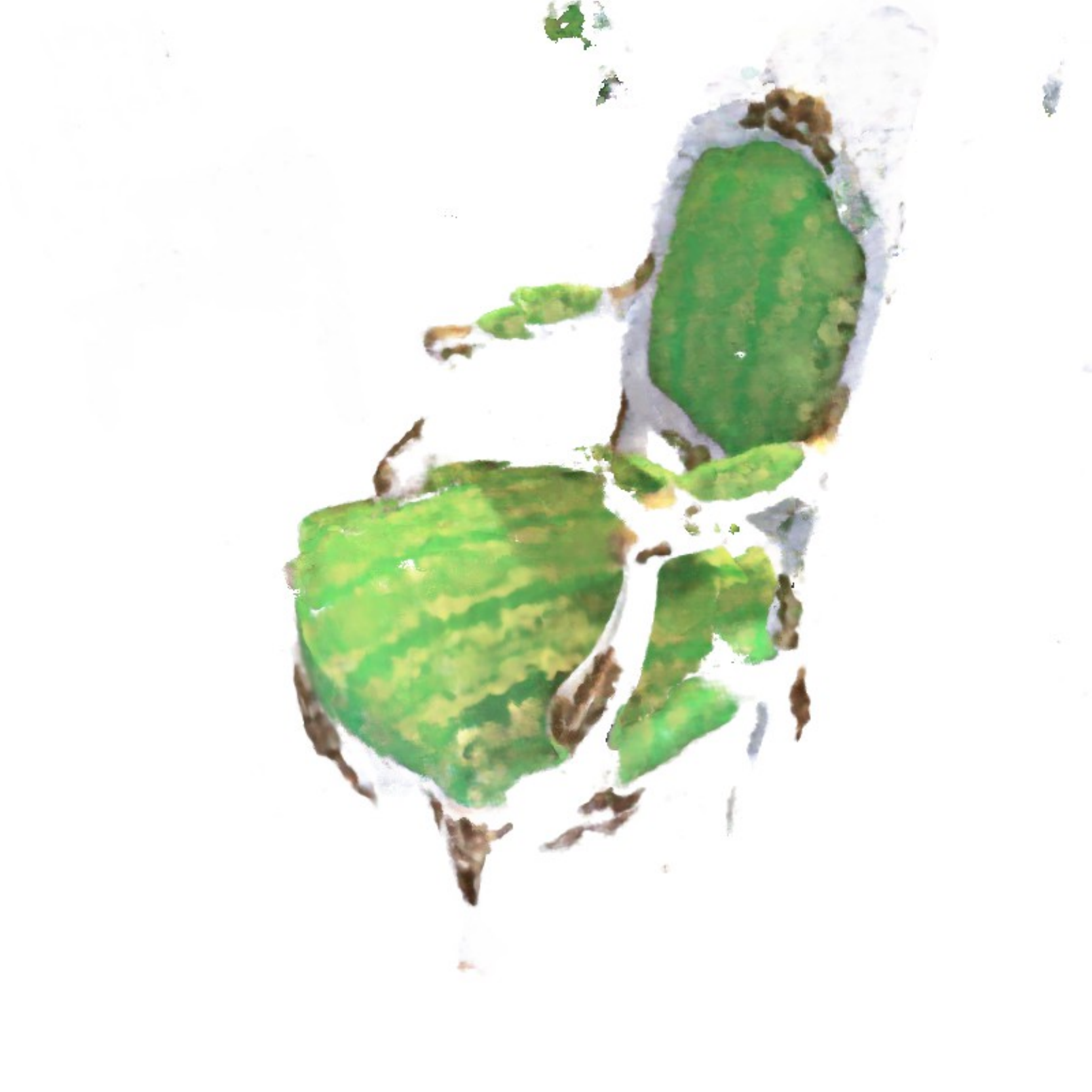}\\
            \includegraphics[width=\linewidth]{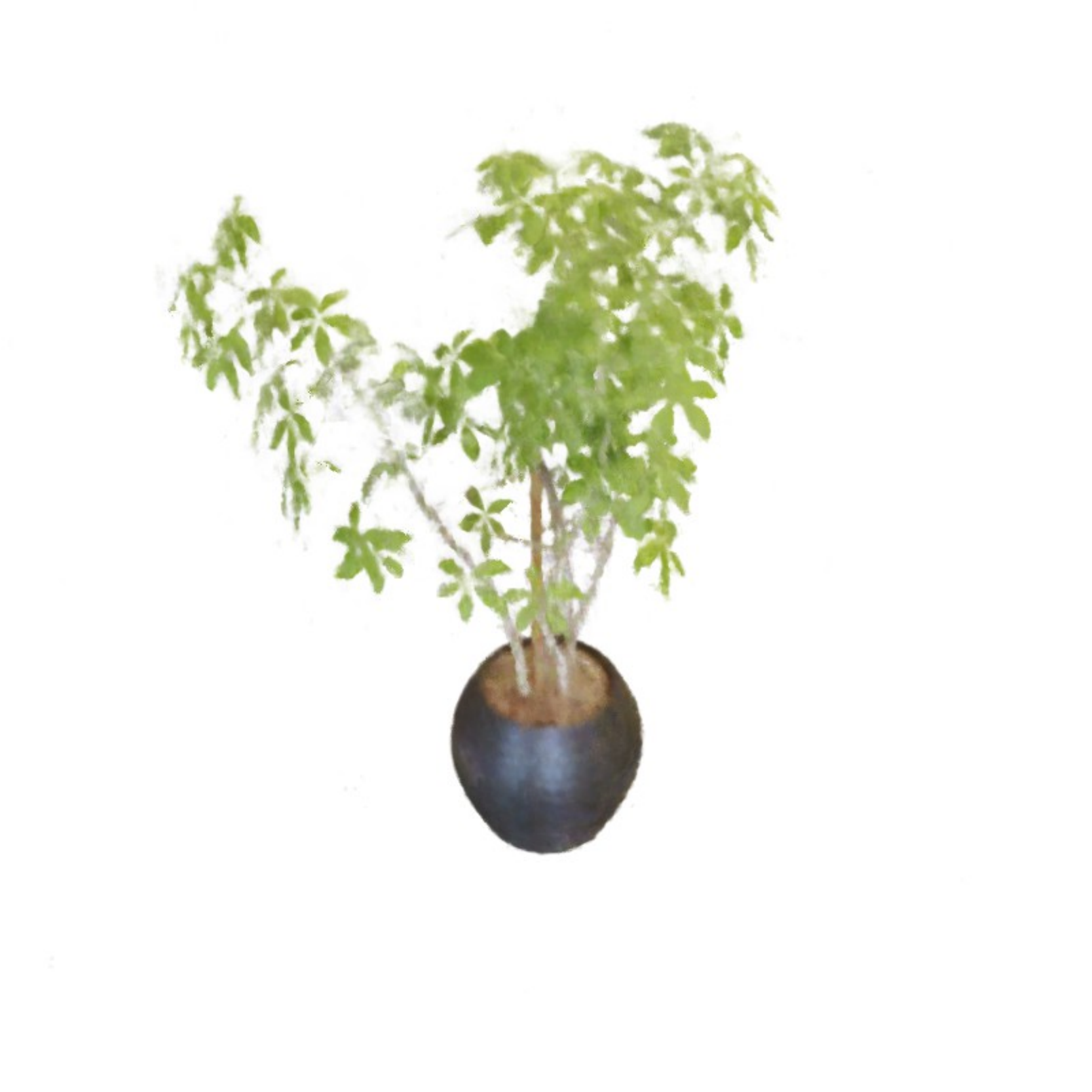}\\
            \includegraphics[width=\linewidth]{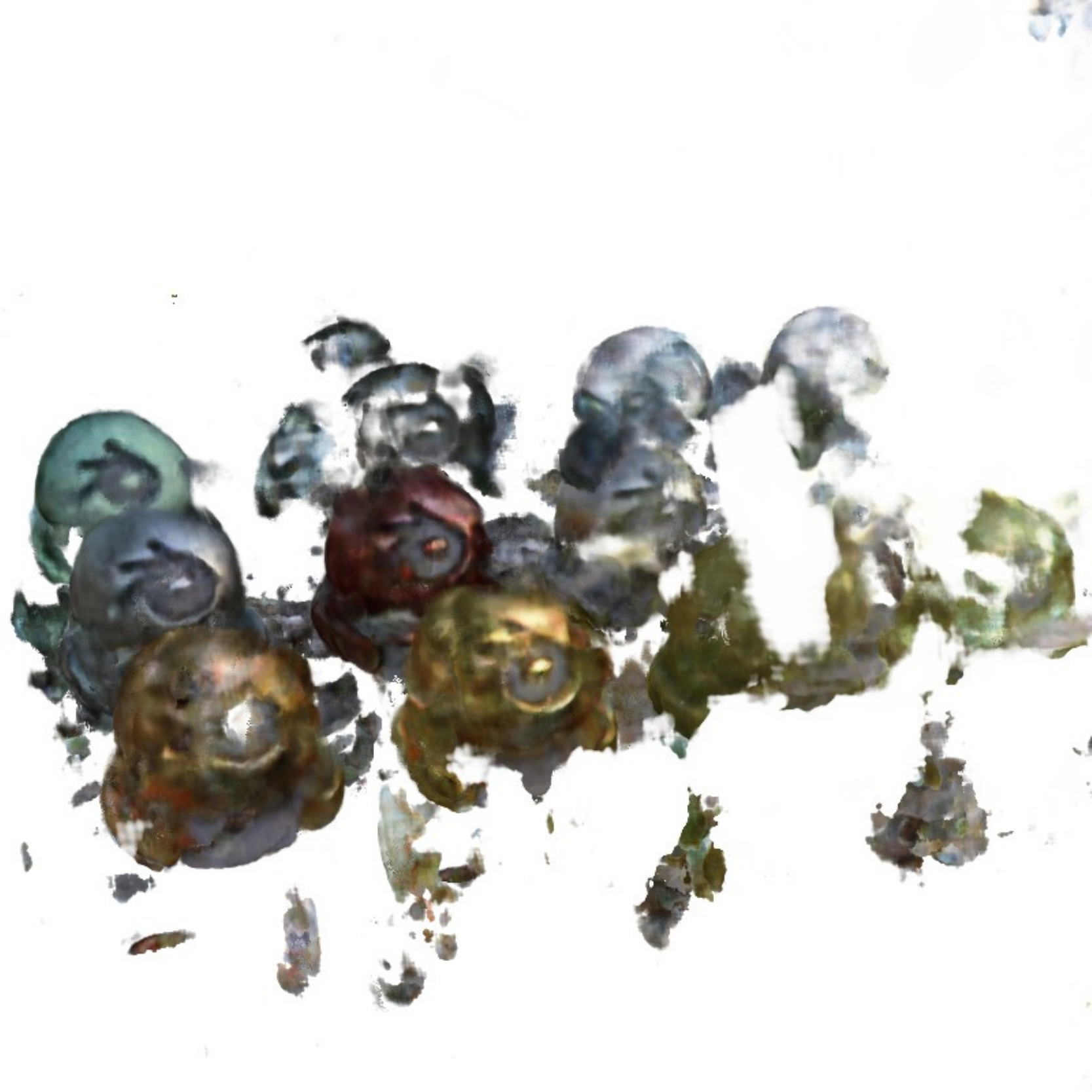}\\
        \caption*{\footnotesize{InfoNeRF~\cite{kim2022infonerf}}\label{edge_a}}
    \end{subfigure}
    \hspace{-1.7mm}
    \begin{subfigure}{.17\linewidth}
    \centering
            \includegraphics[width=\linewidth]{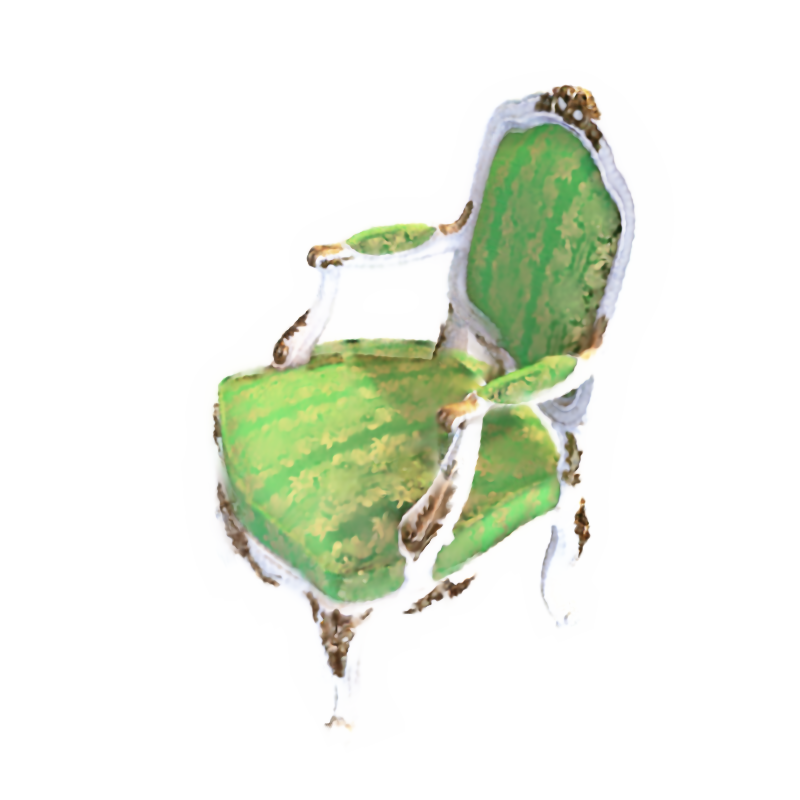}\\
            \includegraphics[width=\linewidth]{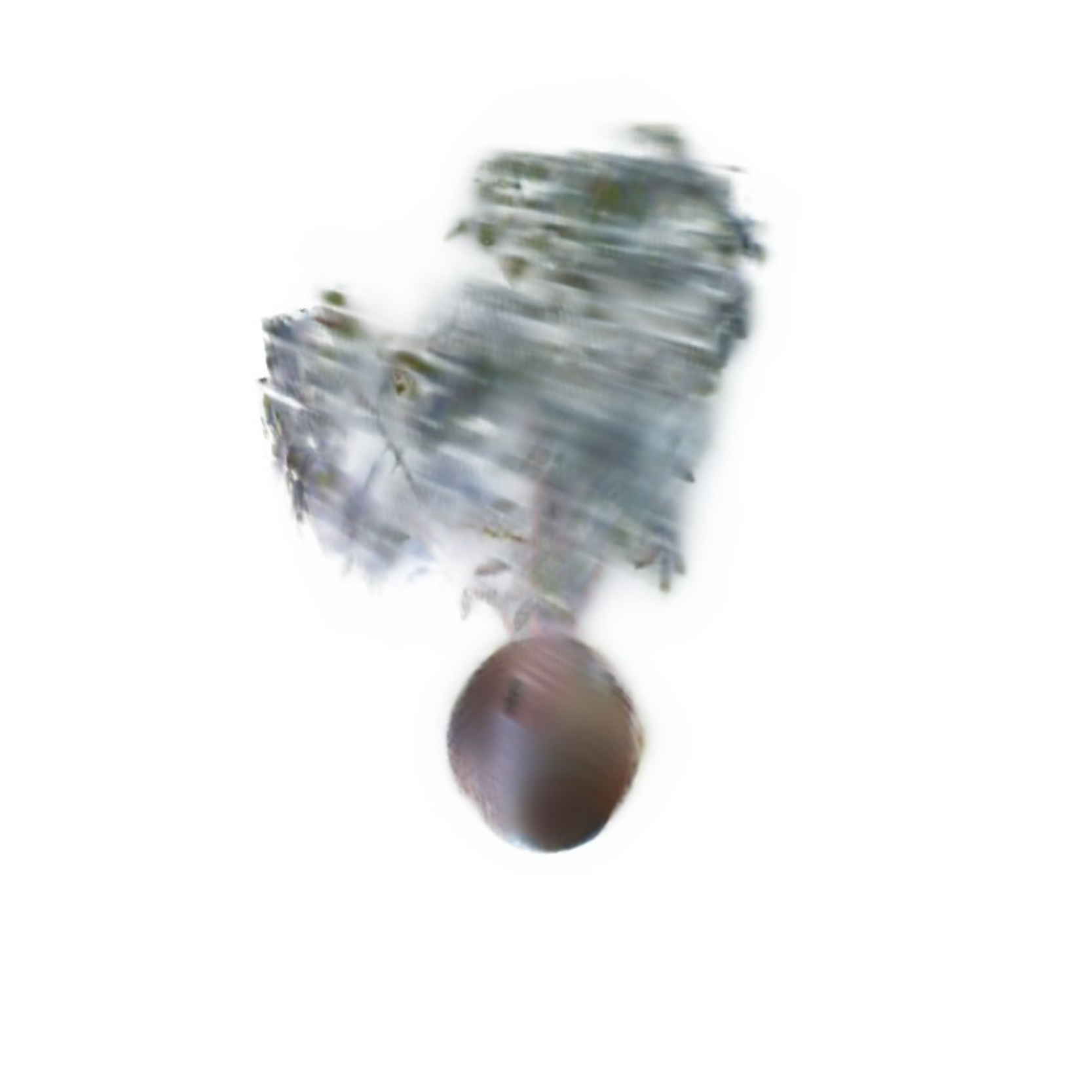}\\
            \includegraphics[width=\linewidth]{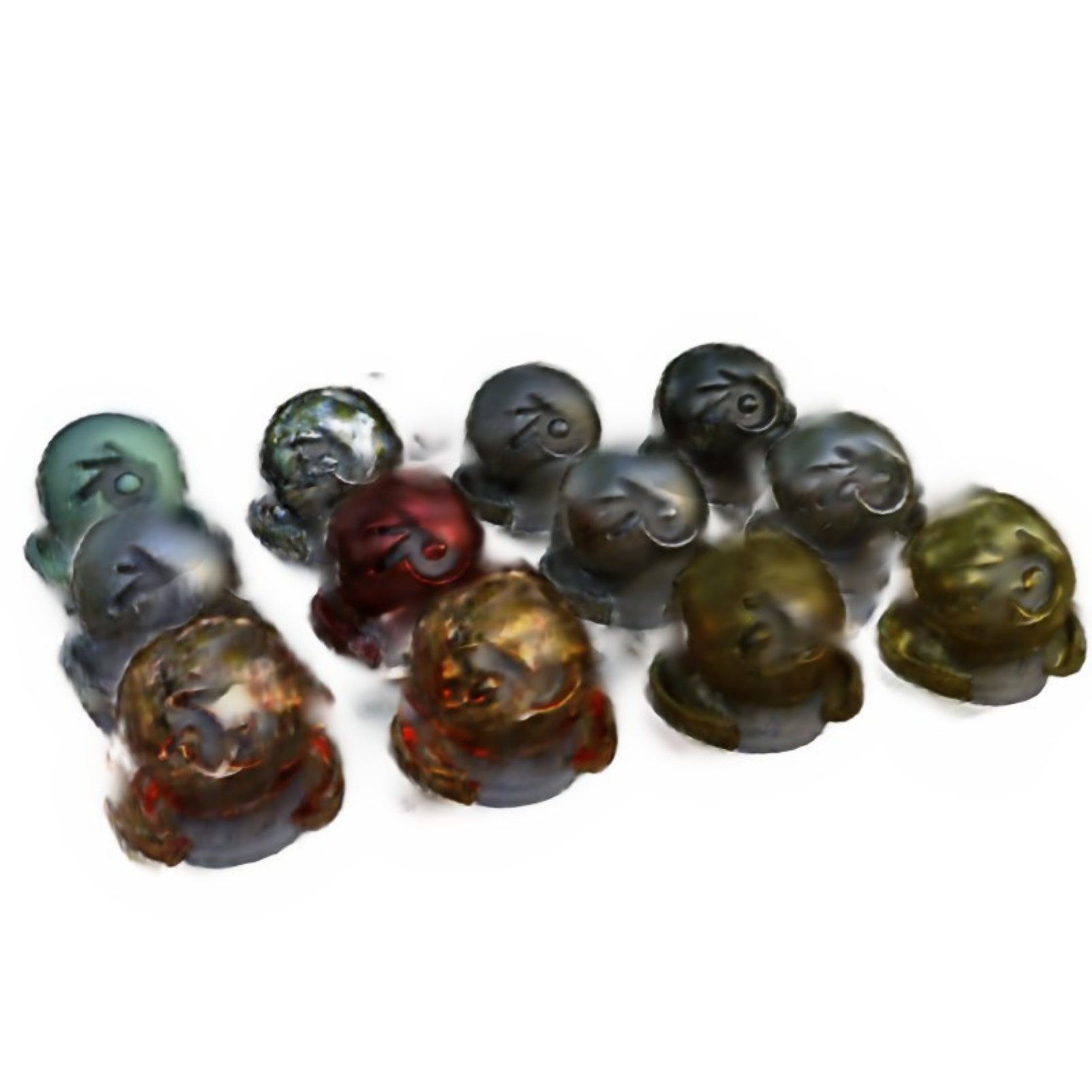}\\
        \caption*{\footnotesize{VGOS~\cite{ijcai2023p157}}\label{edge_a}}
    \end{subfigure}
    \hspace{-1.7mm}
    \begin{subfigure}{.17\linewidth}
        \centering
        \includegraphics[width=\linewidth]{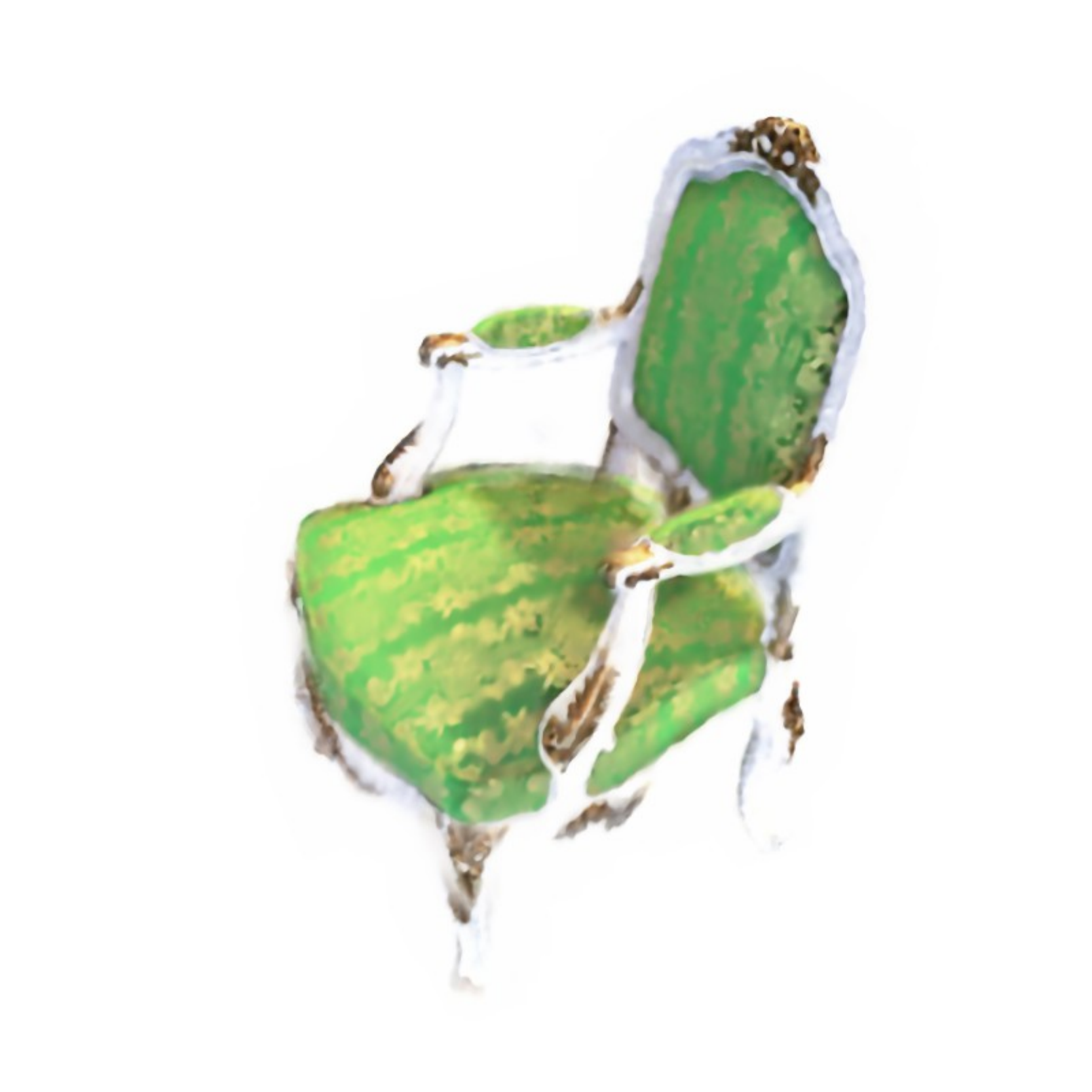}\\
        \includegraphics[width=\linewidth]{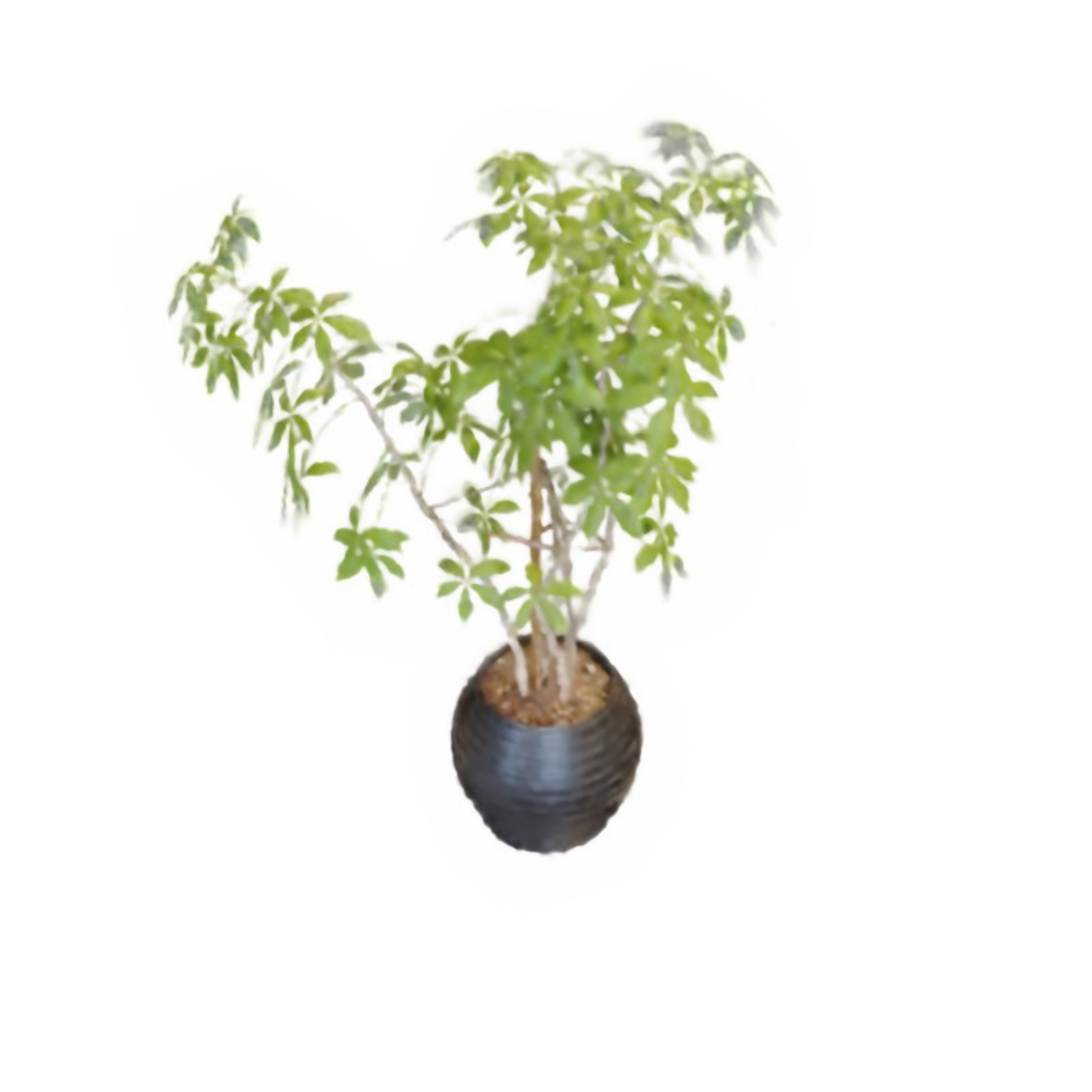}\\
        \includegraphics[width=\linewidth]{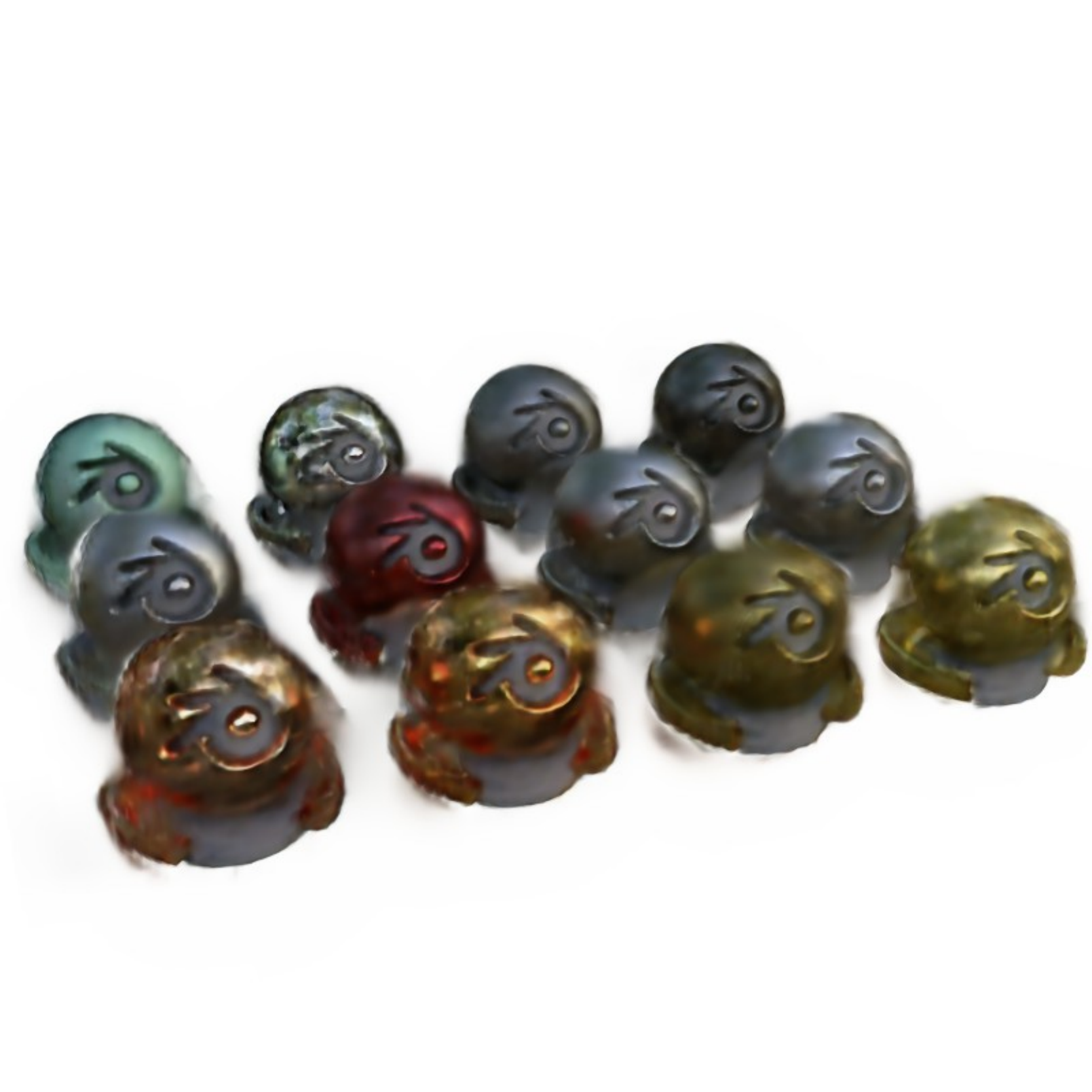}\\
        \caption*{\footnotesize{ReVoRF (Ours)}\label{edge_b}}
    \end{subfigure}
    \hspace{-1.7mm}
    \begin{subfigure}{.17\linewidth}
    \centering
    \includegraphics[width=\linewidth]{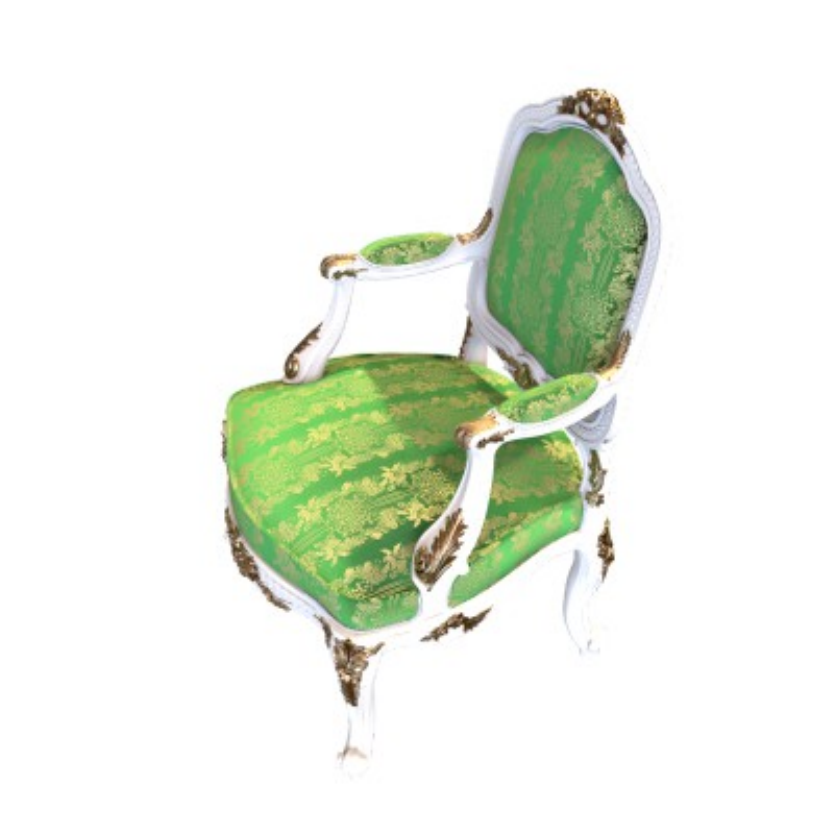}\\
    \includegraphics[width=\linewidth]{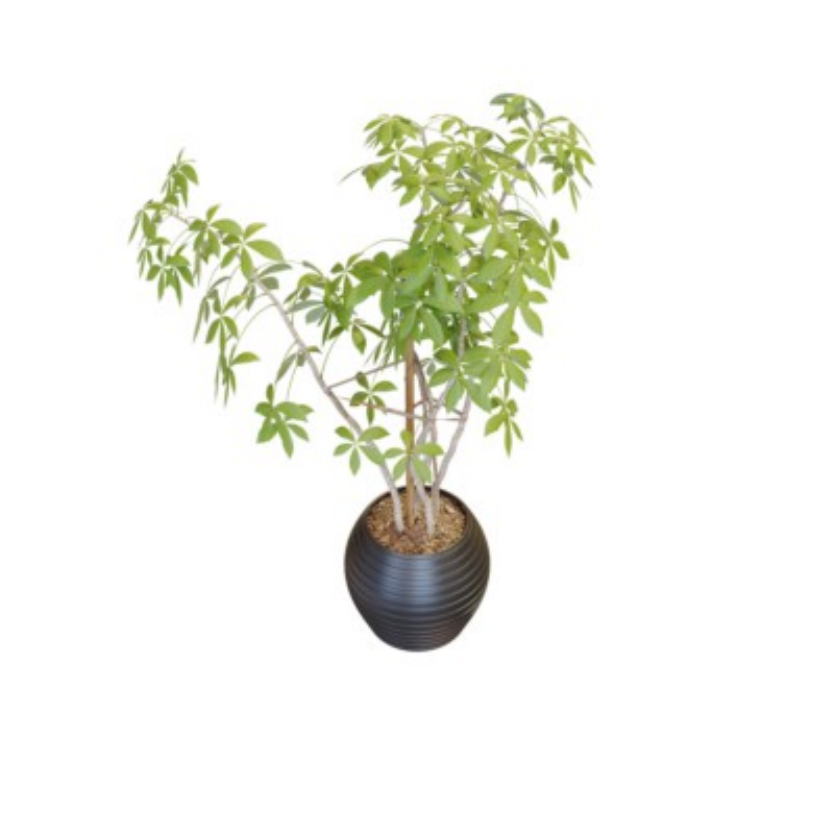}\\
    \includegraphics[width=\linewidth]{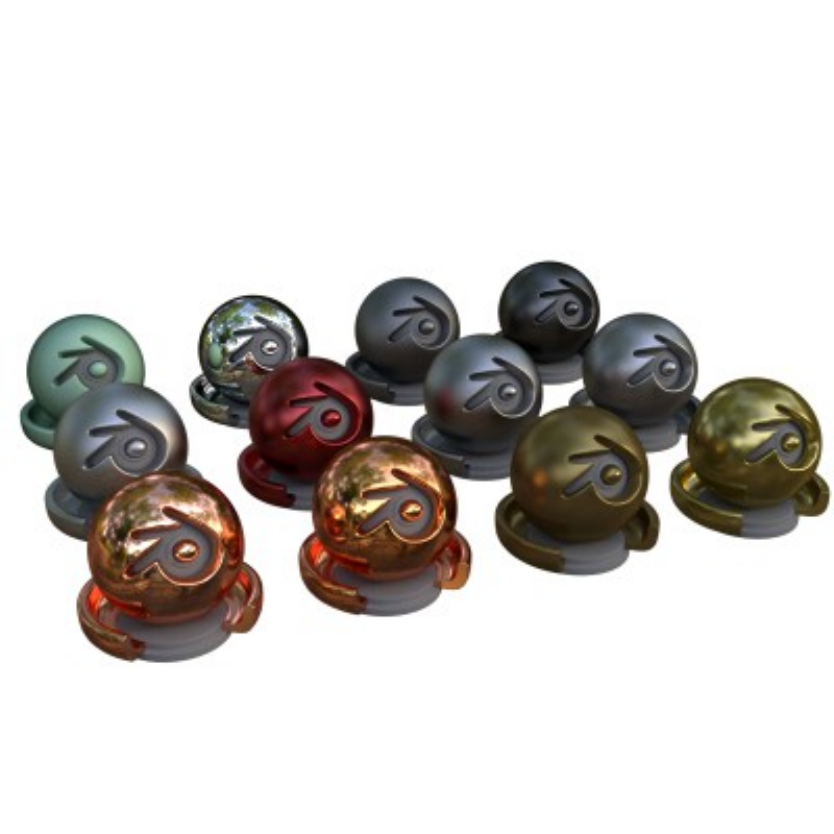}\\
        \caption*{\footnotesize{GT}\label{edge_ours}}
    \end{subfigure}
    \caption{Comparisons on chair, ficus, and materials (from the top to the bottom) of the Realistic Synthetic 360° dataset~\cite{MildenhallSTBRN20} in 4-views setting. ReVoRF enables more consistent reconstruction with detailed appearance. Please zoom in for more details.}
    \label{fig:morecomparisons}

\end{figure*}

We provide more scene reconstruction results of our proposed ReVoRF together with recent advanced methods~\cite{jain2021putting,ijcai2023p157,kim2022infonerf} on the Realistic Synthetic 360° dataset~\cite{MildenhallSTBRN20}. The visualization of the chair, ficus, and materials are shown in Fig.~\ref{fig:morecomparisons}. With fewer artifacts and finer texture details, ReVoRF exhibits a superior capacity for reconstructing both geometry and appearance details in these scenes than the compared methods. 
We also conduct an additional evaluation for separating $L_{rs}$ and $L_{ds}$ in Realistic Synthetic 360°. The results are shown in Table~\ref{tab:ablation_loss}. 
Note that the $L_{rs}$ used here does not consider the unreliability.
Our method outperforms these two variants.

\begin{table}[h]
    \centering
    \setlength{\tabcolsep}{0.01cm}{
    \resizebox{0.3\textwidth}{!}{
        \begin{tabular}{cccc}
        \hline
      & PSNR↑ & SSIM↑ & LPIPS↓ \\
      \hline
        VGOS+$L_{ds}$              & 17.33 & 0.779 & 0.241 \\
        VGOS+$L_{rs}$              & 18.27 & 0.814 & 0.210 \\
    \hline
        Ours            & \textbf{20.72} & \textbf{0.848} & \textbf{0.179} \\
        \hline
        \end{tabular}
    }
    \caption{Ablation on Realistic Synthetic 360° about $L_{rs}$ and $L_{ds}$}
    \vspace{-4mm}
    \label{tab:ablation_loss}
    }
\end{table}

\subsection{LLFF}

\begin{figure*}
    \centering
    \begin{subfigure}{.24\linewidth}
    \centering
            \includegraphics[width=\linewidth]{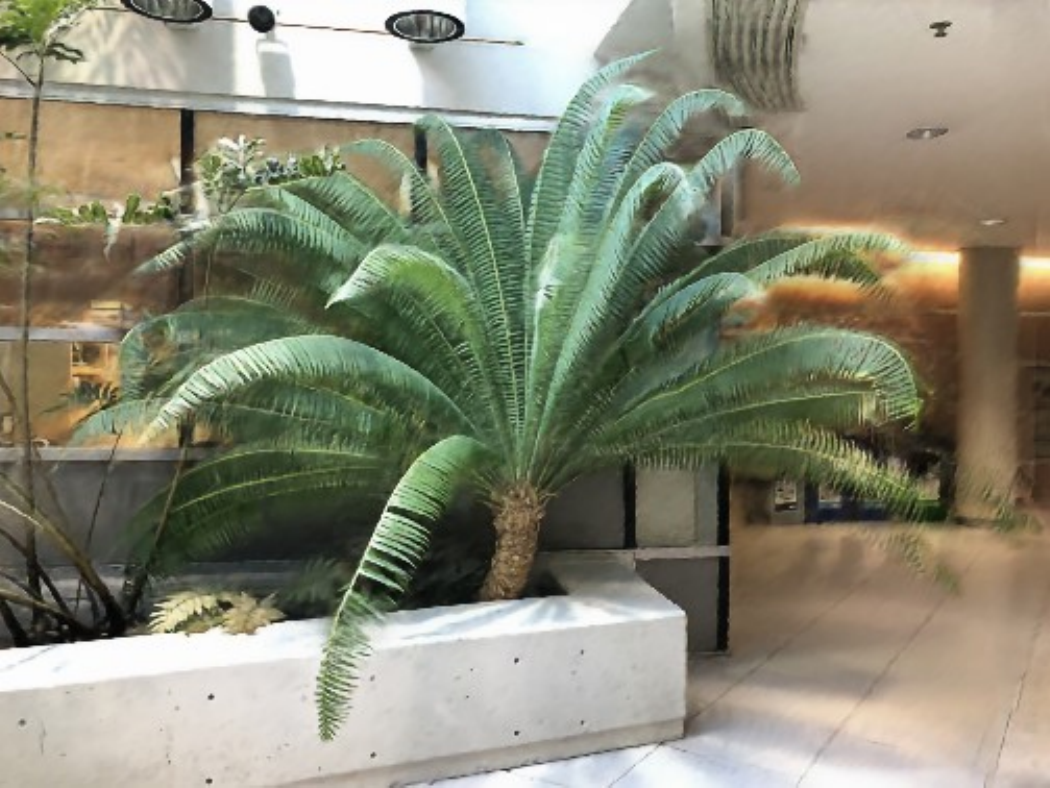} \\
            \includegraphics[width=\linewidth]{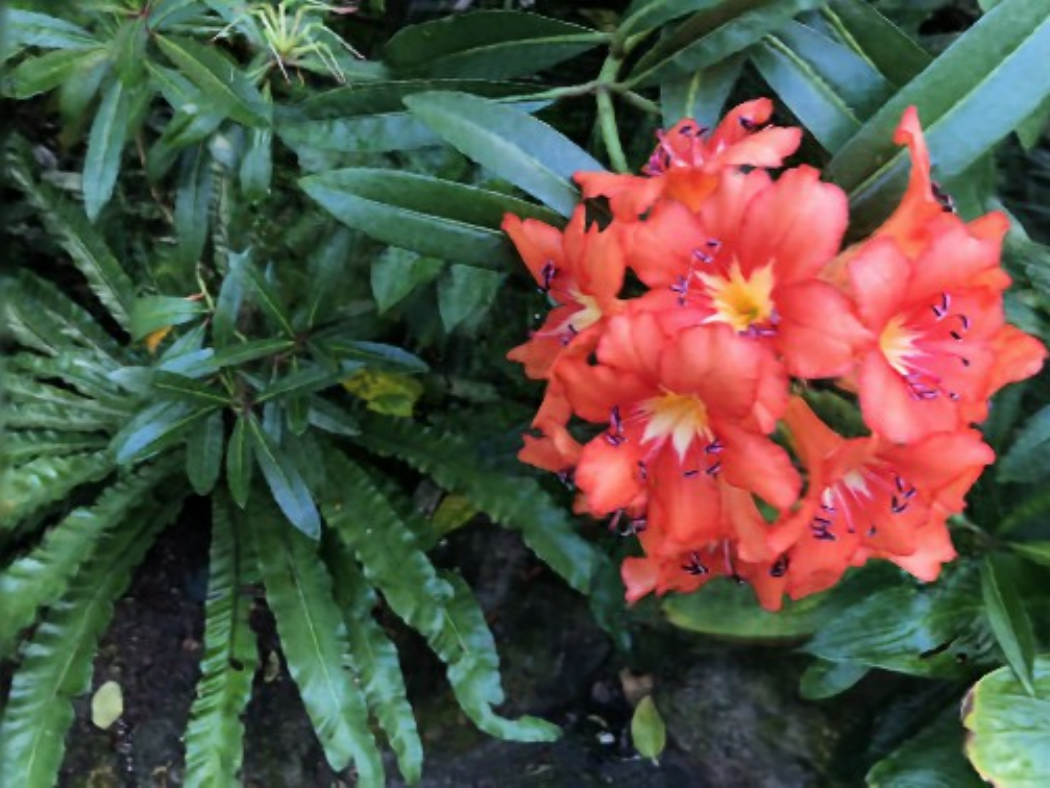} \\
            \includegraphics[width=\linewidth]{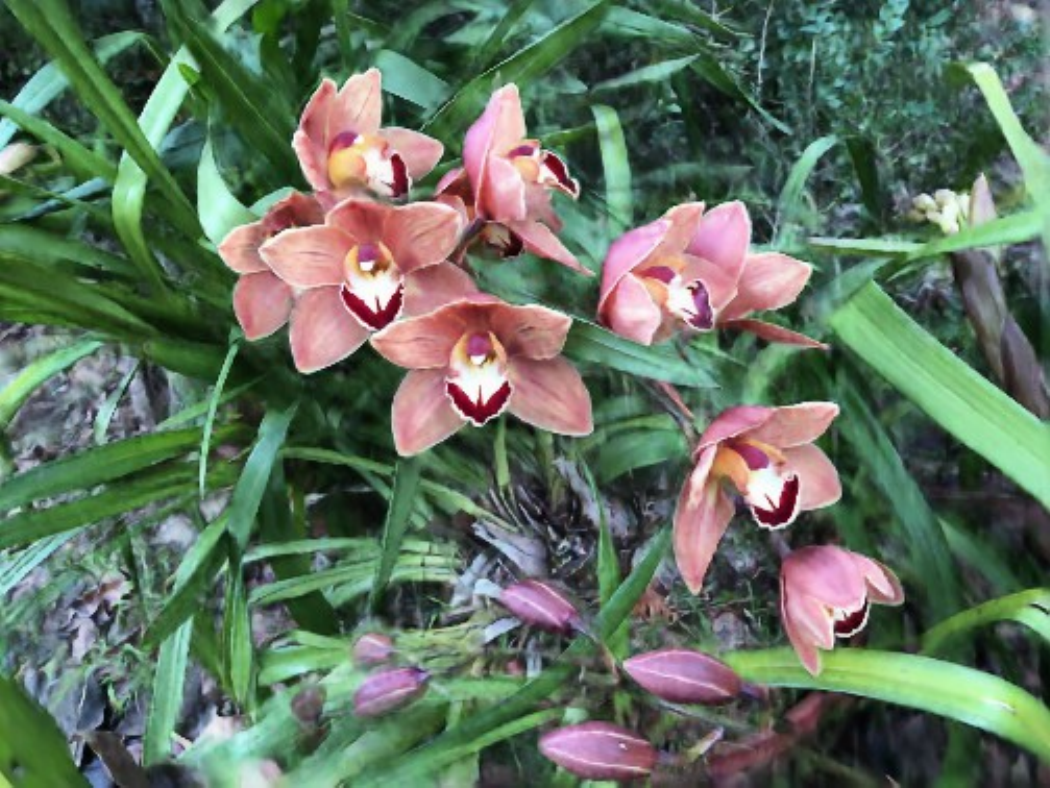}\\
        \caption{\footnotesize{VGOS RGB~\cite{jain2021putting}}\label{llff_vgos}}
    \end{subfigure}
    \hspace{-1.7mm}
    \begin{subfigure}{.24\linewidth}
    \centering
        \includegraphics[width=\linewidth]{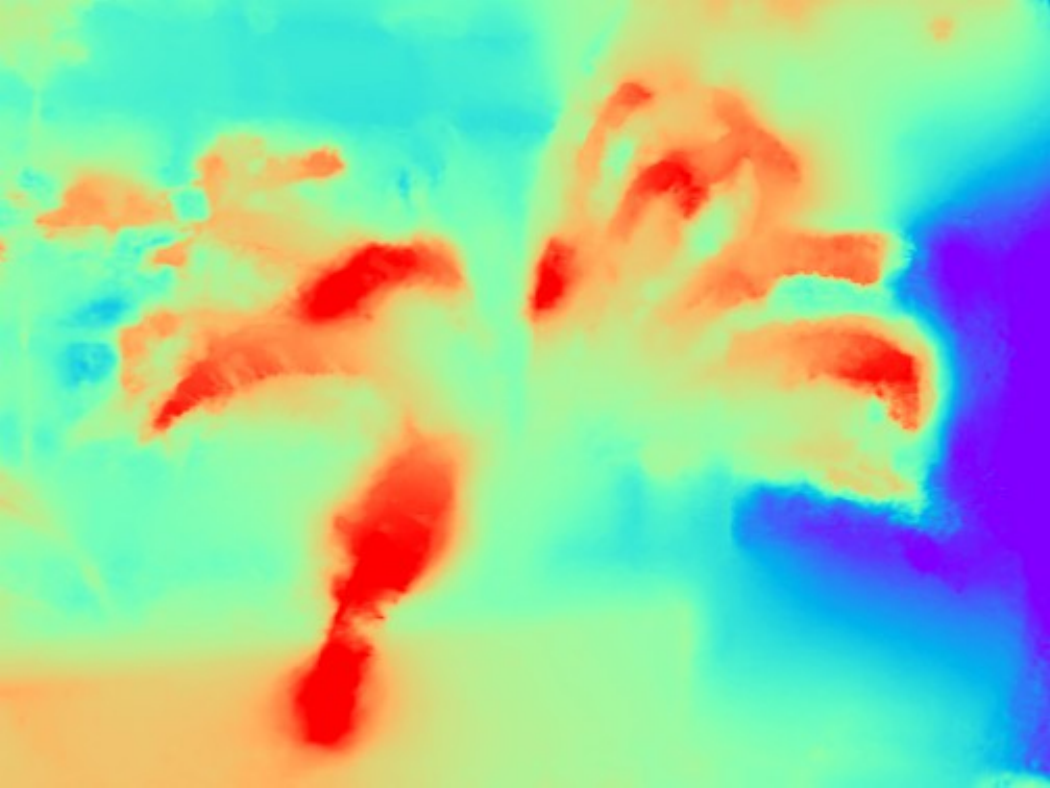} \\
        \includegraphics[width=\linewidth]{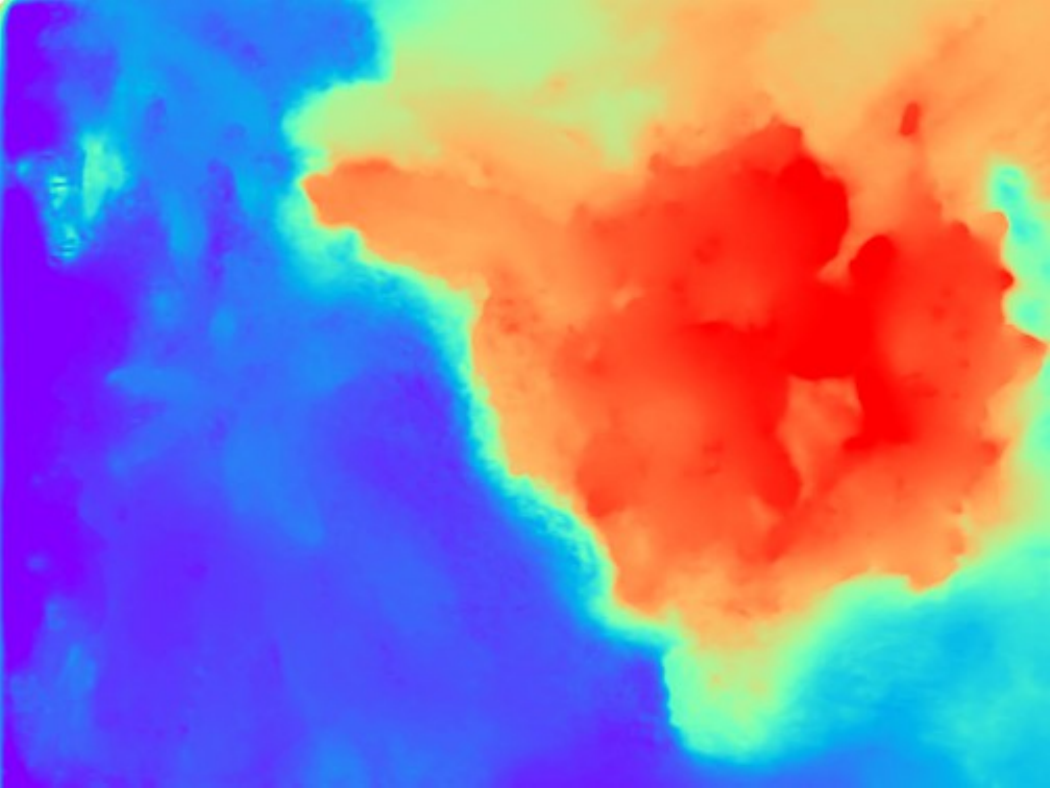} \\
        \includegraphics[width=\linewidth]{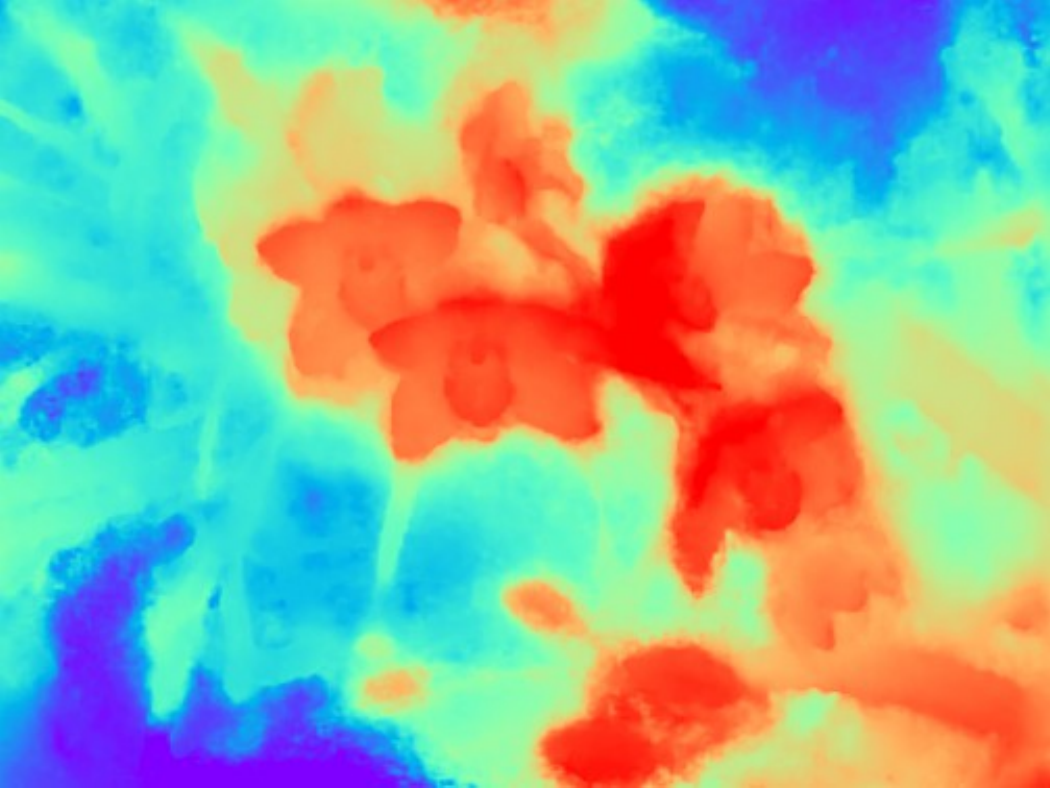}\\
        \caption{\footnotesize{VGOS depth~\cite{kim2022infonerf}}\label{llff_vgos_depth}}
    \end{subfigure}
    \hspace{-1.7mm}
    \begin{subfigure}{.24\linewidth}
    \centering
    \includegraphics[width=\linewidth]{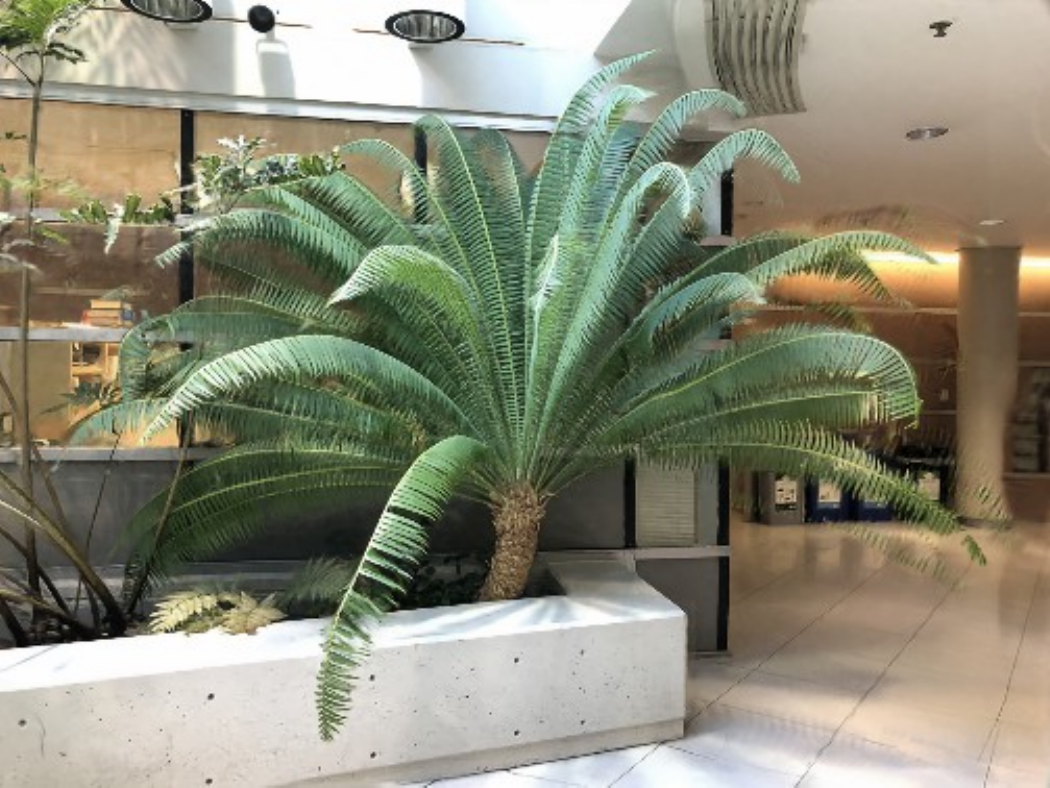} \\
    \includegraphics[width=\linewidth]{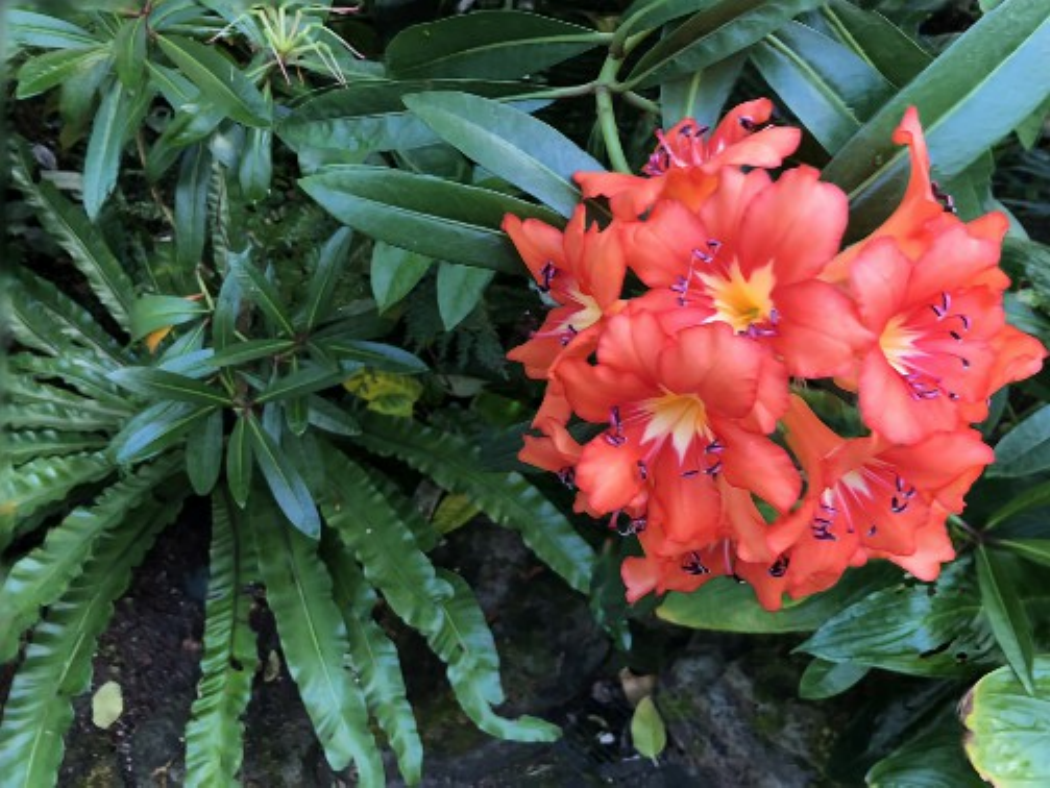} \\
    \includegraphics[width=\linewidth]{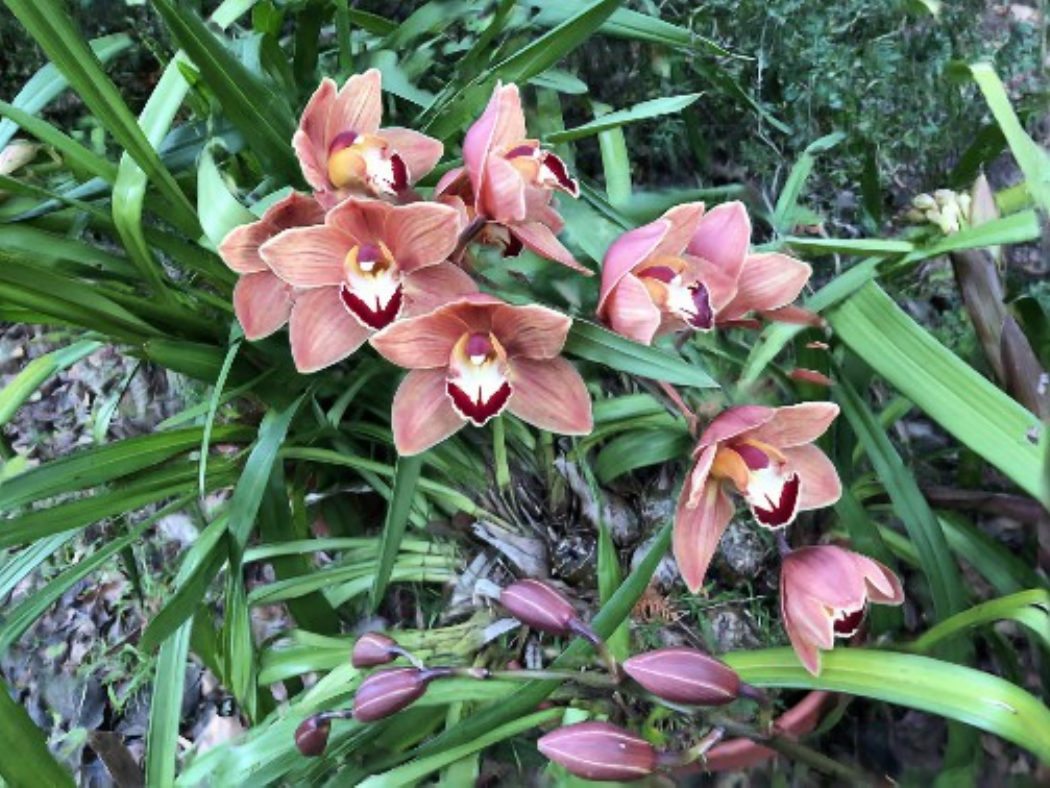}\\
        \caption{\footnotesize{ReVoRF (Ours) RGB}\label{llff_ours}}
    \end{subfigure}
    \hspace{-1.7mm}
    \begin{subfigure}{.24\linewidth}
        \centering
        \includegraphics[width=\linewidth]{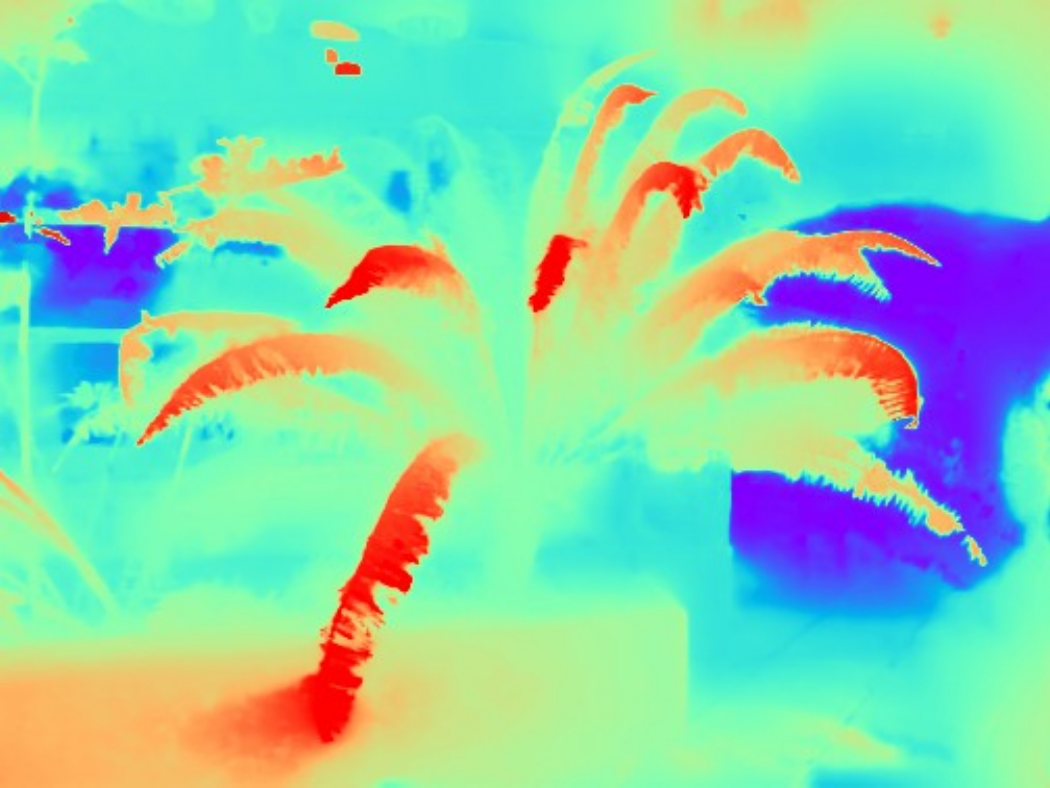} \\
        \includegraphics[width=\linewidth]{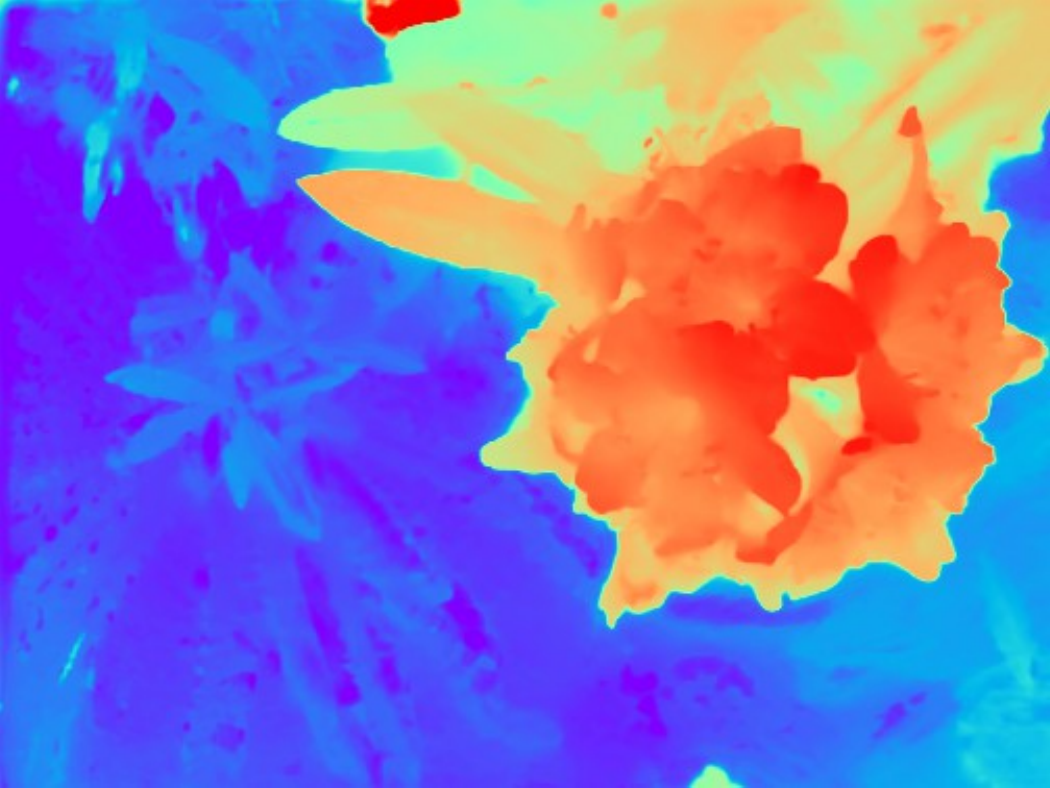} \\
        \includegraphics[width=\linewidth]{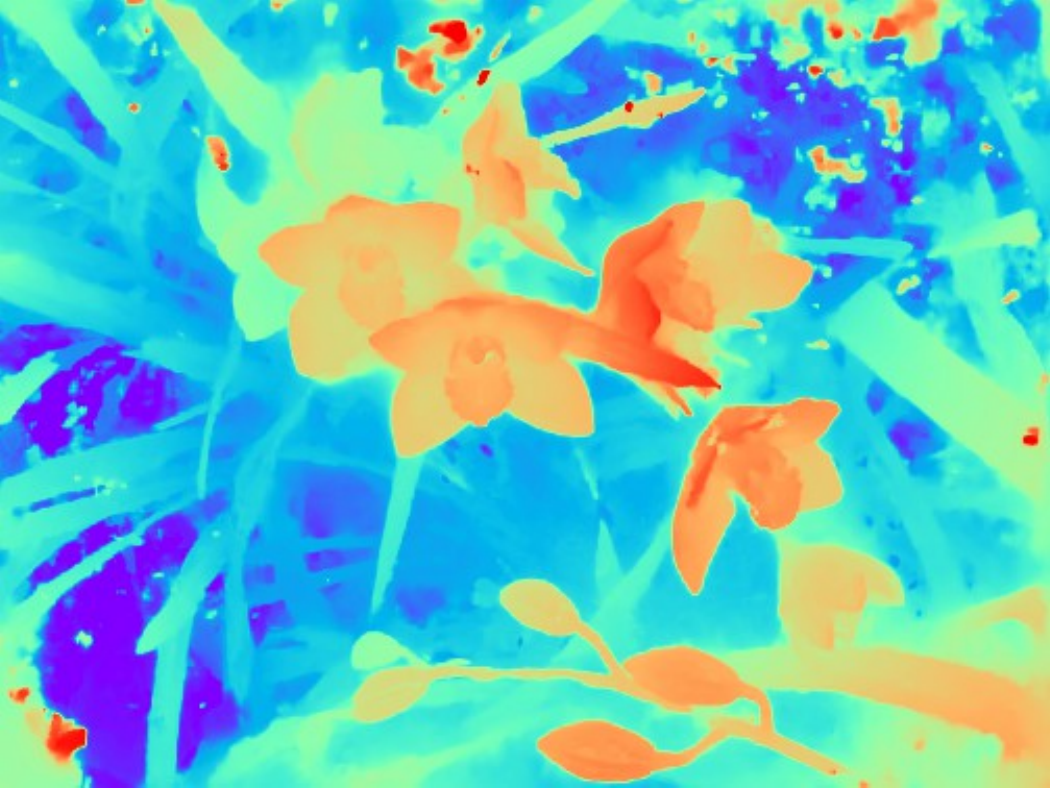}\\
        \caption{\footnotesize{ReVoRF (Ours) depth}\label{llff_ours_depth}}
    \end{subfigure}
    \caption{Comparisons on LLFF dataset~\cite{MildenhallSCKRN19} in 3-views setting. ReVoRF exhibits greater clarity in depth and improved geometric shapes.}
    \label{fig:morellff}

\end{figure*}

We provide the visualizations of rendering images and corresponding depth map of our ReVoRF and the state-of-the-art voxel-based few-shot nerf method VGOS on the LLFF dataset~\cite{MildenhallSCKRN19}. As shown in Fig.~\ref{llff_vgos} and Fig.~\ref{llff_ours}, our method achieves better rendering quality, in terms of the clearer boundary of objects and less blurring. Compared to VGOS, our method also preserves better geometric consistency, which has finer reconstructed depth (Fig.~\ref{llff_vgos_depth} vs. Fig.~\ref{llff_ours_depth}).

\section{Ablations on the Choices of Warping Degree}

We investigate the impact of warping angles $\gamma$ on rendering quality by reconstructing a Lego scene from the Realistic Synthetic 360° dataset. For each selected $\gamma$, we randomly vary the values of pitch angle $\mathbf{\theta}$ and azimuth angle $\mathbf{\phi}$ in a range of $\left[\gamma-5, \gamma\right]$. As illustrated in Fig.~\ref{fig:warptables}, when $\gamma$ is either too large or too small, the rendering quality deteriorates. On the one hand, a small $\gamma$ may not be able to provide sufficient multi-view information due to the slight variance of views. On the other hand, increasing the degree of $\gamma$ could encounter more erroneously warped regions at the very beginning of training. Therefore, a moderate deformation angle yields the optimal rendering result.

\begin{center}
    \includegraphics[width=0.5\textwidth]{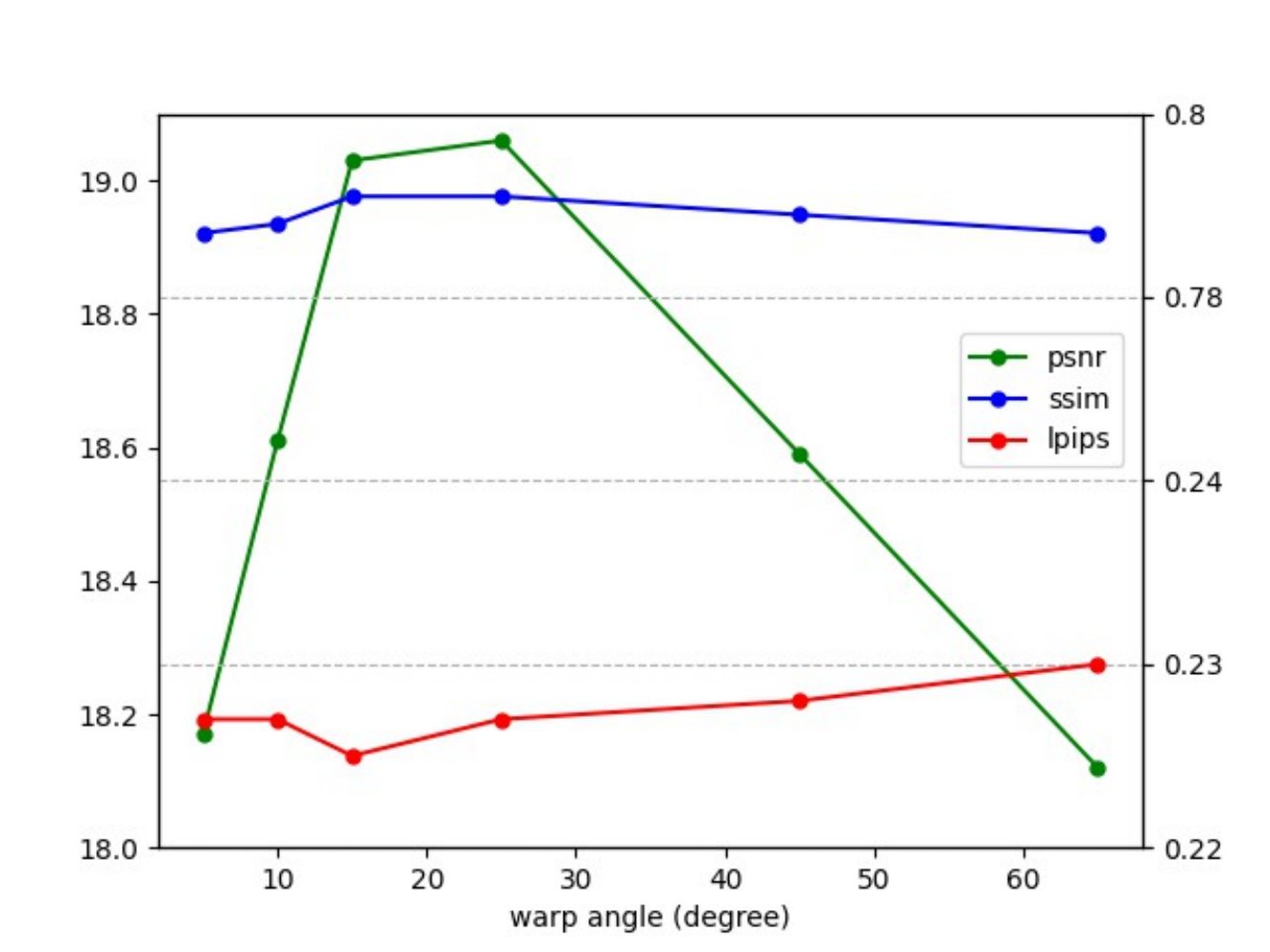}\\
    \captionof{figure}{The ablation of warping angles $\gamma$ on the Lego scene of Realistic Synthetic 360°. The horizontal axis in the graph represents the warping angles $\gamma$. }
    \label{fig:warptables}

\end{center}

\section{Additional Implementation Details in LLFF}

In the LLFF dataset~\cite{MildenhallSCKRN19}, we only use a fine optimization scheme to stabilize the training of ReVoRF and gradually improve the geometric details. During the whole training period, we set the values of $\lambda_{rel}$ and $\lambda_{unr}$ as $10^{-1}$ and $10^{-3}$, respectively. The values of $\lambda_{d}, \lambda_{f}$, and $\lambda_{ds}$ are set as \scinum{5}{-5}, \scinum{5}{-6}, and \scinum{5}{-4} in the fine stage.